\documentclass{article}
\usepackage[utf8]{inputenc}
\pdfoutput=1
\usepackage{fullpage}
\usepackage{hyperref}

\usepackage{graphicx}
\usepackage{comment}
\usepackage{amsmath,amssymb} 
\usepackage{color}
\usepackage[utf8]{inputenc}
\usepackage{epsfig}
\usepackage{graphicx}
\usepackage{amsmath}
\usepackage{amssymb}
\usepackage{optidef}
\usepackage{floatrow}
\usepackage[font=scriptsize]{caption}

\usepackage[font=footnotesize]{caption}
\newfloatcommand{capbtabbox}{table}[][\FBwidth]

\usepackage[square,sort,comma,numbers]{natbib}

\usepackage{natbib}

\usepackage{booktabs}
\usepackage{tabularx}
\usepackage{adjustbox}
\usepackage[caption=false]{subfig}
\usepackage[flushleft]{threeparttable}
\usepackage{multirow}
\usepackage{array,etoolbox}
\preto\tabular{\setcounter{magicrownumbers}{0}}
\newcounter{magicrownumbers}
\usepackage{authblk}

\def\rownumber{}

\begin{document}

\title{Class-Agnostic Segmentation Loss and Its Application to Salient Object Detection and Segmentation}

\author[$\dag$]{Angira Sharma}
\author[$\dag$]{Naeemullah Khan}
\author[$\S$]{Ganesh Sundaramoorthi}
\author[$\dag$]{Philip Torr}

\affil[$\dag$]{University of Oxford} \affil[$\S$]{KAUST}
\date{}

\maketitle

\begin{abstract}
	In this paper we present a novel loss function, called class-agnostic segmentation (CAS) loss. With CAS loss the class descriptors are learned during training of the network. We don't require to define the label of a class a-priori, rather the CAS loss clusters regions with similar appearance together in a weakly-supervised manner. Furthermore, we show that the CAS loss function is sparse, bounded, and robust to class-imbalance. We apply our CAS loss function with fully-convolutional ResNet101 and DeepLab-v3 architectures to the binary segmentation problem of salient object detection. We investigate the performance against the state-of-the-art methods in two settings of low and high-fidelity training data on \textbf{seven} salient object detection datasets. For low-fidelity training data (incorrect class label) class-agnostic segmentation loss outperforms the state-of-the-art methods on salient object detection datasets by staggering margins of around  \textbf{50\%}.  For high-fidelity training data (correct class labels) class-agnostic segmentation models perform as good as the state-of-the-art approaches while beating the state-of-the-art methods on most datasets. In order to show the utility of the loss function across different domains we also test on general segmentation dataset, where class-agnostic segmentation loss outperforms cross-entropy based loss by huge margins on both region and edge metrics\footnote{Code available at \url{https://github.com/sofmonk/class_agnostic_loss_saliency}}.
	
\end{abstract}

\section{Introduction}

\begin{figure}[t!]
	\centering
	
	\begin{adjustbox}{width=0.7\textwidth}
		\begin{tabular}{p{1cm}|cccc||cccc|}
			Data & \multicolumn{4}{c}{Training Data }& \multicolumn{4}{c}{Testing} \\ 
			\hline	 & Image & Ground Truth & Image & Ground Truth& Image & Ground Truth & CE & CAS  \\ \hline
			\multirow{2}{*}{HFD }& & & & 
			\\		
			& 
			\includegraphics[width=1.8cm]{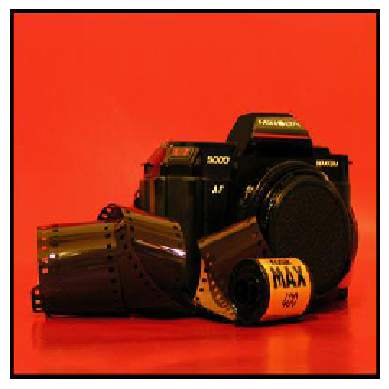} &
			
			\includegraphics[width=1.8cm]{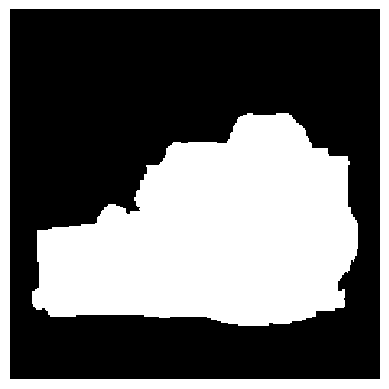} &  		
			\includegraphics[width=1.8cm]{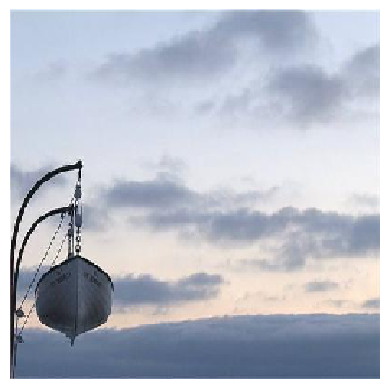} &
			
			\includegraphics[width=1.8cm]{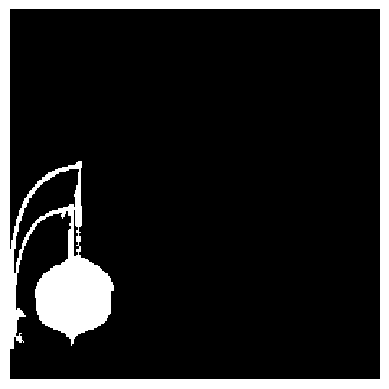} &  	\includegraphics[width=1.8cm]{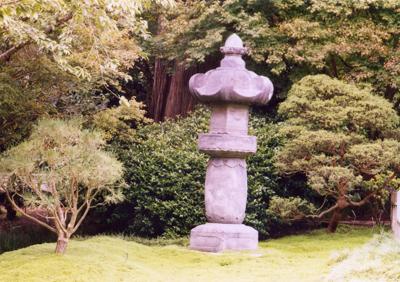} & \includegraphics[width=1.8cm]{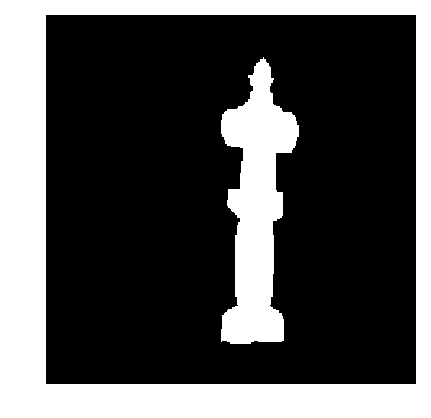}   & 	\includegraphics[width=1.8cm]{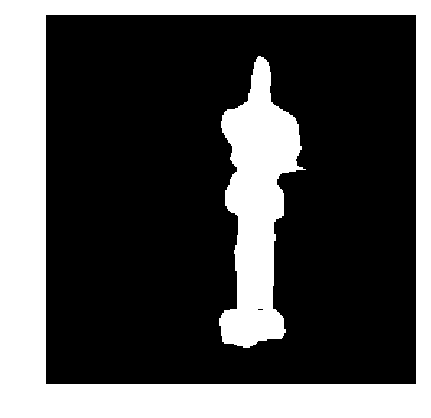} & \includegraphics[width=1.8cm]{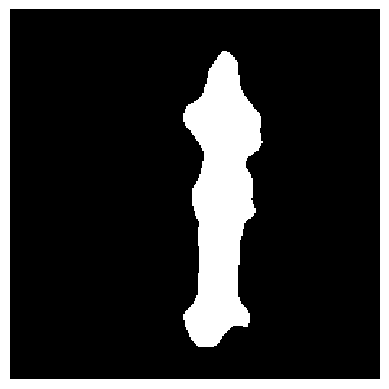} \\
			
			& \includegraphics[width=1.8cm]{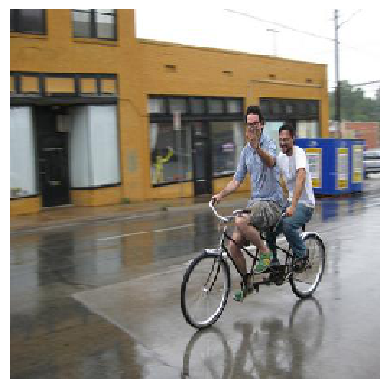} &
			\includegraphics[width=1.8cm]{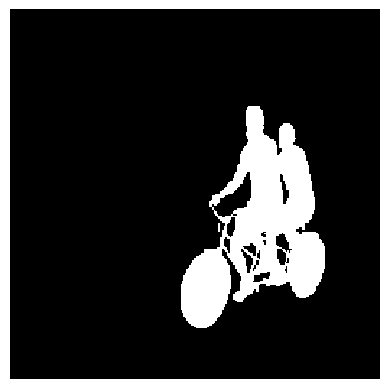} & 	\includegraphics[width=1.8cm]{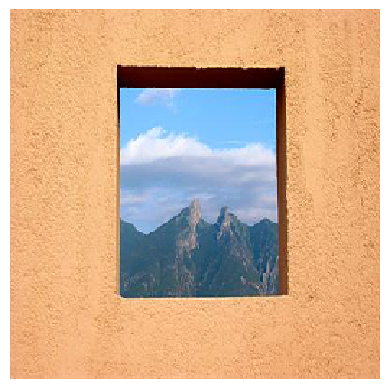} &
			\includegraphics[width=1.8cm]{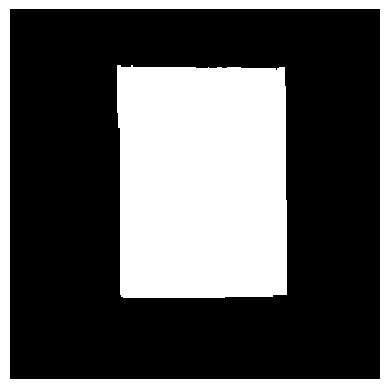} &
			\includegraphics[width=1.8cm]{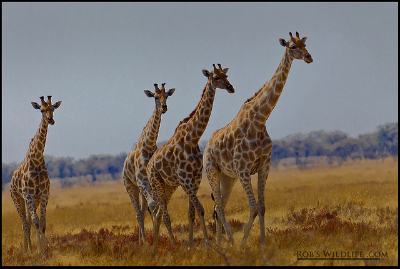} & \includegraphics[width=1.8cm]{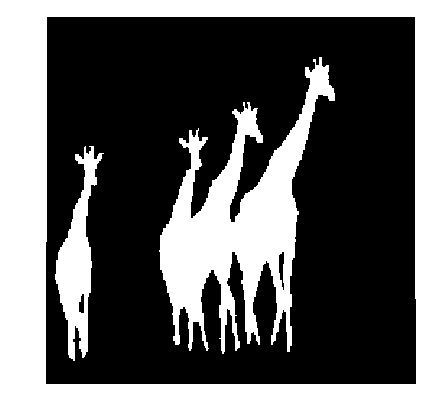}   & 	\includegraphics[width=1.8cm]{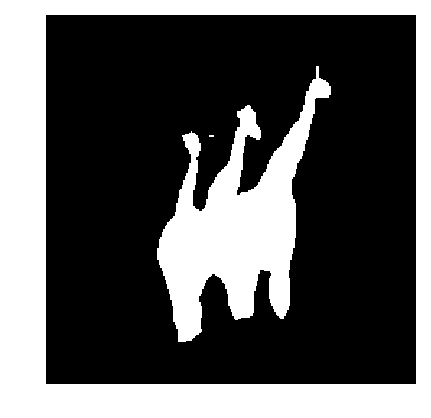}  & \includegraphics[width=1.8cm]{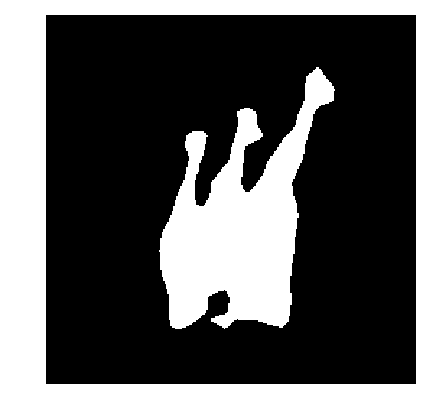} \\ \hline

			\multirow{2}{*}{LFD} & & & & &
			\\

			& \includegraphics[width=1.8cm]{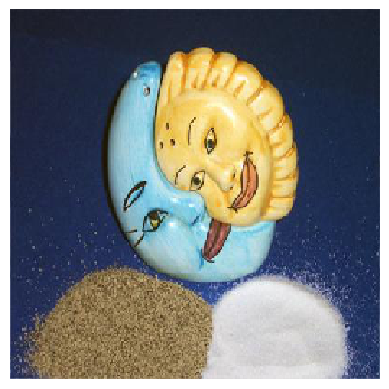} & \includegraphics[width=1.8cm]{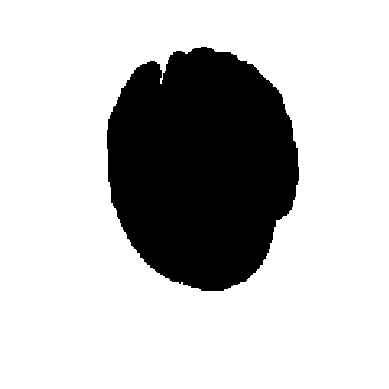} 
			& \includegraphics[width=1.8cm]{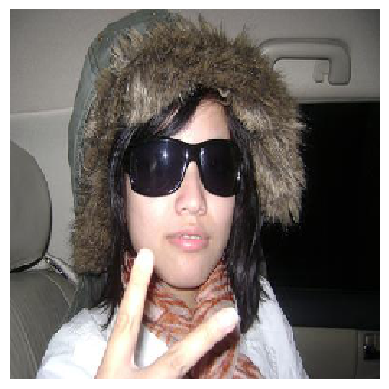} & \includegraphics[width=1.8cm]{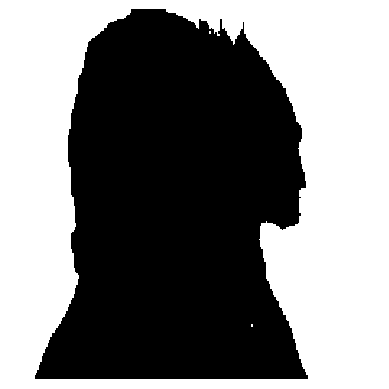} & \includegraphics[width=1.8cm]{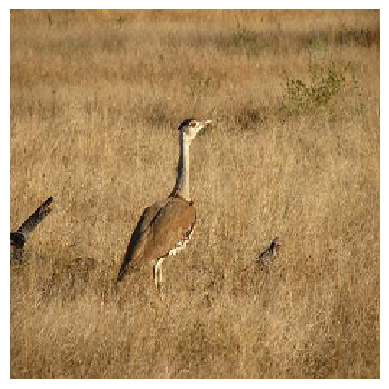} & \includegraphics[width=1.8cm]{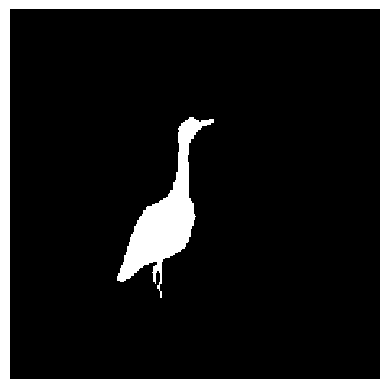}   & 	
			\includegraphics[width=1.8cm]{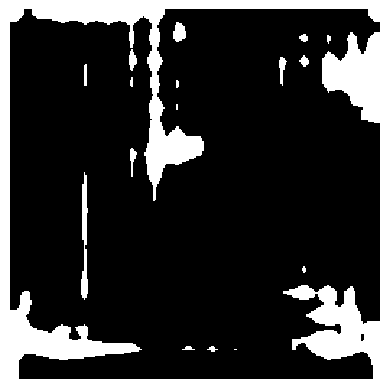} &	\includegraphics[width=1.8cm]{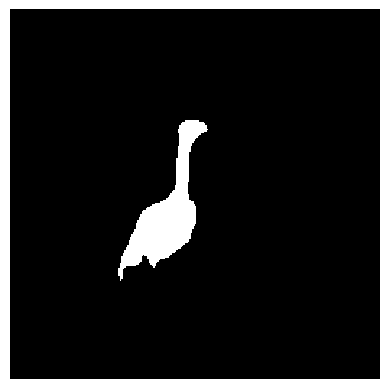} 	 \\

			& \includegraphics[width=1.8cm]{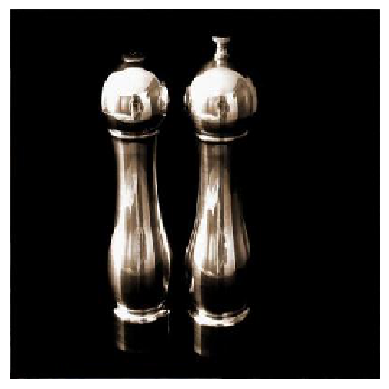} & \includegraphics[width=1.8cm]{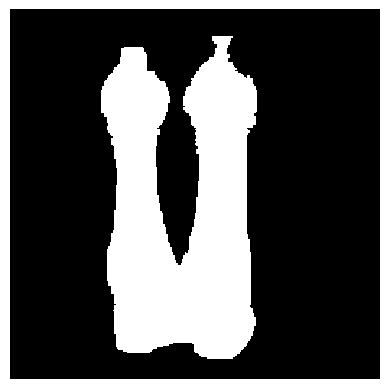}& \includegraphics[width=1.8cm]{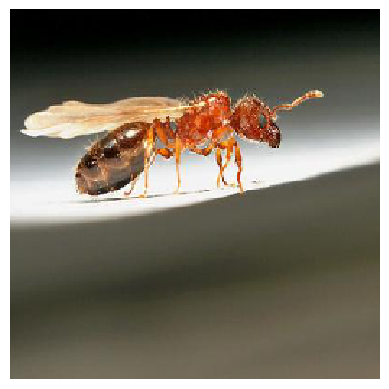} & \includegraphics[width=1.8cm]{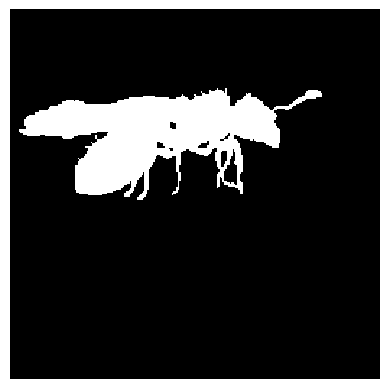} & \includegraphics[width=1.8cm]{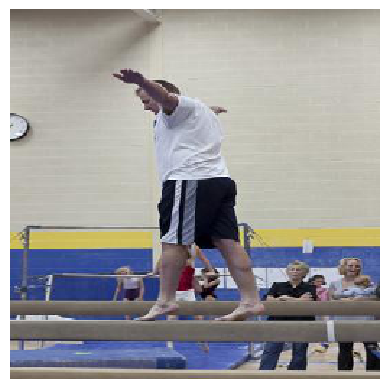} & \includegraphics[width=1.8cm]{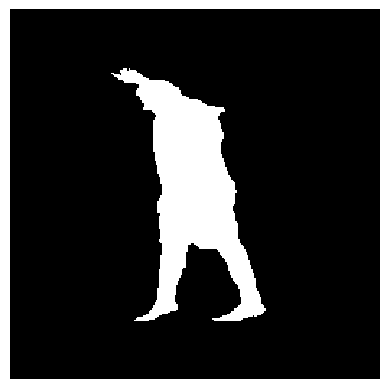}   & 		\includegraphics[width=1.8cm]{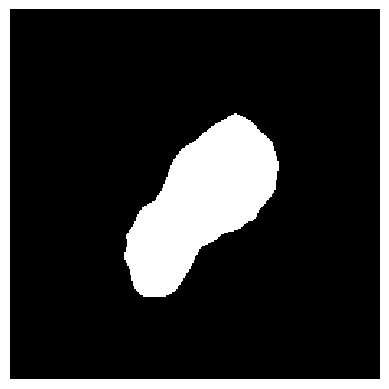} & 	\includegraphics[width=1.8cm]{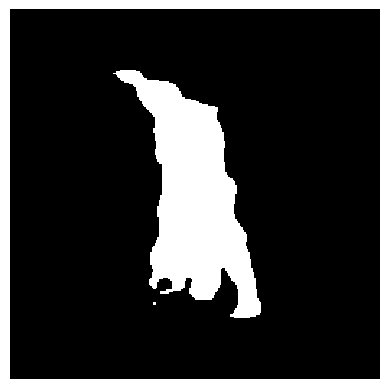}  \\
			
			\bottomrule
			
		\end{tabular}
	\end{adjustbox}
	\caption{\textbf{Motivation Example:} Traditional segmentation/detection methods work well in HFD setting (high-fidelity training data, where class labels are available and accurate) but fail completely in LFD setting (low-fidelity training data, where class labels are either not available or incorrect). Our Class-Agnostic Segmentation (CAS) loss performs well in both cases, since it is independent of the class label and learns to performs unsupervised clustering of the learned descriptors during training to obtain class labels. Whereas, conventional cross-entropy (CE) based methods completely fail in LFD setting since they are reliant on class label. }
	\label{intro}
\end{figure}

Deep learning based methods have achieved state-of-the-art results in many vision applications \cite{Liu2017}. The success of these methods is primarily reliant on the fidelity of the datasets they are trained on. For applications like segmentation, recognition and detection, most deep learning based methods use a classification based training approach, where cross-entropy (CE) or a variant of cross-entropy is used to fit the descriptors (network output) to an arbitrarily pre-assigned class label. Generating these class labels on a dataset of sufficient size is labour-intensive, restricts research and is prone to human errors \cite{Everingham2015}. For network trained with standard loss functions like cross entropy, these labelling errors result in significant degradation of performance (Table \ref{table:adv}). We present a class-agnostic segmentation (CAS) loss using which we can train networks independent of these class labels and avoid the consequent degradation of performance. An example of this for salient object detection is shown in Figure \ref{intro}.

In segmentation we divide an image into regions of unique statistics \cite{Haralick1985}. Segmentation has applications in compression, tracking, and recognition. There is a significant amount of literature that tackles segmentation. Classical approaches of segmentation either relied on region-based descriptors, where descriptors were grouped to achieve unique segments, or on edge-based approach, where edges are extracted from an image and post-processing steps like watershed methods are used to obtain region segments.

The current approaches for segmentation and detection, which conventionally use cross-entropy loss or a variant of cross entropy loss,  have the following drawbacks: 1) The arbitrarily pre-assigned class label might not necessarily correspond to the representation of the class in the descriptor space. 2) This pre-assignment introduces a hard constraint in labelling the data where the same object has to be labelled with the  same label across the entire dataset (thousands of images). For large datasets, data annotation will be performed by a huge pool of moderately trained annotators and there will inevitably be errors in label of objects (since these labels are arbitrary and do not correspond to any fundamental notion of appearance). 3) The number of classes that one can sample is limited and consequently can not be generalised to the infinite number of classes that exist in real life scenes. 4) Using the conventional loss functions the output components of a neural network will have to match the number of classes for these datasets, which will result in very large output vector for large number of classes. 5) The loss functions used are agnostic to the notion of class appearance and simply learn to group similarly labelled objects together.

In this paper we tackle the above challenges. We present a class-agnostic segmentation loss, which does not  need the class label for segmentation, rather the loss's construct is  based on unsupervised clustering of learned descriptor to obtain unique segments and relies only on forms of inexpensive ground truth  annotations. This allows us to cast the general segmentation problem with deep networks, since we do not need the class labels anymore (general segmentation as opposed to semantic segmentation divides image into unique regions and is not limited to a few classes). Rather the network is constructed to cluster similar appearances together while maximising the inter-cluster distance. This eases the task of annotation (and data generation) as one can use off-the-shelf (edge-based) segmentation methods to semi-automate the data-segmentation task (and assign no class labels whatsoever).

The contributions of our work are: 1) A new class-label agnostic segmentation loss function, which relies on ground truth annotation only and clusters similar segments together by grouping pixels with similar appearance (learned descriptor) together. 2) The class-agnostic segmentation loss function is applied to  salient object segmentation and  achieves state-of-the-art results despite the fact that we don't use any pre-training or data augmentation that other state-of-the-art methods use. 3) The class-agnostic loss function is applied to general segmentation task and outperforms cross-entropy based losses by a significant margin.

Notice, that we have used saliency detection and general segmentation as applications for our class-agnostic segmentation loss. However, the application of our loss is not restricted to these applications and they can be applied to any general segmentation problem.

\section{Related Work}

\label{sod} \label{seg}

Segmentation methods generally follow one of the two approaches: general approach, where all segments are labelled region-wise based on appearance, and semantic approach, where objects are labelled with class-labels in the dataset. Hence, semantic approach simplifies a general segmentation task to dense classification problem limited to a few classes. 
Successful segmentation models \cite{Chen2017} are based on semantic approach, where the models' abilities are limited by the number of pre-defined classes.

General segmentation algorithms can be broadly classified into region-based methods and edge-based methods. In region-based methods segments are obtained by grouping descriptors together. Traditional region-based methods suffer from inaccurate segmentation results near the boundaries of objects because statistics are aggregated across the boundaries. In \cite{Khan2015} better shape-tailored descriptors were introduced to tackle this problem. Building on \cite{Khan2015}, in \cite{Khan2017} continuum scale-space of heat equation was used to obtain coarse-to-fine segmentation, but the descriptors here are still hand-crafted and lack the capacity to discriminate between wide range of textures in natural images.

In edge-based segmentation methods pixels in images are classified to either belong to the edge class or otherwise. Post-processing  methods like watershed methods are then applied to the edge maps to obtain region segments. The regions obtained through these approaches are not based on descriptor consistency and hence these methods fails particularly in cases when textures with large textons are present in images \cite{Khan2015}.

Deep learning methods for general segmentation are primarily edge-based methods. These approaches have shown to achieve better results \cite{Xie2015}, \cite{Chen2016}, \cite{Kokkinos2015} on edge detection. Such approaches have used deep networks to derive the probability of a pixel belonging to boundaries between segments. Despite the fact that these approaches achieve impressive results on detecting edges,  generating segmentation from edges is still hard and relies on hand-crafted approaches \cite{Kokkinos2015}, hence,  the segmentation problem remains unsolved. 

An attempt to cast general segmentation as a region-based learning problem came from \cite{Khan2018}, where the authors tried to tackle the problem of segmentation by learning a metric to discriminate between shape-tailored descriptors using the Siamese twin networks. However, \cite{Khan2018} used fully connected layers to learn the discrimination metric on shape-tailored descriptors computed in pre-processing step.  Similar to \cite{Khan2018}, \cite{Kong} uses a metric learning scheme to cluster similar pixels together using mean shift algorithm building on \cite{Newell}, \cite{Fathi}. \cite{Kong} is a metric (and embedding space) learning scheme which can discriminate between pixels pair from similar region and different regions. These are different contrastive loss based metric learning schemes \cite{Chopra2005a}.   \cite{Weinberger2009} uses a distance metric scheme for large margin nearest neighbour classification. \cite{Ke} defines a affinity field loss based on distance metric learning but relies on adversarial weightings to solve semantic segmentation. Such metric learning approaches are computationally very expensive for object detection, $\mathcal{O}(N^2)$ where $N$ is the number of pixels in the image. Also, these methods suffers severely from class imbalances since the sampling for pixel pair exacerbates the class imbalance. These methods also don’t learn a 'class representation' jointly with the metric learning. Our method is novel in terms of learning the ‘class representation’ jointly with the metric learning without adding any computation overhead. Along similar lines \cite{DeBrabandere2017} introduced a discriminative loss. Discriminative loss is used for instance segmentation where different instances are clustered in the descriptor space. The loss does not maximize intra class variance. Discriminative loss is based on two hyperparameters for inter/intra class variance and the loss is simply a penalty which forces the inter/intra class variance to be close to these hyperparameters. This loss is (can) not (be) used for differentiating different classes, it can only classify different instances. \cite{DeBrabandere2017} uses a cross entropy loss for classification of different class. \cite{Neven} implements an embedding loss function for instance segmentation, but relies on learning instance specific margin and  expensive post-processing step. To the best of our knowledge, ours is the first work to successfully apply a loss based on distance metric learning principles for the task of salient object detection.

In this work we present a new class-agnostic loss function which allows the use of any deep network architecture for general segmentation problem. To the best of our knowledge this is the first attempt at general region-based segmentation with deep networks where properties of regions are learned rather than a metric for discriminating pixels. Contrary to most metric learning based methods we learn the proxy class label jointly with the descriptor during training and the complexity of our method is $\mathcal{O}(N)$ where $N$ is the number of pixels.

Since we apply our class-agnostic loss to the binary segmentation problem of salient object detection, we present a brief literature review of salient object detection here. In supervised salient object detection methods, both the input images and the ground truth annotations with class labels, are used for training. The class labels are binary,  generally, 1 is used to represent salient object and 0 for  non-salient objects. 

Some state-of-the-art salient detection methods (which are all supervised), like  \cite{Houa},  are based on combining feature maps from different layers of CNN to obtain saliency map.  PFAN  network \cite{Zhao2019} uses hand-crafted feature extraction method and channel-wise attention mechanism to extract the most important features in the intermediate layers to generate more accurate saliency maps, however, this results in suboptimal solutions as shown in  \citep{Zhang2017a, Wang2018a}.  Other state-of-the-art methods such as PoolNet \cite{Liu2019} aggregate high-level information from customised  global modules built on top of feature pyramid networks coupled with edge detection at intermediate level of the network.

These current state-of-the-art methods require training on large datasets containing real-world data, ideally on cluttered background. However, this kind of data for salient object detection is limited, therefore, majority of the methods such as DeepNet \cite{Pan2016} and DeepFix \cite{Kruthiventi2017} are based on networks pre-trained on the ImageNet \cite{JiaDeng2009} dataset coupled with data augmentation  \cite{Zhao2019}. On the contrary, almost all models in our work do not use any pre-training or data augmentation to keep the complexity minimal.

In  semi-supervised  and weakly-supervised frameworks there are diverse approaches to solve salient object detection. 
Some techniques \cite{Wang2017b} infer potential foreground regions to perform global smooth pooling operation and combine these responses to generate saliency maps. Weak supervision methods, such as CPSNet \cite{Zeng2019}, also rely on multiple sources such as image captions, incomplete or incorrect labels, and multiple images cues, which are  passed to more than one networks followed by  inter-network feature sharing to output the final saliency map. These type of complex operations result in slow forward computation.

Modified versions of cross entropy loss have also been  explored  to solve salient object detection. For instance, \cite{Pan2017}  proposes a variant of cross entropy loss using generative adversarial networks \cite{Goodfellow2014}. Another work, BAS-Net \cite{Qin2019}, proposes a hybrid loss, which is a combination of cross entropy, structural similarity index (SSIM) and intersection over union (IOU).  As these loss functions  are majorly modified versions of  cross entropy loss, they depend on the true class labels for computation. This dependence becomes an obstacle in the scarcity of labelled data.

\subsection{Problem Statement}

We design a loss function which trains the network to clusters pixels of similar appearance together in a weakly-supervised manner. Our loss function forces the descriptors to have low variance on regions/objects, while at the same time the descriptor learns to discriminate between different regions.

\section{Class-Agnostic Segmentation (CAS) Loss}

In this section we present the class-agnostic segmentation loss and derive the backpropagation equation when this loss is applied to standard networks.

\subsection{CAS Loss Function}

The class-agnostic segmentation loss is defined as:
\begin{align} \label{CAS_formula}
\textit{CAS} = \sum_{i = 1}^{N} \int_{r_i}^{} \underbrace{\dfrac{\alpha ||\mathbf{s}(x)-\mathbf{\hat{s}}(r_i)||_{2}^{2}}{|r_i|}dx}_\text{Uniformer}  - \sum_{i=1}^{N}\sum_{\substack{j=1 \\ i\neq j}}^{N} (1-\alpha)\underbrace{||\mathbf{\hat{s}}(r_i) - \mathbf{\hat{s}}(r_j)||_{2}^{2}}_\text{Discriminator}
\end{align}
where $N$ is the number of regions in the ground truth mask;     $r_1,..., r_i, r_j, ..., r_N$ denotes the region of the ground truth mask (a particular segment);    $|r_i|$ denotes the number of pixels in the region $r_i$; $\mathbf{s} = \{s^1, ..., s^m,..., s^M\}$ is  a vector of output descriptor components (or softmax output) of the network; $m \in \{1,...,M\}$ where $M$ denotes the number of output (softmax) channels i.e., number of units in the last layer of the network;   $\alpha \in [0, 1]$ is a scalar, a weighing hyper-parameter which assigns weight to each term; for a region $r$  we have that,  $\mathbf{\hat{s}}(r) = \{\hat{s}(r)^{1}, ...,\hat{s}^{m}(r),..., \hat{s}(r)^{M}\}$ is a vector containing channel-wise mean of the descriptor values; where for a channel $m$,  $\hat{s}^{m}(r)= \frac{1}{|r|}\int_{r} s^{m} (x)\ dx$. In our formulation $\mathbf{\hat{s}}(r)$ acts as a proxy for class label for region $r$.

The uniformer term of the loss function reduces the variance of the learned descriptor on the regions (segments). The discriminator term increases the distance between the learned descriptors for different regions. These are the two essential properties required of any successful descriptor for segmentation i.e. small intra-class variance and large inter-class discriminability. The uniformer term ensures the invariance of a descriptor on a region of interest and the discriminator term ensures that different region have different descriptors. Hence, the combination of the two terms trains the model to perform segmentation based on the appearance (rather than class labels). One aspect to note is that we have used squared euclidean distances for both the terms; this is for the simplicity of implementation, but the general framework of class-agnostic segmentation loss will work for any suitable norm.

\subsection{Gradient of CAS Loss}

The gradient of the loss function with respect to the weights $\boldsymbol{\omega}$ of a deep network is (for more details see Appendix \ref{a_cal}), 
\begin{align} \label{casl_b1}
& \nabla_{\boldsymbol{\omega}} CAS = \sum_{i=1}^{N} \int_{r_i}^{} 2\dfrac{\alpha  (\mathbf{s}(x)-\mathbf{\hat{s}}(r_i))(\nabla_{\boldsymbol{\omega}}\mathbf{s}(x) - \nabla_{\boldsymbol{\omega}}\mathbf{\hat{s}}(r_i))}{|r_i|}dx  \notag \\
&-  \sum_{i=1}^{N}\sum_{\substack{j=1 \\ i\neq j}}^{N} 2(1-\alpha)(\mathbf{\hat{s}}(r_i) - \mathbf{\hat{s}}(r_j))(\nabla_{\boldsymbol{\omega}}\mathbf{\hat{s}}(r_i) - \nabla_{\boldsymbol{\omega}}\mathbf{\hat{s}}(r_j)) 
\end{align}
From Equation \ref{CAS_formula} we've, $
\nabla_{\boldsymbol{\omega}} \hat{s}^{m}(r_i) = \frac{1}{|r_i|}\int_{r_i}^{} \nabla_{\boldsymbol{\omega}} s^m(x) dx$,  where we have $\nabla_{\boldsymbol{\omega}} s^m(x)$ from the backpropagation of the network as $s^m(x)$ is a component of the softmax output of the network.

\subsection{Properties of CAS loss} \label{prop}
\vspace{-5pt}
The construction of the discriminator term introduces some interesting properties to the CAS loss and here we analyse them (the empirical results for these can be found in  Appendix \ref{a_prop} ).  The global minima for the uniformer term is any piece-wise constant descriptor where each component of descriptor is  constant on each individual region. This will make the variance (uniformer term) of the descriptor zero on each region. The discriminator term eliminates the trivial minima of  uniformer term where all descriptors on image are either zero or equal constant values. Hence the discriminator term is necessary to introduce the crucial discriminability properties to the learned descriptor.

The discriminator term can be considered as an optimization problem of the form given below; the constraints below are due to the softmax layer at the end of the network. Setting $\alpha=0$ in Equation \ref{CAS_formula} for a binary segmentation problem (with regions $r_0$ and $r_1$) leaves us with the task to,
\begin{maxi} 
	{}{||\hat{{ \textbf{s}}}(r_0)-\hat{{\textbf{ s}}}(r_1)||_2^2 }
	{}{}
	\addConstraint {\sum_{m=1}^M\hat{ s}^m(r_0)&=1 \qquad \sum_{m=1}^M\hat{ s}^m(r_1)=1 }  
	\addConstraint {\hat{ s}^m(r_0)&\geq0 \qquad \hat{ s}^m (r_1)\geq0 \qquad \forall m} 
	\label{prop}
\end{maxi}

\begin{figure}[h!]
	\centering
	
	\begin{adjustbox}{width=0.8\textwidth}
		\begin{tabular}{c|cccccccc}
			\tiny Channel 1 & 	\includegraphics[width=0.8cm]{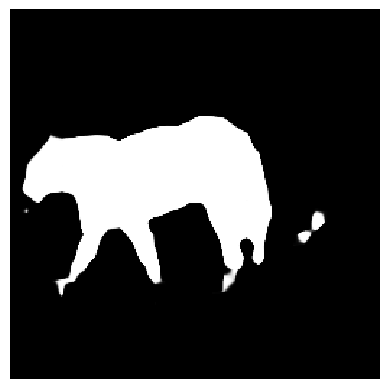} & \  \includegraphics[width=0.8cm]{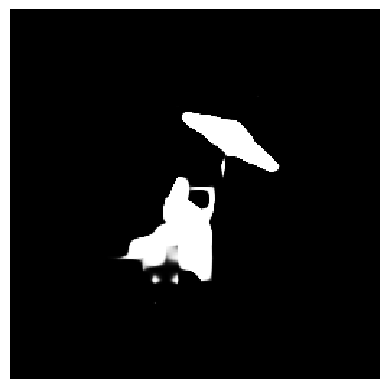}  & \ \includegraphics[width=0.8cm]{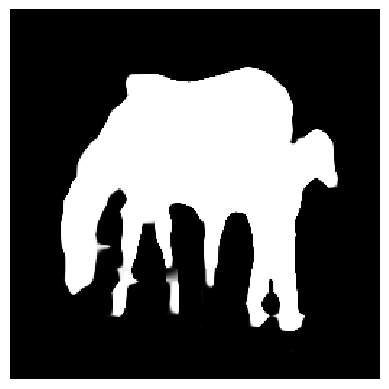} 	&\  \includegraphics[width=0.8cm]{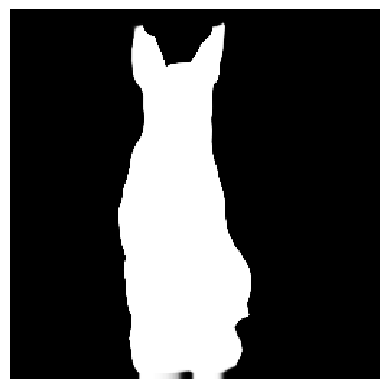} 	& \ 	\includegraphics[width=0.8cm]{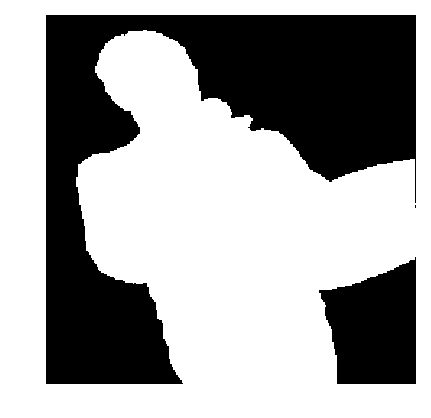} &\               \includegraphics[width=0.8cm]{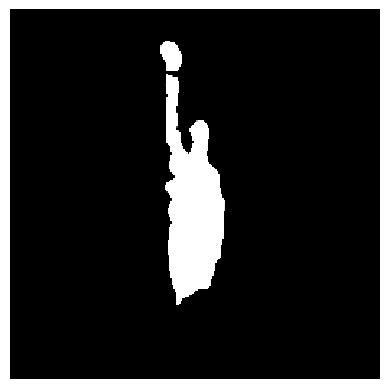} &\  \includegraphics[width=0.8cm]{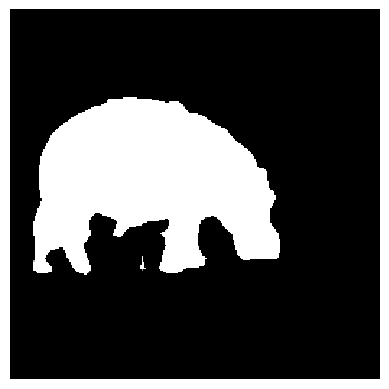} & \ \includegraphics[width=0.8cm]{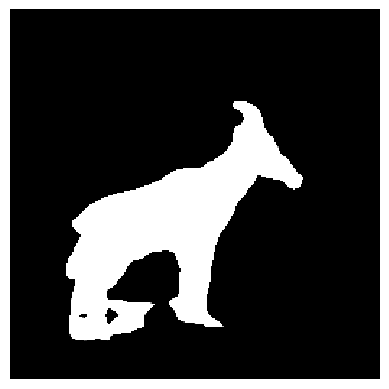}
			\\ \tiny Channel 2 
			&\  	\frame{\includegraphics[width=0.8cm]{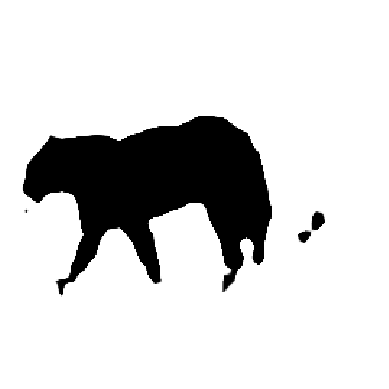}}  
			& \ 	\frame{\includegraphics[width=0.8cm]{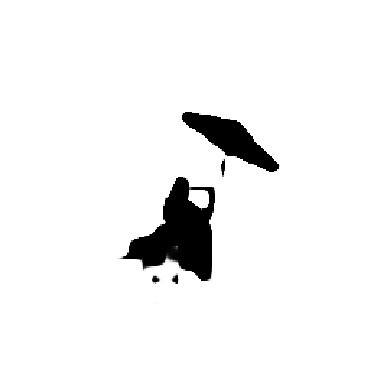}}  
			& 	 \ \frame{\includegraphics[width=0.8cm]{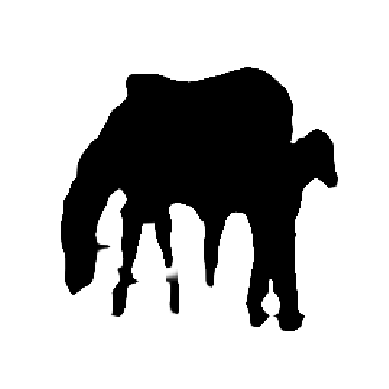}} 
			&  \	\frame{\includegraphics[width=0.8cm]{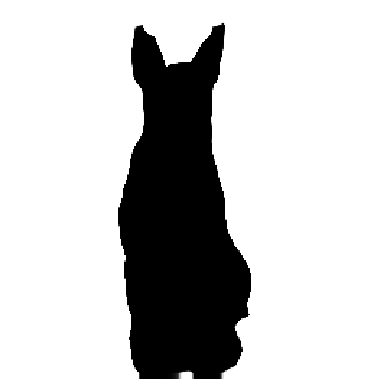}} 
			&\ 	\frame{\includegraphics[width=0.8cm]{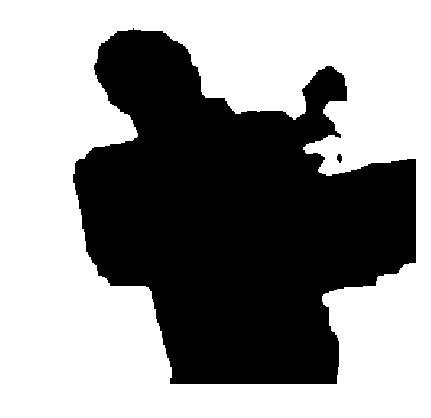}} & \	\frame{\includegraphics[width=0.8cm]{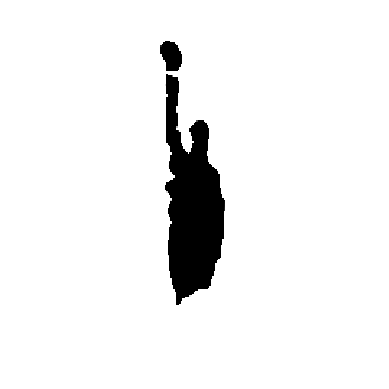}}
			& \	\frame{\includegraphics[width=0.8cm]{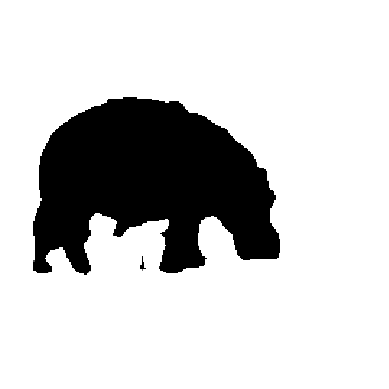}} & \
			\frame{\includegraphics[width=0.8cm]{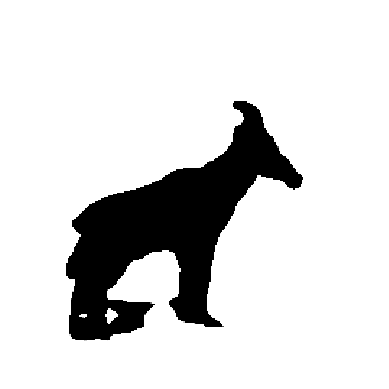}} \\
			
		\end{tabular}
	\end{adjustbox}
	\caption{\textbf{Empirical Results for Sparsity}: Sum of components of both channels is 1 and only one channel is active at a time, thus, resulting in sparse output descriptors } 
	\label{tbl:sp}
\end{figure}

\noindent Its properties are: 

\textbf{Sparsity:}
With the inequality constraints in Eq. \ref{prop} we get the feasible region of the objective. Because of the equality constraints this feasible region is bounded.  In $M$-dimensions each inequality constraint will represent a half-space. With the intersection of these half-spaces we will get a convex polytope. In this case, the optimal solutions will occur at the corners of the convex polytope, which are sparse. Here one component of each $\hat{{ \textbf{s}}}(r_0)$ and $\hat{{ \textbf{s}}}(r_1)$ is $1$ and all other components are 0, and the non-zero components for $\hat{{ \textbf{s}}}(r_0)$ and $\hat{{ \textbf{s}}}(r_1)$ lie at different indices, as shown in Figure \ref{tbl:sp}.  Thus, each unique region (or texture) will be represented by a sparse descriptor.

\begin{figure}[h!]
	\centering
	
	\begin{adjustbox}{width=0.9\textwidth}
		\begin{tabular}{c|cc  c | c c }
			
			\tiny Ground Truth & \tiny CE & \tiny CAS &  \quad  \quad \quad \tiny Ground Truth & \tiny CE & \tiny CAS \\
			\includegraphics[width=1cm]{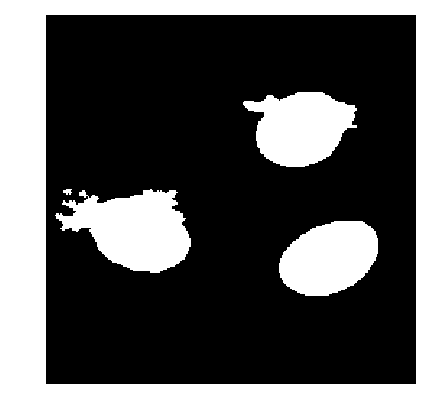}  & 
			\includegraphics[width=1cm]{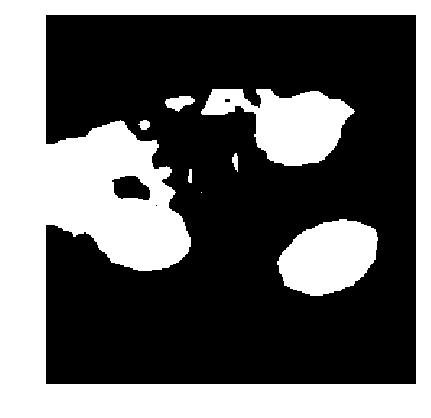} & 
			\includegraphics[width=1cm]{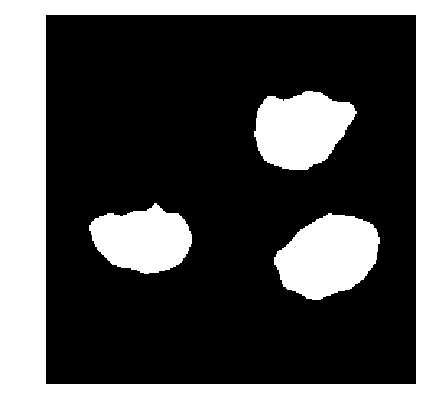} & \quad  \quad \quad	\includegraphics[width=1cm]{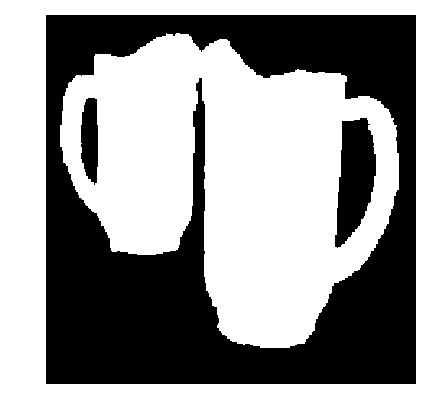}  & 
			\includegraphics[width=1cm]{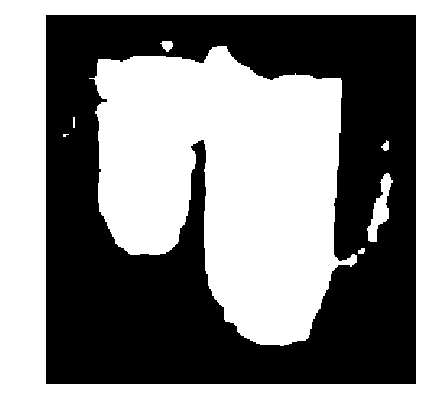} & 
			\includegraphics[width=1cm]{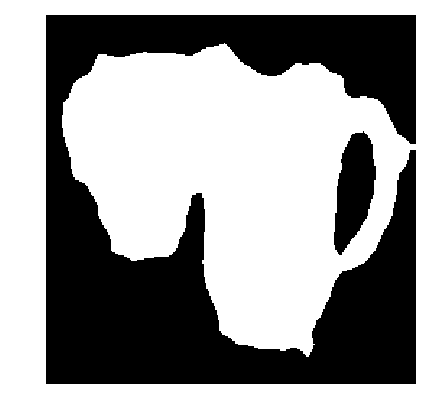} \\
			\includegraphics[width=1cm]{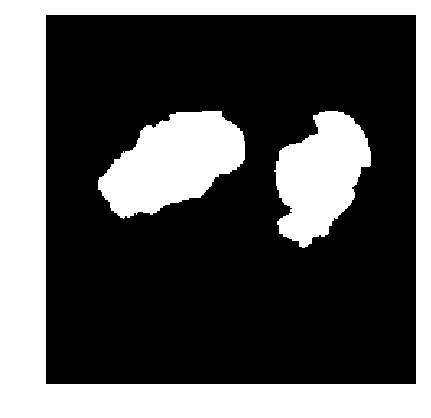} & 
			\includegraphics[width=1cm]{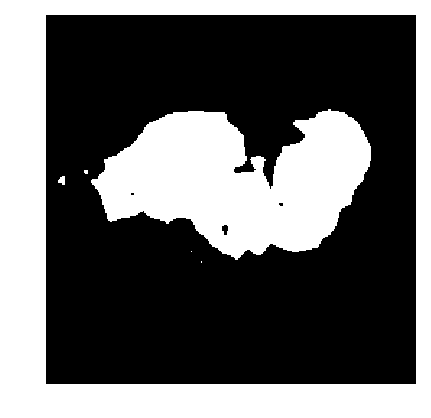} & 
			\includegraphics[width=1cm]{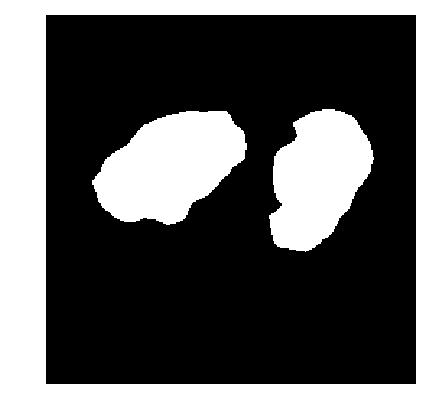} & \quad  \quad \quad	\includegraphics[width=1cm]{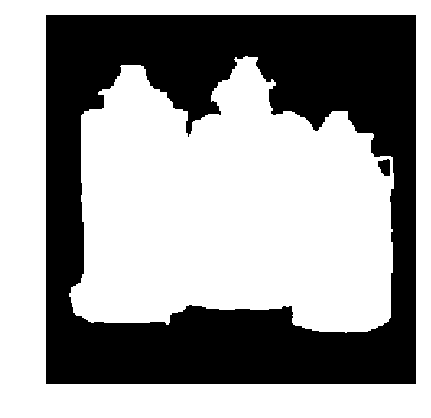}  & 
			\includegraphics[width=1cm]{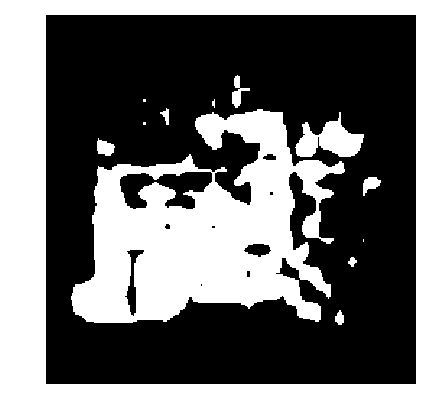} & 
			\includegraphics[width=1cm]{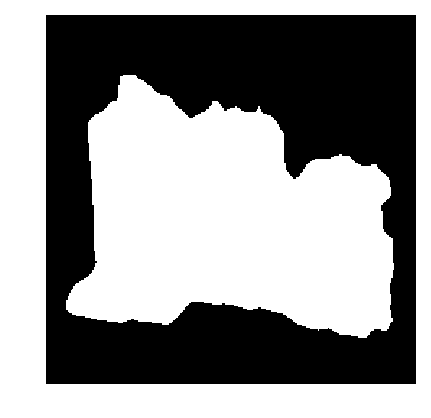}  \\
			
		\end{tabular}
	\end{adjustbox}
	
	\caption{\textbf{Empirical Results for Class Imbalance}: The CE loss fails to output disconnected or clear salient objects because of the size bias of salient objects, whereas, the CAS loss is robust to such size bias or class imbalance } 
	\label{tbl:ci}
\end{figure}

\textbf{Robustness to Class Imbalance:}
Cross entropy loss based methods for general and salient object segmentation methods suffer from  class imbalance where the larger class weighs the learning more \cite{Kingma2014}. In contrast, the CAS loss function is immune to class imbalance-induced training issues as all the terms in  Eq. \ref{CAS_formula} used for calculation are normalised by class (region) size. Therefore,  regardless of the size of  salient objects (number of pixels they cover), all regions contribute equally to the loss function. Hence, even a  small region will be represented well in the descriptor space as shown in Fig. \ref{tbl:ci}.   We didn't normalise for salient or non-salient class in CAS loss and achieved state-of-the-art results (Section \ref{tbl:sup_results}). 

\begin{figure}[h!]
	\centering
	\includegraphics[scale=0.35]{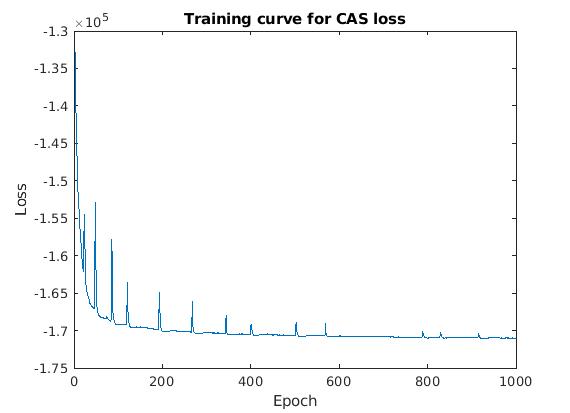}   
	\caption{Training curve for CAS loss }
	\label{training_curve}
\end{figure}

\textbf{Boundedness:}
As a consequence of using softmax outputs in the last layer of  neural network, the CAS  loss function has a defined upper  and lower bound. 
Since $s^c \in [0,1]^{w \times h}$ (where $w$ and $h$ are width and height of the output channel respectively), and we have all mean values in the loss, thus both the uniformer and discriminator terms are bounded by 1. Thus, the value of loss function lies in a bounded interval  $\left( \alpha N_i,-(1-\alpha)N_i \right]$, where $N_i$ is the number of regions in an image $i$. A loss value of $-(1-\alpha)N_i$ indicates perfect segmentation of all samples in the training set. Figure \ref{training_curve} shows the training loss for training with 
3000 images. Notice that the loss in the Figure \ref{training_curve} is well bounded within $\left( \alpha N_i,-(1-\alpha)N_i \right]$.

\section{Experimental Setup}

This section describes the experimental setup for the experiments in this paper. The codes were setup in PyTorch \cite{Paszke2017} using Python3.7. The experiments were run on Nvidia Quadro RTX 6000 GPU and Intel Xeon 2.60GHz  CPU . 

\subsection{Architecture}

The main aim of this work was  to present a  class-agnostic segmentation loss, regardless of the architecture. Therefore, we  used standard available architectures as  our models.
We used the standard FCN-ResNet-101 \cite{He2016} architecture   and DeepLab-v3 \cite{Chen2017} architecture backbones with softmax layer as the output layer of the model; also, these are being adopted and advised now as the backbone architecture for segmentation and salient object detection \cite{Zhang2017}, \cite{Borji2019b}. 

\subsection{Datasets}

For an extensive  evaluation of our methods, we tested our models on 7 datasets: MARS-B (5000 images) \cite{Ming1}   dataset consists of single salient object in an image; DUTS \cite{Wang2017b}  dataset has explicit training and testing sets which are DUTS-TR (10553 images) and DUTS-TE (5019 images) respectively; ECSSD (1000 images)\cite{Yan2013}  dataset has semantically meaningful and complex objects, and textures, containing salient objects of different sizes; PASCAL-S (850 images) \cite{Li2014}  dataset contains natural images with cluttered backgrounds; HKU-IS (4447 images) \cite{Li2016b}  dataset has multiple disconnected salient objects, some touching the boundary; THUR15k (6232 images)\cite{Cheng2014}  dataset contains random internet images and does not necessarily contain a salient object in every image and  DUT-OMRON (5168 images) \cite{Yang2013a} dataset contains one or two complex salient objects. The datasets used for training were MSRA-B \cite{Cheng2011} (split 6:4 ratio for training and testing) and DUTS-TR  \cite{Zeng}. 

\subsection{Pre-processing and Post-processing}

All images were resized to $256 \times 256$ and  standardised to have mean 0 and unit variance. Standardisation ensures that all parts of the image share equal weights, otherwise the larger pixel values  tend to dominate the weights of the neural network. After standardisation the distribution of pixels resembles a Gaussian curve centred at zero, which helps in faster convergence of the neural network. The images were then divided by 255.0.

The network has 2 sparse output channels, the correct saliency map $S$, and $1-S$.  We choose the channel for the salient object label by calculating the correlation of the output channels with the saliency map on the validation set and selecting the channel with maximum value as saliency map.  In post-processing, the continuous output saliency map $S\in [0,1 ]^{256 \times 256}$ of the neural network is  thresholded using the popular method \cite{Cheng2011}  to get a binary output map $B$, which is calculated as $B(x, y) = 
1,  \text{if}\ S(x, y) > T \text{ else }
0,  \text{if}\ S(x, y) \leq T $ where, $T = 2 \times mean(S)$ is the threshold, and $x$ and $y$ denotes the pixel positions on the map.

\subsection{Evaluation Metrics}

The standard evaluation metrics used in salient object detection are $F_\beta$-score and Mean Absolute Error (MAE). The evaluation metrics  are calculated on the ground-truth mask $G \in \{0, 1\} ^{w \times h}$ and the binary map $B \in \{0,1\}^{w \times h}$ extracted from saliency map $S\in [0,1 ]^{w \times h}$ (where $w$ and $h$ denote the width and height respectively). 

$F_\beta$-score is the weighted harmonic mean of precision and recall, with a non-negative weight $\beta$. The $F_\beta$-score is computed as, $
F_{\beta} = \frac{(1+\beta^2) Precision \times Recall}{\beta^2 Precision + Recall} $. Setting $\beta = 1$ provides the standard F-score formula. However, because the traditional $F$-score suffers  from interpolation flaw, dependency flaw and equal-importance flaw \cite{Margolin2014} which results in unfair comparison, weighted $F_\beta$ score is used \cite{Liu2018}. 
Like all previous works,  the value of $\beta^2$ was set to 0.3. 

To address the true negative saliency assignments (i.e., correctly marked as non-salient) and reward it, the MAE score is calculated  as the average of absolute error between the saliency map $S$ and the ground truth mask $G$.   $MAE = \frac{1}{w \times h} \sum_{x = 1}^{w} \sum_{y=1}^{h}|S(x, y) - G(x, y)| $, 
where $x$ and $y$ denotes the pixel positions on the map. It is desired to be as low as possible.

\subsection{Models} \label{models}

We used two standard models as the backbone for our experiments, \textit{Model-CE} model is FCN-ResNet-101 and DeepLab-v3 architecture with cross entropy loss and \textit{Model-CAS} model is  FCN-ResNet-101 and DeepLab-v3 architecture with CAS loss. Hence, in total we get 4 models, 2 with backbone FCN-ResNet-101 and 2 with DeepLab-v3. For completion of our comparison with CE based methods, we define a class-agnostic version of the cross-entropy loss function for binary segmentation as $\min(-y_i \log{p_i}, -(1-y_i)\log{p_i})$, where for $i^{th}$ example, $y_i$ is the true-class label and $p_i$ is the predicted probability of belonging to class. Notice that as we increase the number of classes  the arguments of the $\min$ function increases combinatorially in the definition of class-agnostic version of CE loss. In the low-fidelity data setting we train the  \textit{ResNet-CACE} model, which is FCN-ResNet-101 architecture with class-agnostic version of cross entropy loss. For these  models  the settings such as architecture, optimisers, activations functions and hyperparameters were same, a comparison of these models consequently results in comparison of the loss function's performance. 

\subsection{Ablation Study for Hyperparameter $\alpha$}

The CAS loss (Equation \ref{CAS_formula}) has one hyperparameter, $\alpha$, which sets the weights on the uniformer and discriminator terms. As argued in Section \ref{prop}, the discriminator term has more important properties, consequently we weighed this term more. $\alpha$ was set to 0.1. This was validated by testing for various values of $\alpha$ as shown in Table \ref{table:alpha}. The learning rate was 1e-3 with Adam \cite{Kingma2014} optimizer.

\section{Experiments}

We conducted experiments for salient object detection (SOD) in two settings, low-fidelity data setting where the data is not correctly labelled and the probability of incorrect class labels  is  0.5 (notice that the error that we interested here is the incorrect class label assignment to a region, see example in Figure \ref{intro}), and in high-fidelity data setting where the training data is correctly labelled and the class labels are reliable. We also applied our class-agnostic segmentation loss to general segmentation problem to show the general nature of the loss function for any segmentation application.

\subsection{Low-fidelity data SOD} \label{6_claims}

We implemented our class-agnostic segmentation loss framework for salient object detection  with low-fidelity training data (probability of incorrect class label 0.5). Training data in standard datasets is labelled correctly. To setup the low-fidelity data setting, we flipped the labels of half of the training samples  in a way that the ground truth segments remain intact i.e., we flipped the class label 'value' (to simulate low-fidelity training data).  We trained 3 state-of-the-art models PFAN \cite{Zhao2019}, BAS-Net \cite{Qin2019} and PoolNet \cite{Liu2019},  and 4 of our models, ResNet-CE, ResNet-CAS, ResNet-CACE and DeepLab-CAS in this setting. 

The purpose of this experiment was to empirically prove the class-agnostic property of CAS loss i.e., even if class labels are not correct, the CAS loss still learns to perform salient object detection (SOD) (see Appendix \ref{a_lfd} for more details). Consequently, the annotators do not need to assign the class labels to the segments as long as labels for different regions are distinct. The experiment simulates a class label error scenario where class labels for segments are incorrect with probability 0.5 (worst scenario in binary classification) i.e., region label was different from class label (as shown in row 3 in Figure \ref{intro}). In other words, the foreground (salient) was labelled as background (non-salient) \& vice-versa with a probability of 0.5. It is to show that even in worst case CAS loss works, consequently it works  for cases where the class label is incorrect for fewer training images which we show in  Table \ref{table:lfd_perc}.  We also show  that the class-agnostic version of cross-entropy fails in this setting.

\subsection{High-fidelity data SOD}

To empirically verify the segmentation ability (tasks of uniformer and discriminator) of  CAS  loss against the state-of-the-art methods, we compared these methods in high-fidelity data setting. Here the training data included the images and ground truth masks with accurate class labels. Following the brief discussion in Section \ref{models}, in total the following 7 models, were trained until convergence: \textit{ResNet-m-CE}, 
\textit{ResNet-m-CAS},  \textit{ResNet-d-CE}, \textit{ResNet-d-CAS}, where $m$ and $d$ denote training on MSRA-B and DUTS-TR datasets respectively,   \textit{ResNet101-pre-CAS} model was pre-trained on MSRA-B using cross entropy loss and then trained  on MSRA-B using CAS loss, \textit{DeepLab-CE} and \textit{DeepLab-CAS} models were trained on DUTS-TR dataset, pretrained on COCO \cite{Lina} dataset.

\subsection{Texture Segmentation}

Experiments were conducted on real-world segmentation dataset presented in \cite{Khan2015}. This is one of the most challenging general segmentation datasets \cite{Khan2015} with huge intrinsic and extrinsic variability of textures. The dataset is divided into 128 training and 128 testing images. We tested our CAS loss and cross-entropy loss and showed that CAS outperforms CE loss by a significant margin. We trained using CAS and CE loss on multiple state-of-the-art architectures (ResNet and DeepLab-v3) and showed that CAS is more suited to general segmentation task compared to CE loss.

\section{Results}

The quantitative results for salient object detection in low-fidelity training data setting are summarised in Table \ref{table:adv} and Table \ref{table:lfd_perc}, and a few qualitative samples are show in Figure \ref{tbl:sup_results} (for the remaining models the results are in Appendix \ref{a_d}). Since state-of-the-art methods are based on CE type loss which is highly reliant on fidelity of class labels, these methods fail completely in low-fidelity training data cases, with performance drops of around 50 \% on most datasets. This essentially means that pixels are randomly labelled as salient or non-salient. Our CAS loss is immune to any performance degradation in low-fidelity training data (and trains the model in a class-agnostic manner). ResNet-CAS results in low-fidelity case degrades only slightly whereas for DeepLab-CAS the performance improves.

We also compare performance of our CAS loss with state-of-the-art methods in high-fidelity training data setting. The quantitative results are summarized in Table \ref{table:supervised} and a few qualitative samples are shown in Figure \ref{tbl:sup_results}. Models trained using CAS loss perform equally well to the sate-of-the-art methods and at times superior to these models. Our models achieve state-of-the-art results on 5 out of the 7 datasets. No other state-of-the-art method performs that well consistently on that many datasets.

Another aspect to note is that without any tweak in the FCN-ResNet101 or DeepLab-v3 architectures our models beat  the state-of-the-art results. There is a good probability that with further tweak or using different networks we might beat the state-of-the-art with even larger margins. However, we would like to restate that the key contribution of this work is the introduction of the CAS loss, and hence we did not tune the network architectures for superior results.

It is also worth mentioning that we performed the most basic level of distortion to the class labels to generate low-fidelity training data; even if the class label \textit{1} was mislabelled as \textit{10} in some images, the CAS loss would still work. This is a consequence of dependence of CAS loss on $s(r_i)$ and $|r_i|$ but not the label of $r_i$ (Equation \ref{CAS_formula}).

Results for texture segmentation experiments are summarised in Table \ref{tab:results_Nets} and Figure \ref{fig:deep_results} (visual results for DeepLab-v3 are in Appendix 	Fig. \ref{fig:text}). Notice that CE loss fails to learn any useful descriptor with both ResNet \cite{He2016} and DeepLab \cite{Chen2017} architectures because the real-world texture dataset is annotated with ground truth segments and not class labels. On the other hand, the CAS loss outperforms CE loss and performs reasonably well in capturing complex texture segments. Undeniably the results are not perfect since we have a very small texture segmentation training set at hand, nonetheless the point about the strength of CAS loss over CE loss is well demonstrated in this experiment. \\ \\ \\ \\

\begin{table*}[h!]
	\begin{center}
		\begin{adjustbox}{width=0.99\textwidth}	
			\begin{threeparttable}
				\begin{tabular}{|c|cc|cc|cc|cc|cc|cc|cc|}
					\toprule
					Model & \multicolumn{2}{c|}{MSRA-B} & \multicolumn{2}{c|}{DUTS-TE} & \multicolumn{2}{c|}{ECSSD} & \multicolumn{2}{c|}{PASCAL-S} & \multicolumn{2}{c|}{HKU-IS} &  \multicolumn{2}{c|}{THUR15k} &
					\multicolumn{2}{c|}{DUT-OMRON} \\ 
					&  $F_\beta \uparrow$ & MAE $\downarrow$ & $F_\beta \uparrow$ & MAE  & $F_\beta\uparrow$ & MAE  $\downarrow$  & $F_\beta \uparrow$ & MAE   $\downarrow$& $F_\beta\uparrow$ & MAE $\downarrow$ & $F_\beta\uparrow$ & MAE  $\downarrow$  &  $F_\beta\uparrow$ & MAE  $\downarrow$  \\
					\hline\hline
					PFAN \cite{Zhao2019} & 0.580 & 0.502 & 0.532 & 0.512 & 0.588 & 0.510 & 0.611 & 0.493 & 0.565 & 0.518 & 0.541 & 0.513 & 0.537 & 0.511 \\
					BAS-Net \cite{Qin2019}  & 0.585 & 0.619 & 0.444 & 0.682 & 0.572 & 0.619 & 0.655 & 0.623 & 0.527 & 0.637 & 0.431 & 0.660 & 0.442 & 0.686 \\
					PoolNet \cite{Liu2019}	& 0.603 &0.502 &  0.566 & 0.501 & 0.597 & 0.503 & 0.648 & 0.480 & 0.582 & 0.503& 0.517 & 0.505 & 0.526 & 0.498 \\
					\midrule
					
					ResNet-CE &  0.691 & 0.140 & 0.427 & 0.191 & 0.625 & 0.178 & 0.592 & 0.206 & 0.633  &0.160 & 0.670& 0.150 &  0.666 & 0.147\\
					ResNet-CACE &  0.872 &\textcolor{blue}{  0.076} & 0.808 & 0.103 & 0.803 & 0.114 & 0.754 & 0.150 & 0.822  & 0.094 & 0.834 & 0.108 & 0.816  & 0.095 \\
					ResNet-CAS & \textcolor{blue}{ 0.920} & \textcolor{red}{ 0.038} & \textcolor{blue}{ 0.836} & \textcolor{blue}{ 0.077} & \textcolor{blue}{ 0.837} & \textcolor{blue}{ 0.085} &  \textcolor{blue}{0.773} & \textcolor{blue}{ 0.126}  & \textcolor{blue}{0.856}& \textcolor{blue}{0.067} & \textcolor{blue}{0.865} & \textcolor{blue}{0.080} & \textcolor{blue}{0.846} & \textcolor{blue}{0.069}\\

					DeepLab-CAS  & \textcolor{red}{0.937} & \textcolor{red}{0.038} & \textcolor{red}{0.868} & \textcolor{red}{0.064} & \textcolor{red}{0.875} & \textcolor{red}{0.068} & \textcolor{red}{0.810} & \textcolor{red}{0.110} & \textcolor{red}{0.896} & \textcolor{red}{0.050} & \textcolor{red}{0.893} & \textcolor{red}{0.070} & \textcolor{red}{0.871} & \textcolor{red}{0.058} \\
					\bottomrule

				\end{tabular}
				\begin{tablenotes}
					\item[\textcolor{red}{red}] represents our best score value on the dataset; \textcolor{blue}{blue} represents the second best score on the dataset\\
				\end{tablenotes}
			\end{threeparttable}
		\end{adjustbox}
	
		\caption{Results on Low-fidelity training on MSRA-B data \vspace{-10pt}}
		\label{table:adv}
	\end{center}
\end{table*}

\begin{table*}[h!]
	\begin{center}
		\begin{adjustbox}{width=0.99\textwidth}	
			\begin{threeparttable}
				\begin{tabular}{|p{3.2cm}|c c|c c|cc|cc|cc|cc|cc|}
					\toprule
					Model & \multicolumn{2}{c|}{MSRA-B} & \multicolumn{2}{c|}{DUTS-TE} & \multicolumn{2}{c|}{ECSSD} & \multicolumn{2}{c|}{PASCAL-S} & \multicolumn{2}{c|}{HKU-IS} &  \multicolumn{2}{c|}{THUR15k}
					& \multicolumn{2}{c}{DUT-OMRON}\\ 
					&  $F_\beta \uparrow$ & MAE $\downarrow$ & $F_\beta \uparrow$ & MAE $\downarrow$ & $F_\beta\uparrow$ & MAE  $\downarrow$  & $F_\beta \uparrow$ & MAE  $\downarrow$& $F_\beta\uparrow$ & MAE $\downarrow$ & $F_\beta\uparrow$ & MAE  $\downarrow$  & $F_\beta\uparrow$ & MAE  $\downarrow$  \\
					\midrule \hline
					
					BAS-Net \cite{Qin2019} &  -  & - & 0.860   &    0.047   &\textcolor{blue}{0.942}   &  0.037 & 0.854 & 0.076 & 0.921 & 0.039 & - & - & 0.805 & 0.056\\
					
					PoolNet \cite{Liu2019}	 & -&-& 0.892 & \textcolor{red}{0.036}& \textcolor{red}{0.945} & 0.038 & \textcolor{blue}{0.880}&\textcolor{red}{0.065} & \textcolor{blue}{0.935}&\textcolor{blue}{0.030}&-&- & 0.833 & \textcolor{blue}{0.053}\\
					
					CPSNet \cite{Zeng2019}	&-&-&-&-& 0.878& 0.096 & 0.790 &0.134   &-&-&-&- & 0.718 & 0.114\\
					
					PFAN \cite{Zhao2019} &  - & - & 0.870 & \textcolor{blue}{0.040} & 0.931 & \textcolor{red}{0.032} & \textcolor{red}{0.892} & \textcolor{blue}{0.067} & 0.926 & 0.032 & -&- & 0.855 & \textcolor{red}{0.041} \\		
					PAGENET+CRF\cite{Wang2019} & - & - & 0.817 & 0.047  &  0.926 & \textcolor{blue}{0.035}  & 0.835 & 0.074 & 0.920 & \textcolor{blue}{0.030} & - & -  & 0.770 & 0.063 \\ 
					PAGENET\cite{Wang2019}  & -   & -   & 0.815 & 0.051  & 0.924 & 0.042 &  0.835 & 0.078 & 0.918 & 0.037 & -&- & 0.770 & 0.066\\ 
					
					HED \cite{Houa}	 & 0.927 & \textcolor{blue}{0.028}& -   &  -  & 0.915 & 0.052  & 0.830&0.080 & 0.913 & 0.039 &-&-&0.764 & 0.070 \\
					
					DNA 	\cite{Liu2019a} & - &-& 0.873 & \textcolor{blue}{0.040} & 0.938 & 0.040 & -&-& 0.934 &\textcolor{red}{0.029}&  0.796 & 0.068 & 0.805 & 0.056\\

					\midrule
					ResNet-CAS (ours)& \textcolor{red}{$0.985^{`}$}& \textcolor{red}{$0.010^{`}$} & $0.871^{+}$ & $0.071^{+}$ & $0.888^{+}$ & $0.071^{`}$ & $0.840^{+}$ & $0.112^{`}$ & \textcolor{red}{$0.939^{+}$} & $0.050^{+}$ & $0.931^{+}$ & $0.073^{+}$ & \textcolor{blue}{$0.876^{`}$} & $0.066^{`}$ \\
					
					ResNet-CE (ours)& \textcolor{blue}{$0.958^{*}$}  & $0.030^{*}$  & \textcolor{red}{$0.919^{+}$}  &$0.055^{+}$  & $0.905^{*}$ & $0.068^{*}$ & $0.876^{+}$ & $0.091^{+}$ & $0.928^{+}$ & $0.044^{+}$ & \textcolor{red}{$0.935^{+}$} & \textcolor{red}{$0.057^{+}$}  & \textcolor{red}{$0.920^{*}$} & $0.059^{*}$\\ 
					
					DeepLab-CAS (ours) & 0.931 & 0.040 & 0.850 & 0.070 & 0.864 & 0.072 & 0.800 & 0.111 & 0.882 &0.054 & 0.888 & 0.069 & 0.865 & 0.060\\
					
					DeepLab-CE (ours)& 0.928 &  0.039 & 0.847 & 0.070 & 0.867 & 0.069 & 0.805 & 0.110 & 0.880 & 0.052 & 0.881 & 0.070 & 0.856 & 0.061\\ 
					
					\bottomrule
					
				\end{tabular}
				\begin{tablenotes}
					\item[]	* represents model trained on MSRA-B dataset, + represents model trained on DUTS-TE dataset, ` represents model pre-trained on cross-entropy, 		\textcolor{red}{red} represents the  best score value on the dataset, \textcolor{blue}{blue} represents the second best score on the dataset, - represents the dataset was not tested by the method\\
				\end{tablenotes}
			\end{threeparttable}
		\end{adjustbox}
		\caption{ Numerical Results on High-fidelity data setting }
		\label{table:supervised}
	\end{center}
\end{table*}

\begin{figure*}[h!]
	\centering

	\begin{adjustbox}{width=1\textwidth}
		\begin{tabular}{@{\makebox[3em][r]{\rownumber\space}} |cc||c|c|c|c|c||c|c|c||c|c|c|c|c|}
			\toprule
			Image & Ground Truth & ResNet-pre-CAS&  ResNet-m-CAS & ResNet-d-CAS & ResNet-m-CE & ResNet-d-CE & ResNet-CAS &ResNet-CE& ResNet-CACE& BAS-Net \cite{Qin2019} &PoolNet \cite{Liu2019} & CPSNet \cite{Zeng2019} &PFAN \cite{Zhao2019} & Low-fidelity trained sota\\
			
			\midrule
			
			&& \multicolumn{5}{c|}{High-fidelity trained Our Models} & \multicolumn{3}{c|}{Low-fidelity trained Our Models} & \multicolumn{4}{c|}{State-of-the art} &\\ 
			\gdef\rownumber{\stepcounter{magicrownumbers}\arabic{magicrownumbers}} \\
			\midrule
			
			\includegraphics[width=1.8cm]{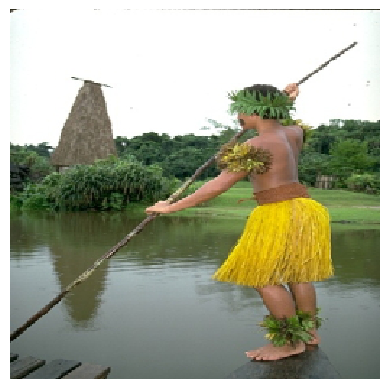} & \includegraphics[width=1.8cm]{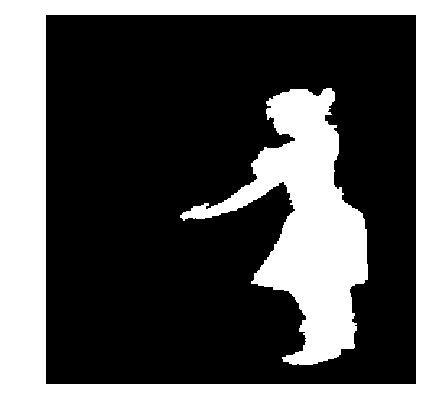}   & 		\includegraphics[width=1.8cm]{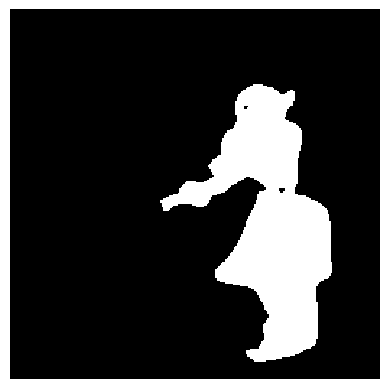} & 		\includegraphics[width=1.8cm]{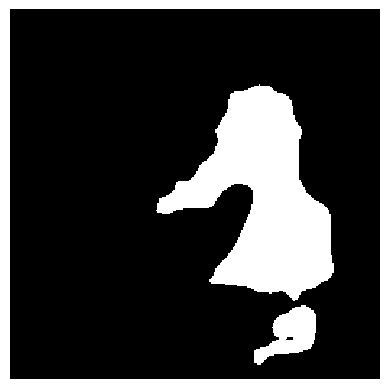} & 
			
			\includegraphics[width=1.8cm]{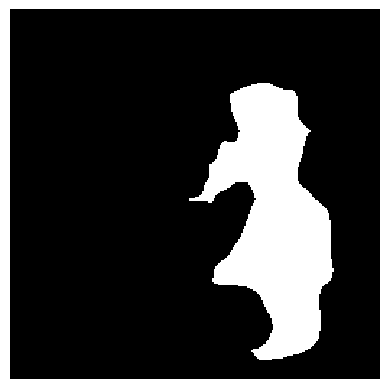} &  
			\includegraphics[width=1.8cm]{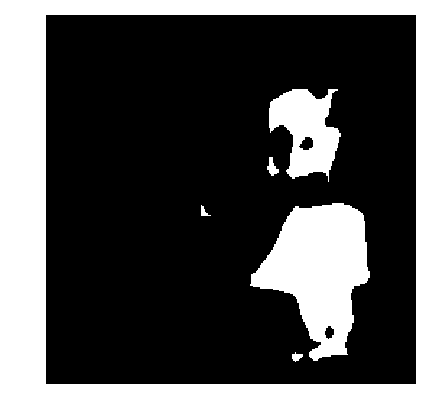} & \includegraphics[width=1.8cm]{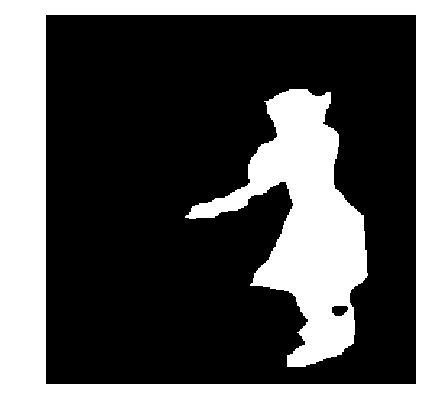} 
			& 	
			\includegraphics[width=1.8cm]{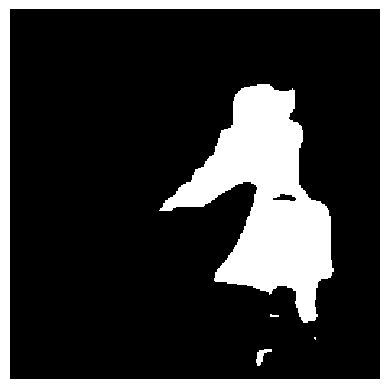} & \includegraphics[width=1.8cm]{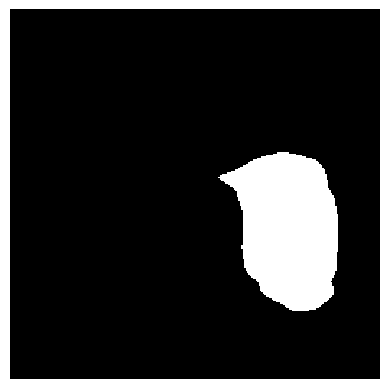} 
			& \includegraphics[width=1.8cm]{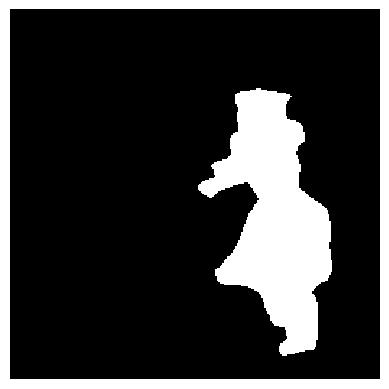} 	&

			\includegraphics[width=1.2cm]{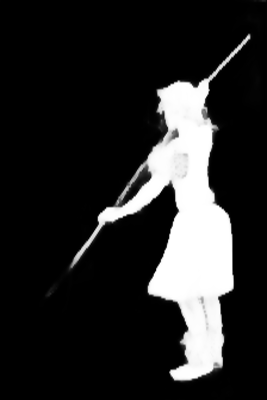} & 		\includegraphics[width=1.2cm]{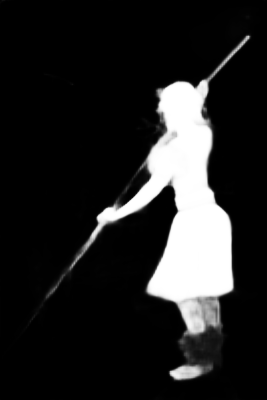}
			&\includegraphics[width=1.2cm]{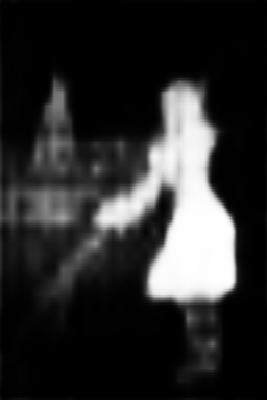} & 		\includegraphics[width=1.2cm]{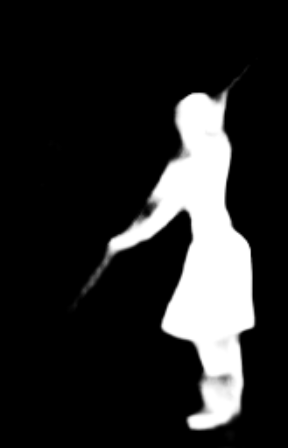} &  \includegraphics[width=1.8cm]{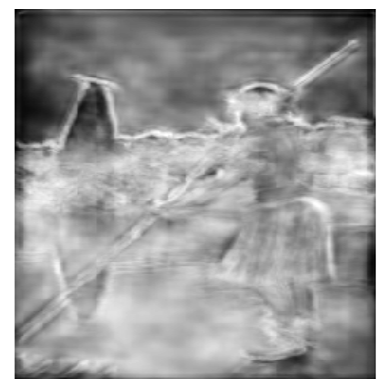}(PFAN)\\
			\midrule

			\includegraphics[width=1.8cm]{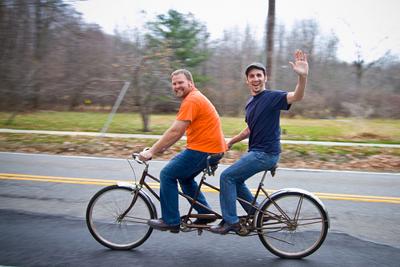} & \includegraphics[width=1.8cm]{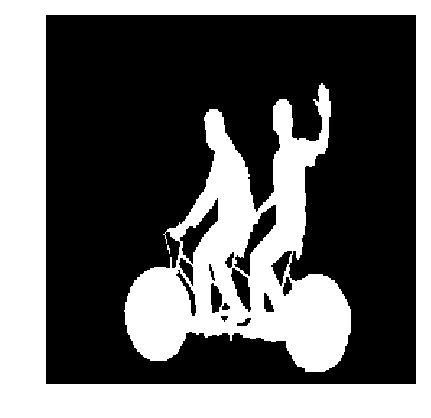}   & 		\includegraphics[width=1.8cm]{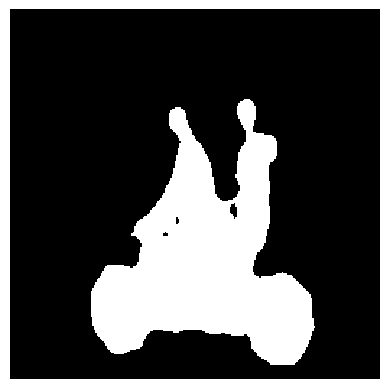} & 		\includegraphics[width=1.8cm]{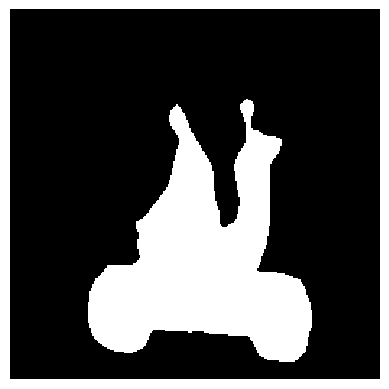} &
			\includegraphics[width=1.8cm]{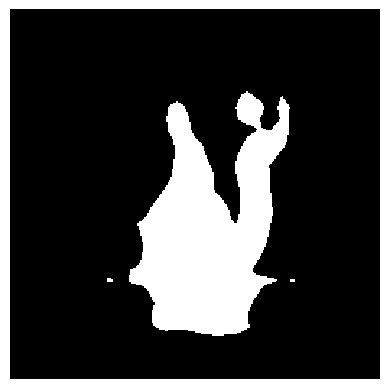}  	&
			\includegraphics[width=1.8cm]{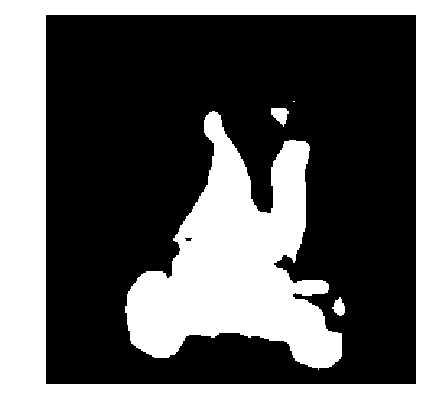} &
			\includegraphics[width=1.8cm]{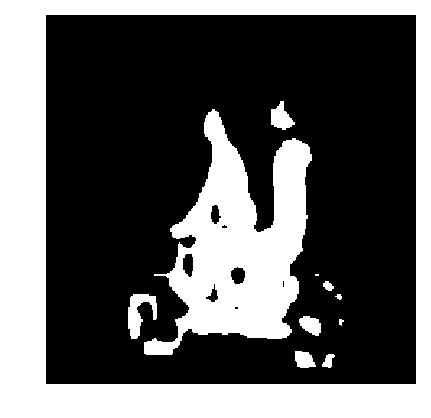}  & 
			
			\includegraphics[width=1.8cm]{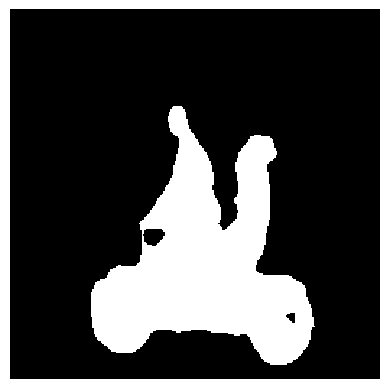} & \includegraphics[width=1.8cm]{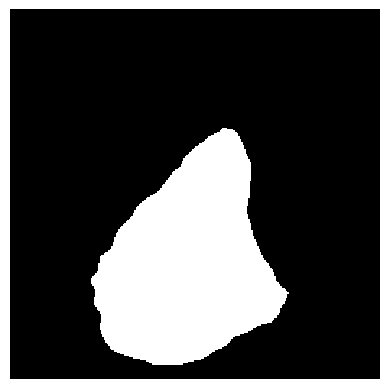} 
			& \includegraphics[width=1.8cm]{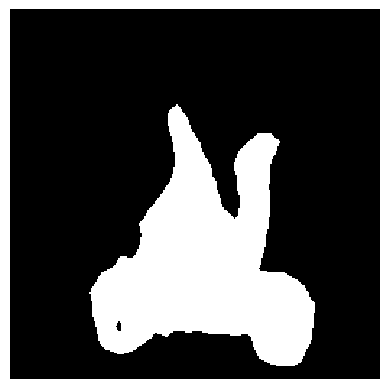} 	& 
			
			\includegraphics[width=1.8cm, height=1.8cm]{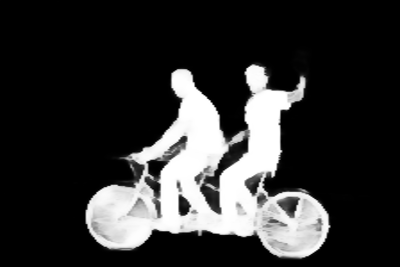} & 		\includegraphics[width=1.8cm, height=1.8cm]{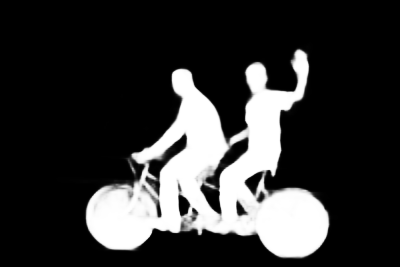}
			&\includegraphics[width=1.8cm, height=1.8cm]{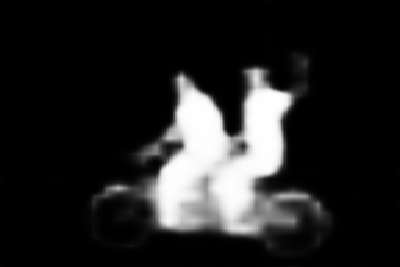} & 		\includegraphics[width=1.8cm, height=1.8cm]{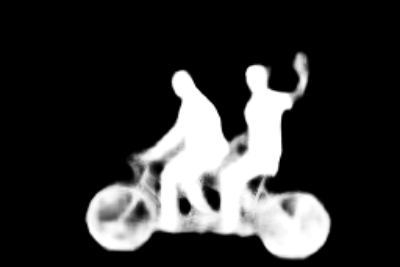} & \includegraphics[width=1.8cm, height=1.8cm]{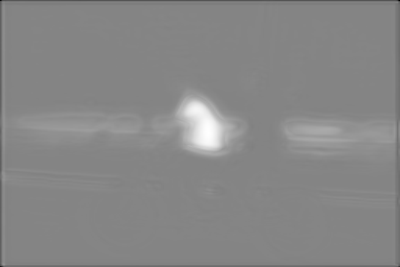}(BasNet)\\
			\midrule

			\includegraphics[width=1.8cm]{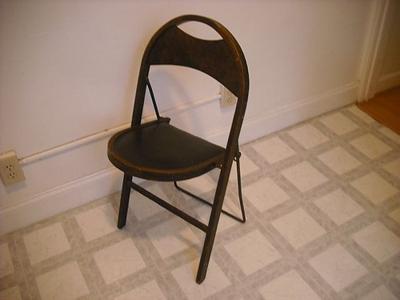} & \includegraphics[width=1.8cm]{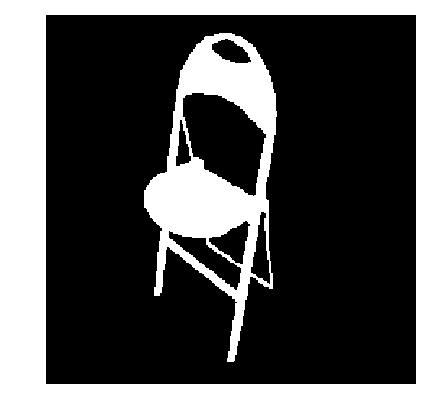}   & 		\includegraphics[width=1.8cm]{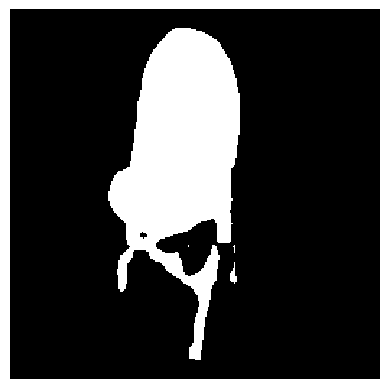} & 		\includegraphics[width=1.8cm]{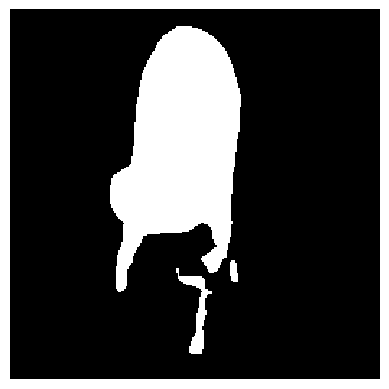} & 	
			\includegraphics[width=1.8cm]{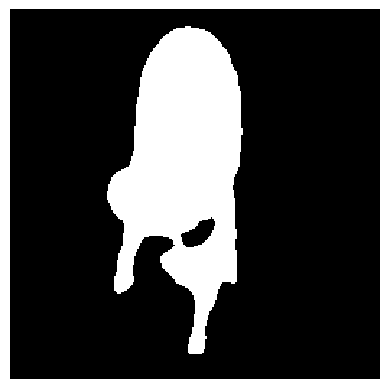}& 
			\includegraphics[width=1.8cm]{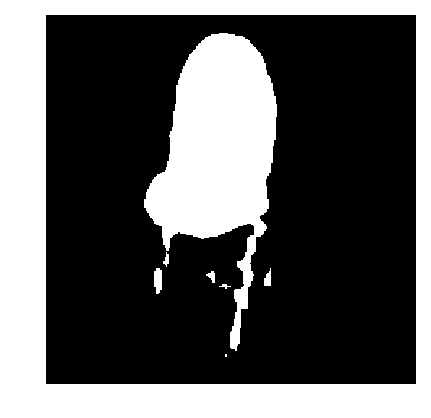}&
			\includegraphics[width=1.8cm]{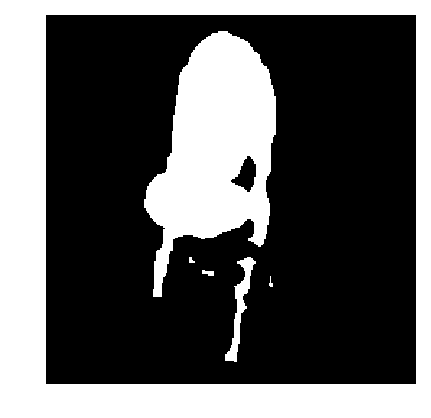}& 
			\includegraphics[width=1.8cm]{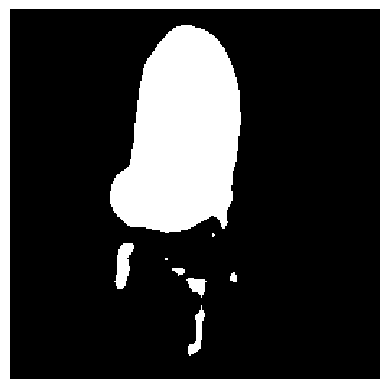} & \includegraphics[width=1.8cm, height=1.8cm]{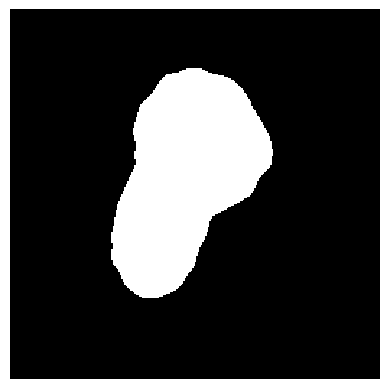} 
			& \includegraphics[width=1.8cm, height=1.8cm]{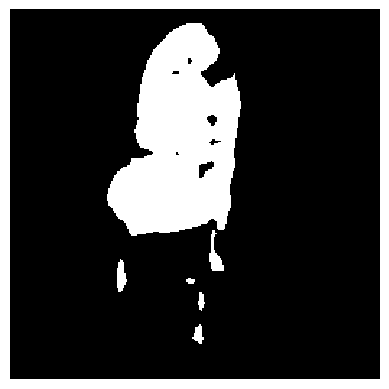} 	&

			\includegraphics[width=1.8cm, height=1.8cm]{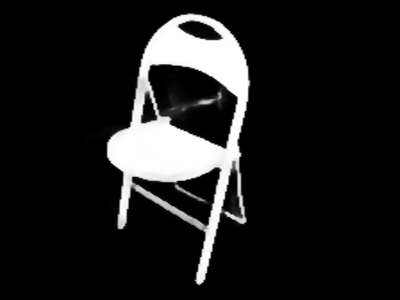} & 		\includegraphics[width=1.8cm, height=1.8cm]{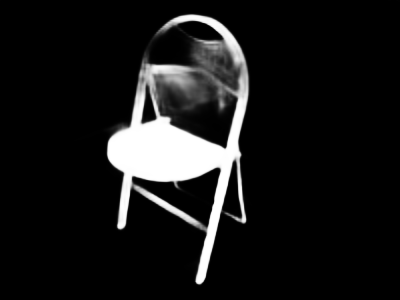}
			&\includegraphics[width=1.8cm, height=1.8cm]{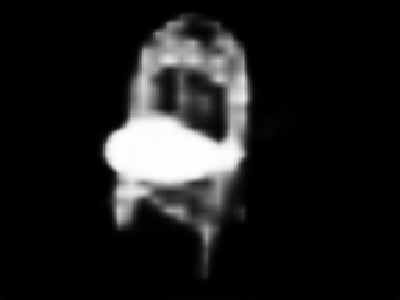} & 		\includegraphics[width=1.8cm, height=1.8cm]{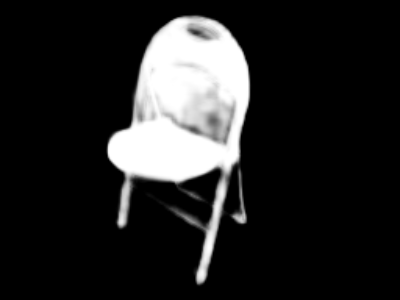} & \includegraphics[width=1.8cm,height=1.8cm]{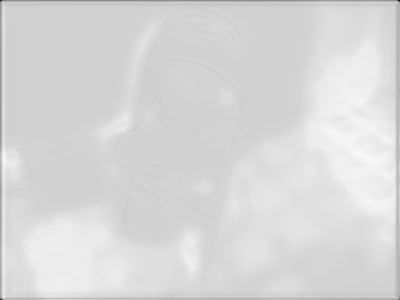} (BasNet)\\
			\midrule

			\includegraphics[width=1.8cm]{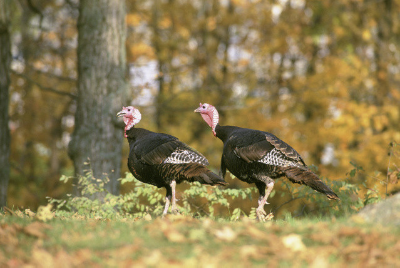} & \includegraphics[width=1.8cm]{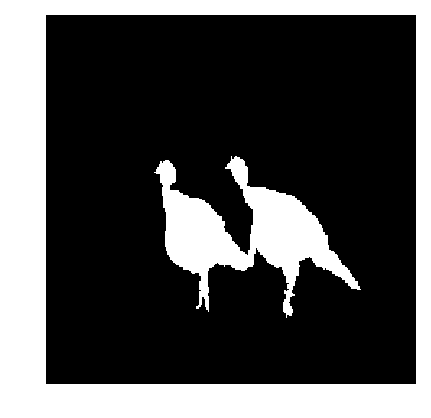}   & 		\includegraphics[width=1.8cm]{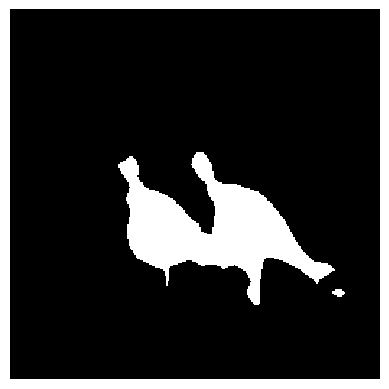} & 		\includegraphics[width=1.8cm]{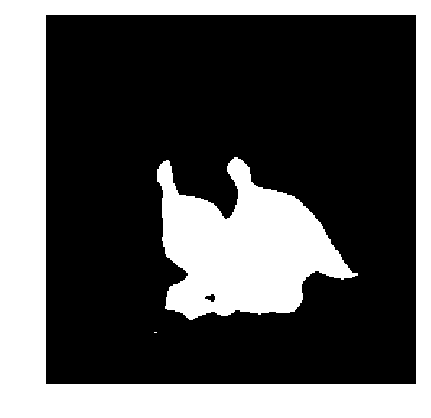} & 	\includegraphics[width=1.8cm]{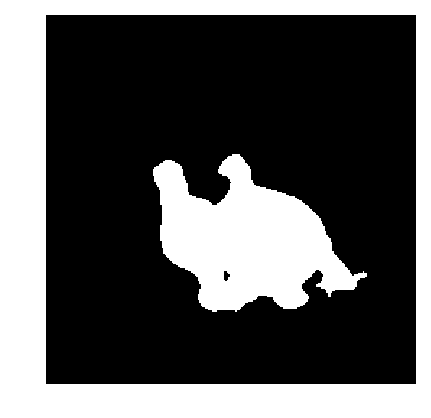}&
			\includegraphics[width=1.8cm]{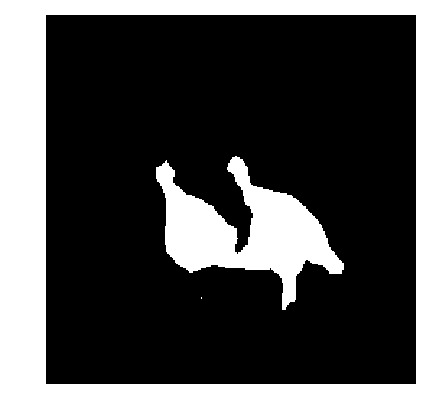} &
			\includegraphics[width=1.8cm]{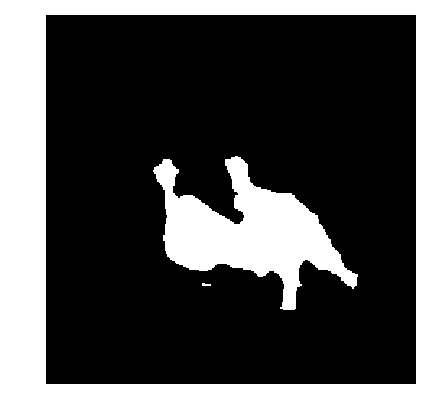} &
			
			\includegraphics[width=1.8cm]{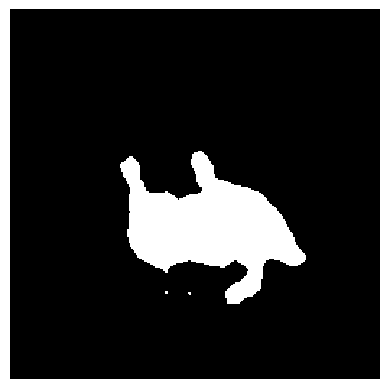} & \includegraphics[width=1.8cm]{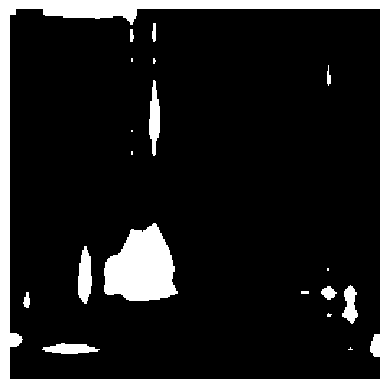} 
			& 	\includegraphics[width=1.8cm]{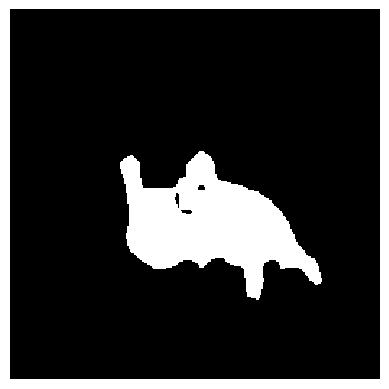}  &

			\includegraphics[width=1.8cm, height=1.8cm]{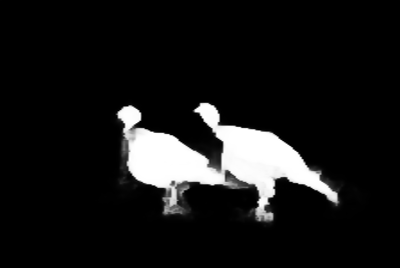} & 		\includegraphics[width=1.8cm, height=1.8cm]{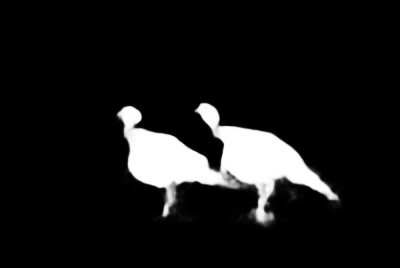}
			&\includegraphics[width=1.8cm, height=1.8cm]{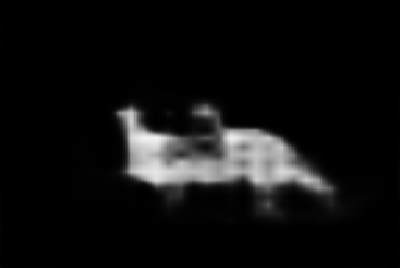} & 		\includegraphics[width=1.8cm, height=1.8cm]{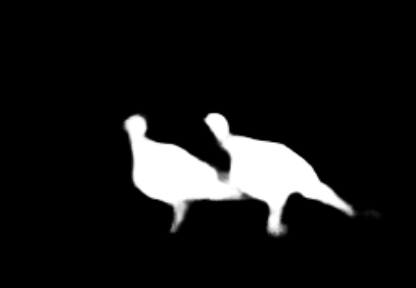} & 	\includegraphics[width=1.8cm, height=1.8cm]{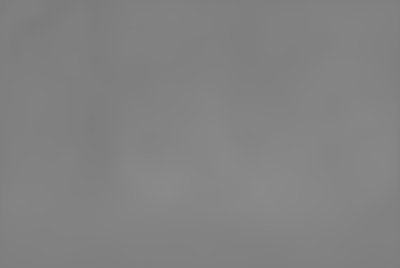}(PoolNet)\\
			\midrule

			\includegraphics[width=1.8cm]{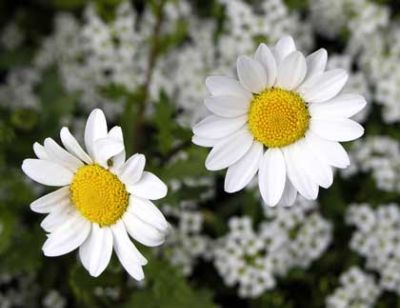} & \includegraphics[width=1.8cm]{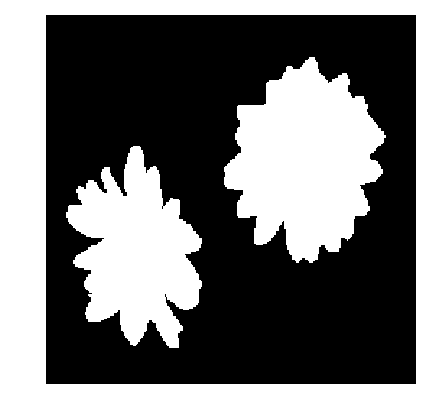}   & 		\includegraphics[width=1.8cm]{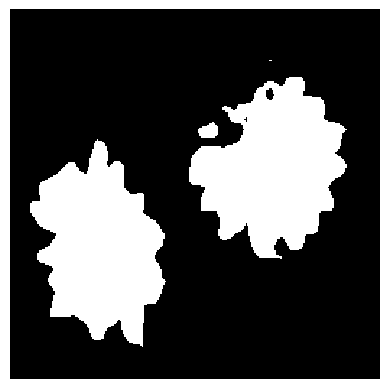} & 		\includegraphics[width=1.8cm]{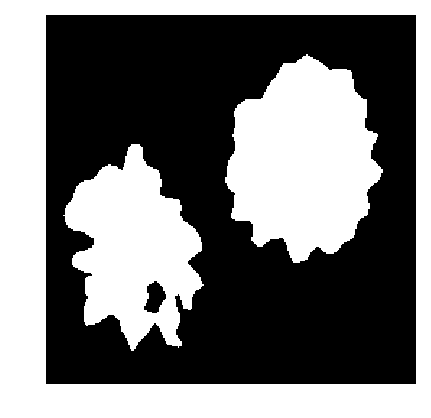} & 	\includegraphics[width=1.8cm]{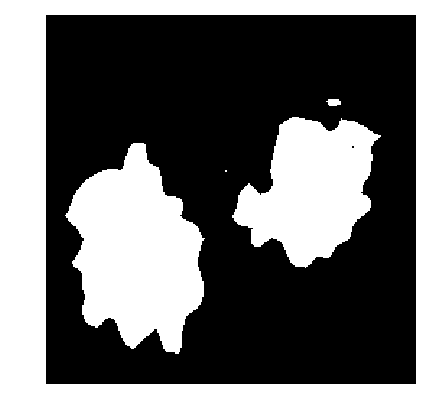}&
			\includegraphics[width=1.8cm]{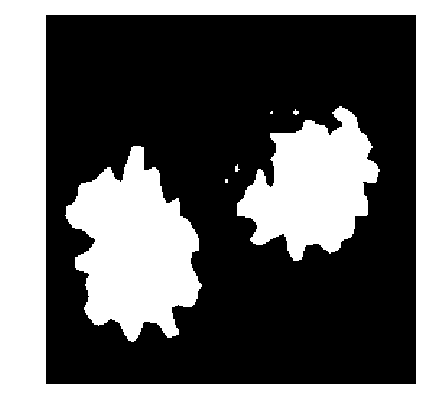} &
			\includegraphics[width=1.8cm]{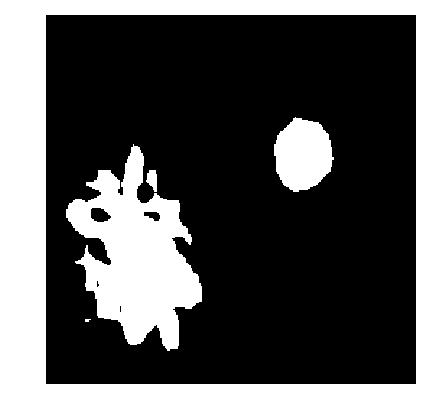}& 
			\includegraphics[width=1.8cm]{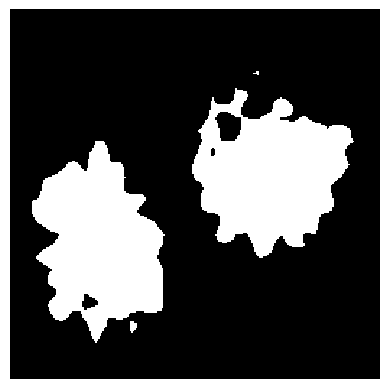} & \includegraphics[width=1.8cm]{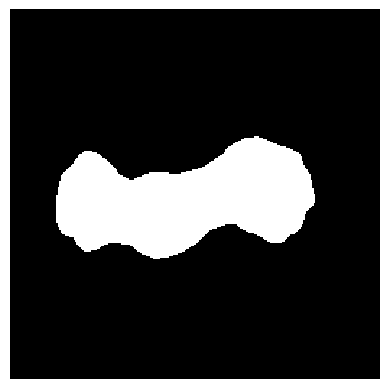} 
			& \includegraphics[width=1.8cm]{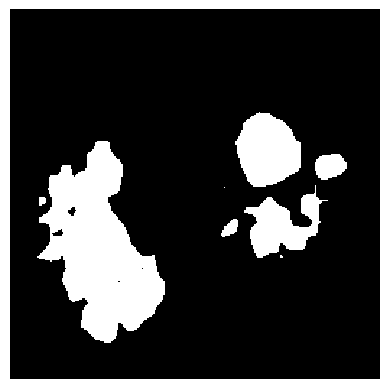}	&

			\includegraphics[width=1.8cm, height=1.8cm]{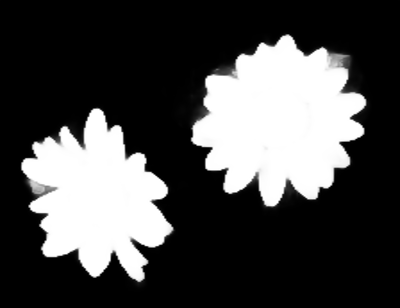} & 		\includegraphics[width=1.8cm, height=1.8cm]{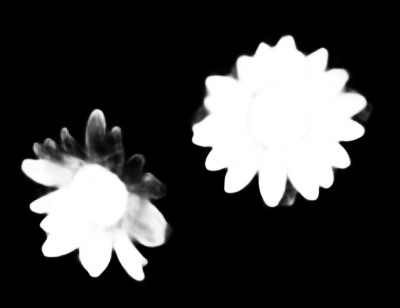}
			&\includegraphics[width=1.8cm, height=1.8cm]{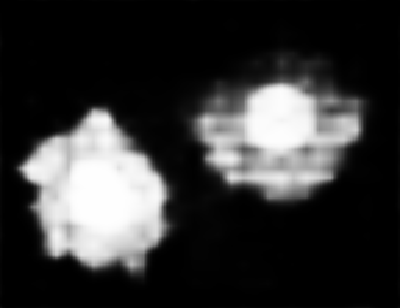} & 		\includegraphics[width=1.8cm, height=1.8cm]{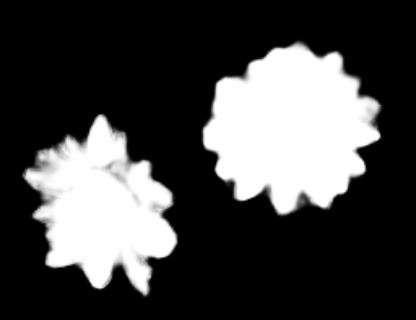} & 	\includegraphics[width=1.8cm, height=1.8cm]{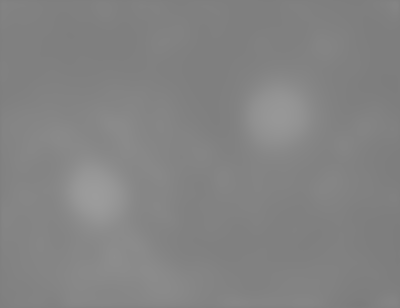}(PoolNet) \\
			\midrule

			\includegraphics[width=1.8cm, height=1.8cm]{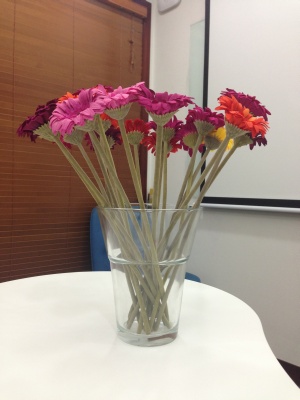} & \includegraphics[width=1.8cm]{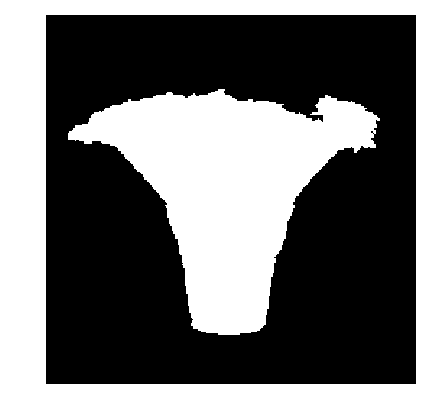}   & 		\includegraphics[width=1.8cm]{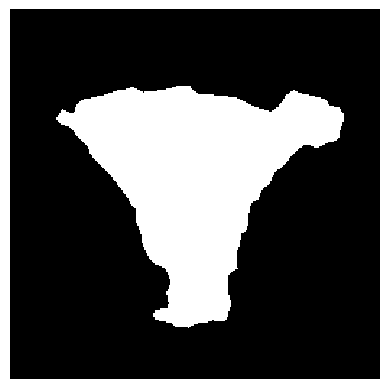} & 	\includegraphics[width=1.8cm]{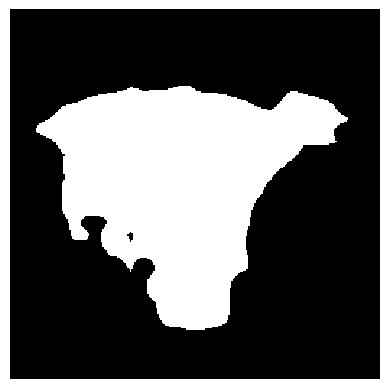} & 	
			\includegraphics[width=1.8cm]{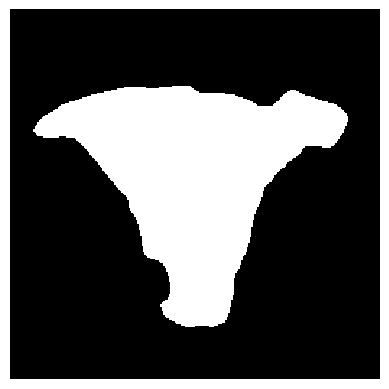}&

			\includegraphics[width=1.8cm]{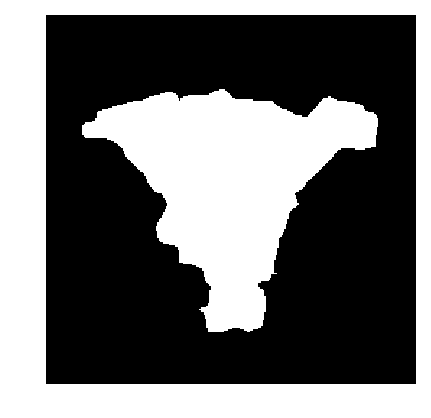} &
			\includegraphics[width=1.8cm]{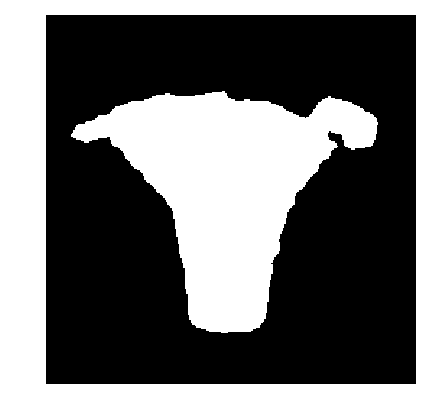}

			& 	
			\includegraphics[width=1.8cm]{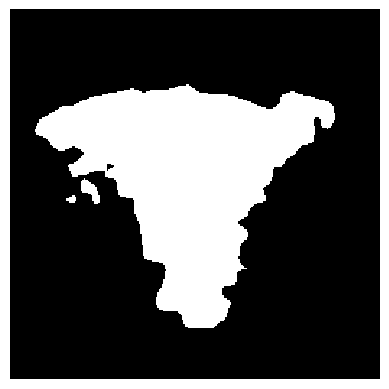} & \includegraphics[width=1.8cm]{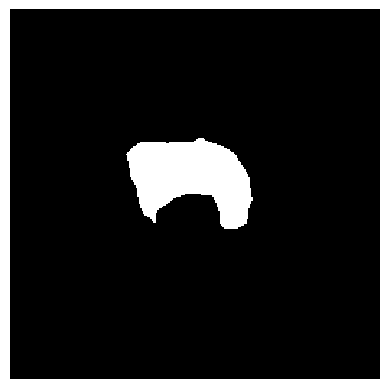} 
			& \includegraphics[width=1.8cm]{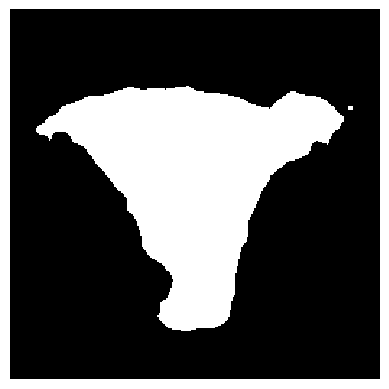} & 
			\includegraphics[width=1.8cm, height=1.8cm]{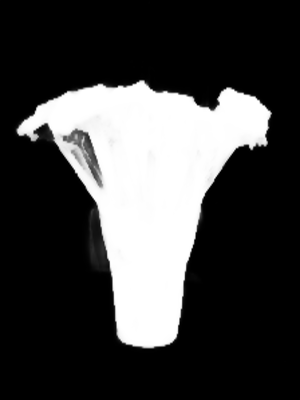} & 		\includegraphics[width=1.8cm, height=1.8cm]{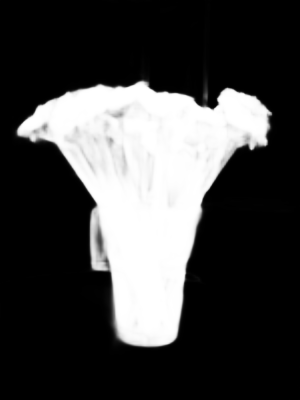}
			&\includegraphics[width=1.8cm, height=1.8cm]{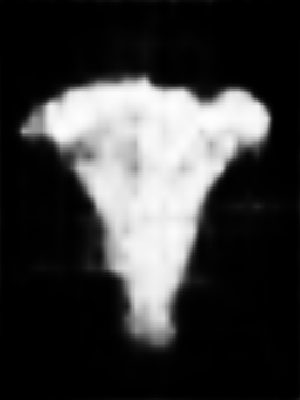} & 		\includegraphics[width=1.8cm, height=1.8cm]{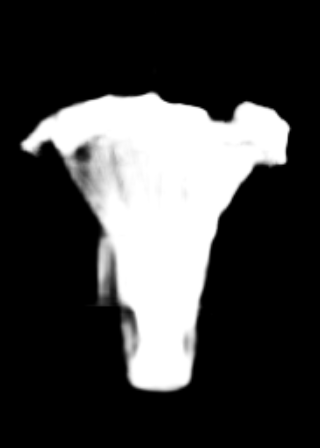} & \includegraphics[width=1.8cm, height=1.8cm]{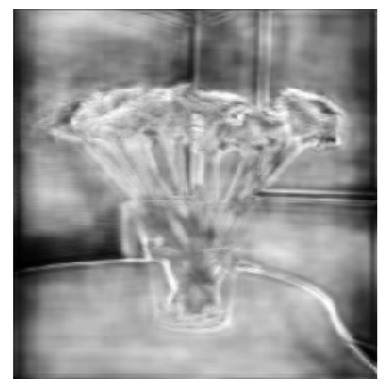}(PFAN)\\
			\midrule

			\includegraphics[width=1.8cm]{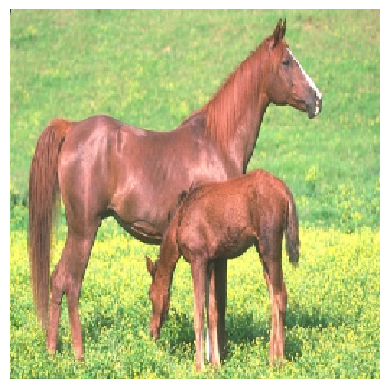} & \includegraphics[width=1.8cm]{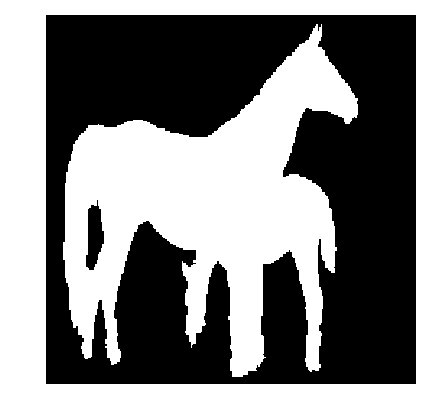}   & 		\includegraphics[width=1.8cm]{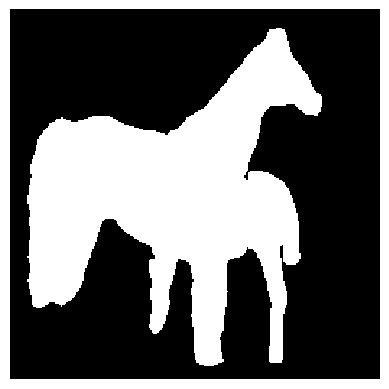} & 		\includegraphics[width=1.8cm]{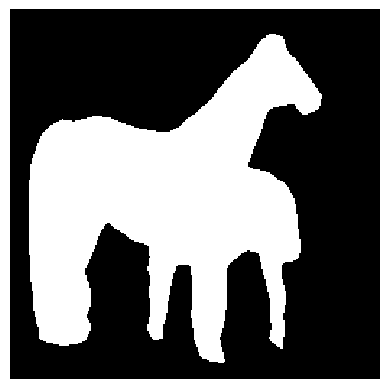} & 	\includegraphics[width=1.8cm]{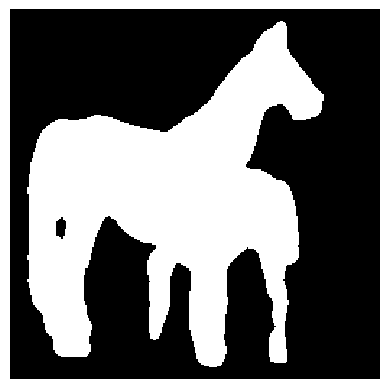}	&
			\includegraphics[width=1.8cm]{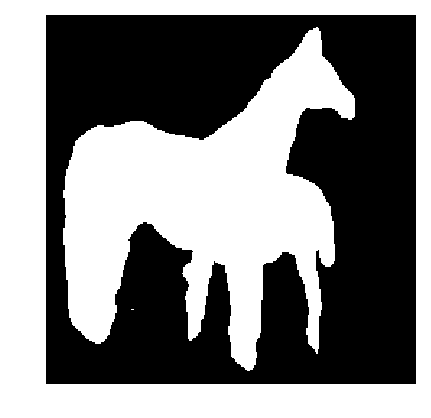} &
			\includegraphics[width=1.8cm]{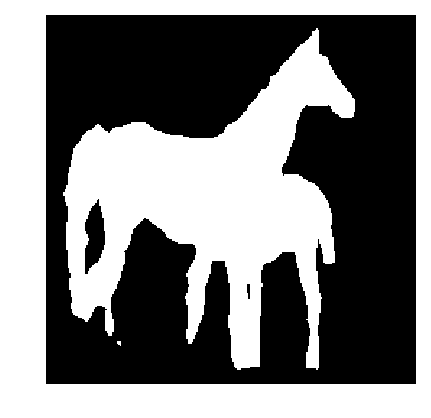}  &
			
			\includegraphics[width=1.8cm]{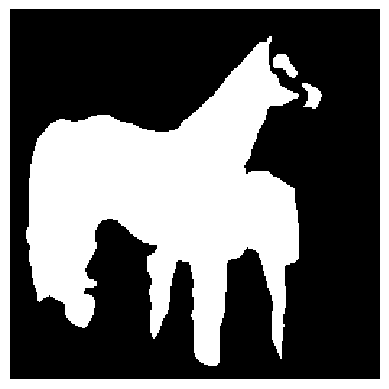} & \includegraphics[width=1.8cm]{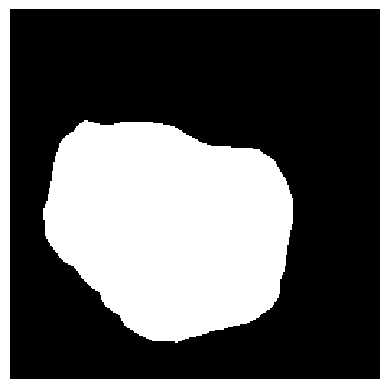} 
			& \includegraphics[width=1.8cm]{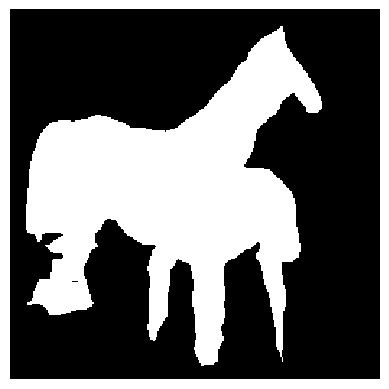} 	 &

			\includegraphics[width=1.8cm, height=1.8cm]{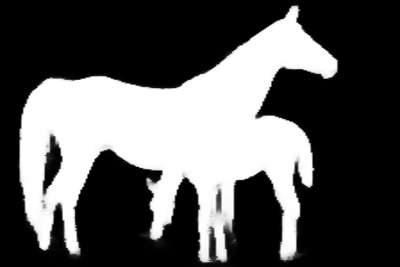} & 		\includegraphics[width=1.8cm, height=1.8cm]{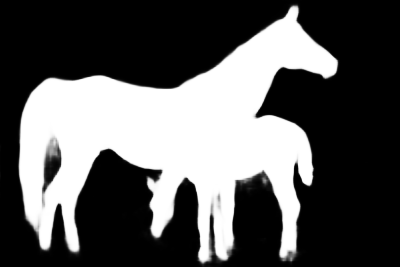}
			&\includegraphics[width=1.8cm, height=1.8cm]{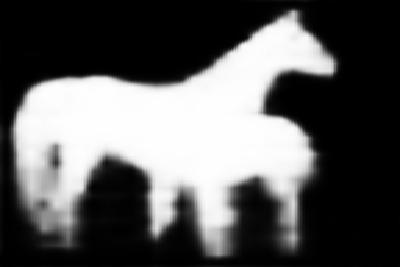} & 		\includegraphics[width=1.8cm, height=1.8cm]{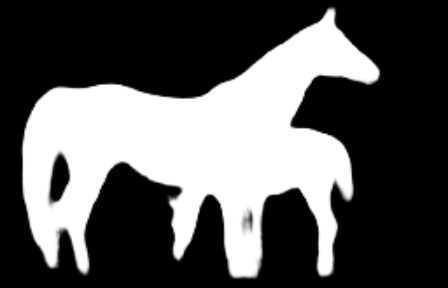} & 	\includegraphics[width=1.8cm, height=1.8cm]{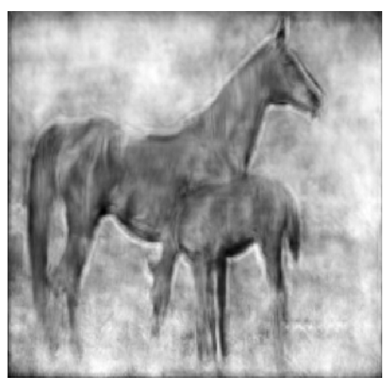} (PFAN)\\
			\midrule

			\includegraphics[width=1.8cm]{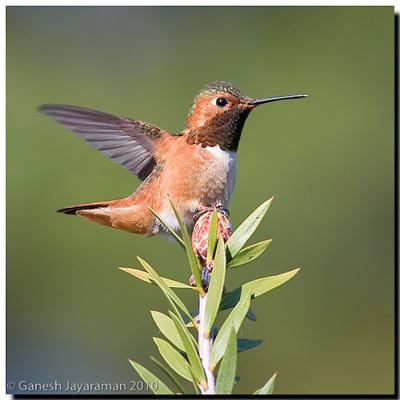} & \includegraphics[width=1.8cm]{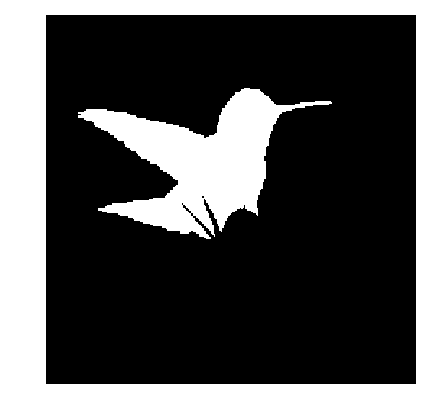}   &

			\includegraphics[width=1.8cm]{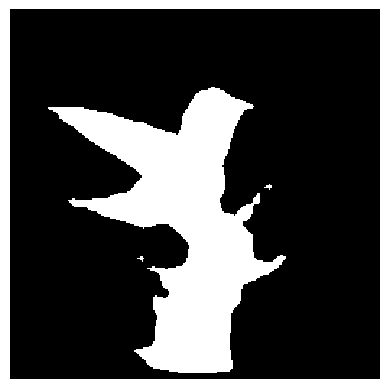} & 		\includegraphics[width=1.8cm]{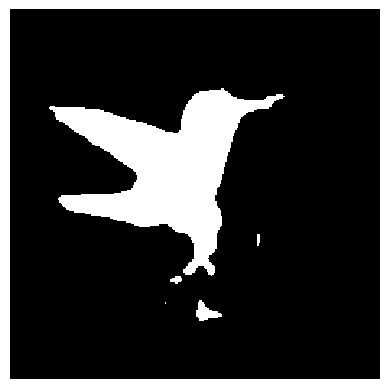} & 			
			\includegraphics[width=1.8cm]{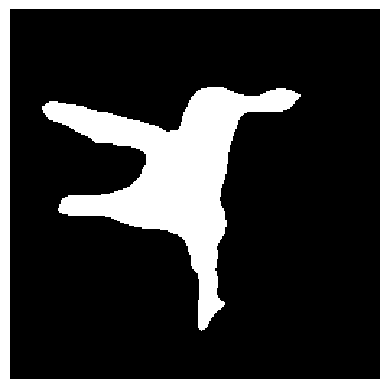}	&
			\includegraphics[width=1.8cm]{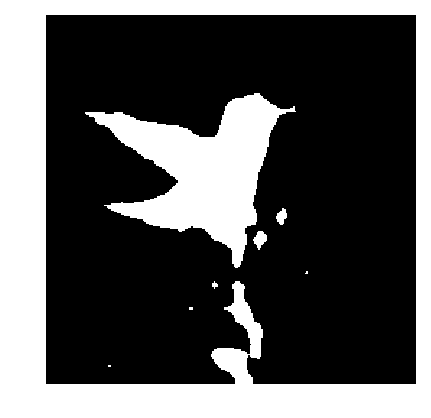} &
			\includegraphics[width=1.8cm]{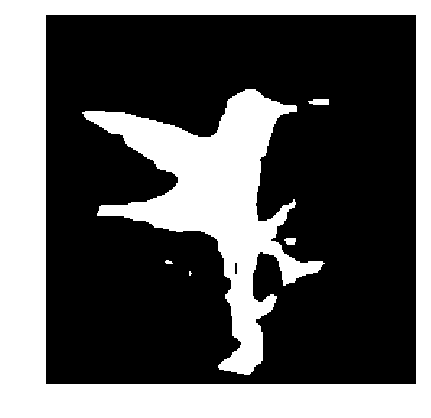} & 
			
			\includegraphics[width=1.8cm]{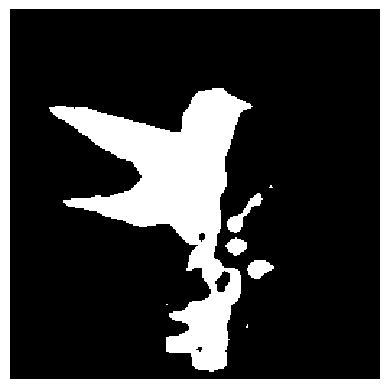} & \includegraphics[width=1.8cm]{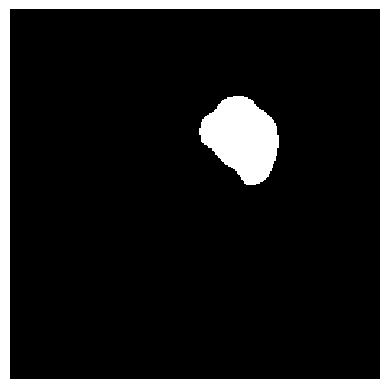} 
			& \includegraphics[width=1.8cm]{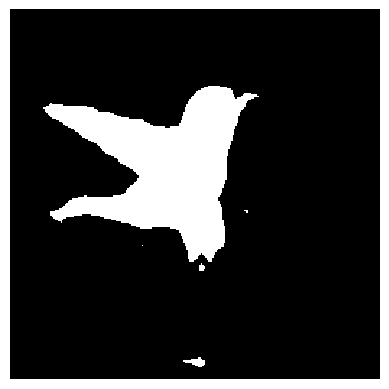}	&

			\includegraphics[width=1.8cm]{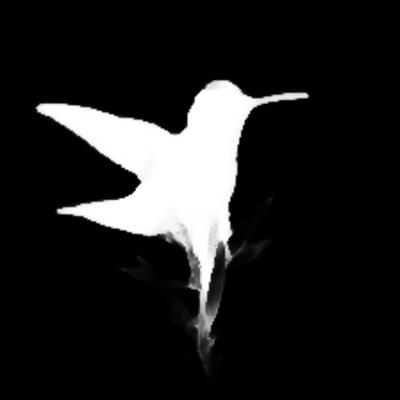} & 		\includegraphics[width=1.8cm]{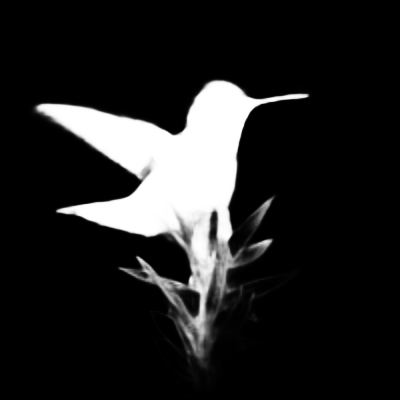}
			&\includegraphics[width=1.8cm]{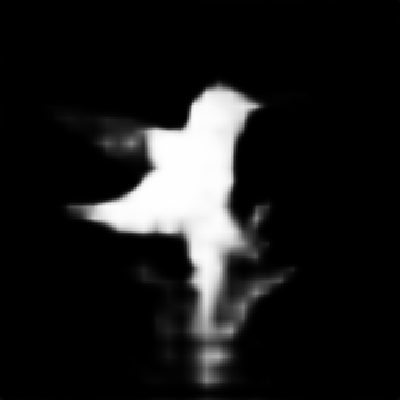} & 		\includegraphics[width=1.8cm]{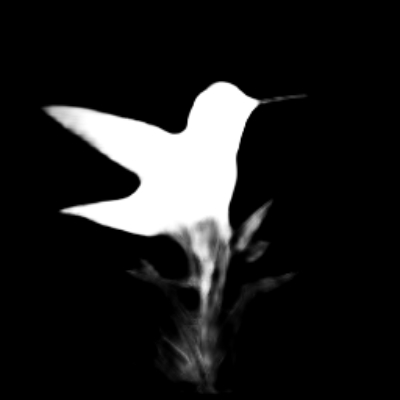} & \includegraphics[width=1.8cm]{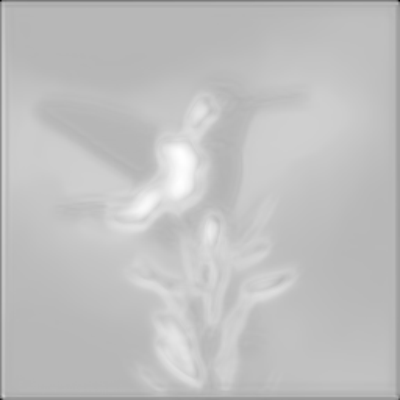}(BasNet)\\
			\midrule

			\includegraphics[width=1.8cm]{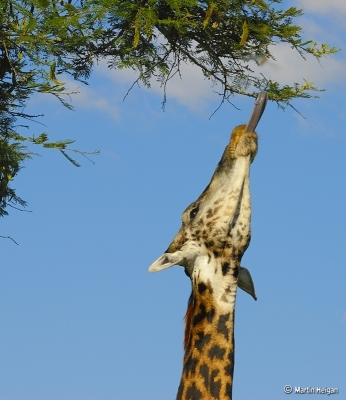} & \includegraphics[width=1.8cm]{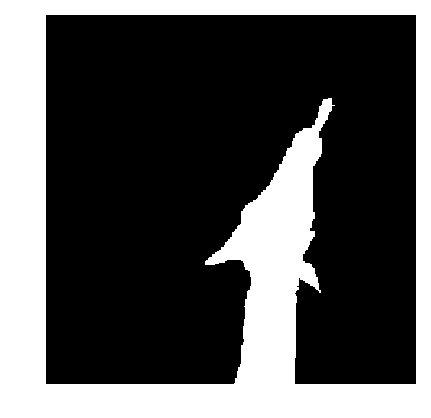}   & 		\includegraphics[width=1.8cm]{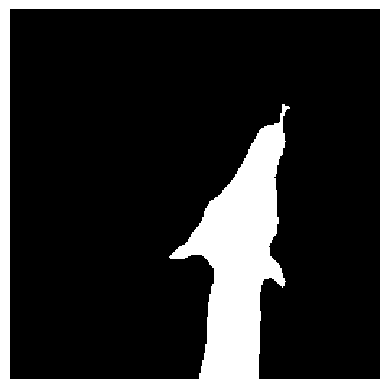} & 		\includegraphics[width=1.8cm]{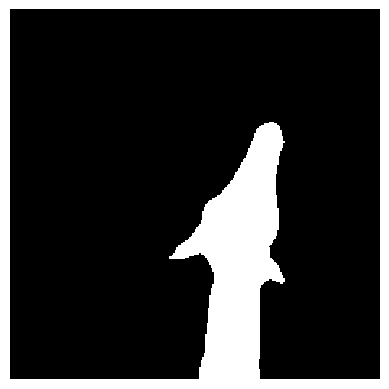} &
			\includegraphics[width=1.8cm]{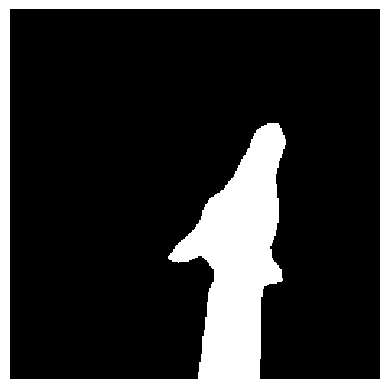} 	&
			\includegraphics[width=1.8cm]{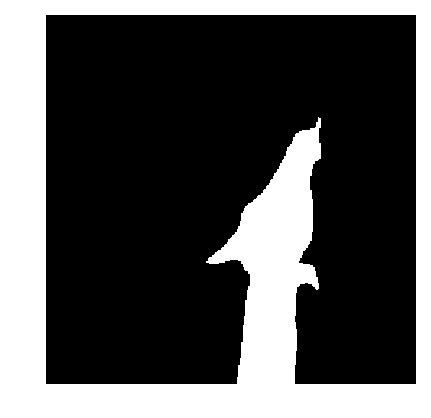} &
			\includegraphics[width=1.8cm]{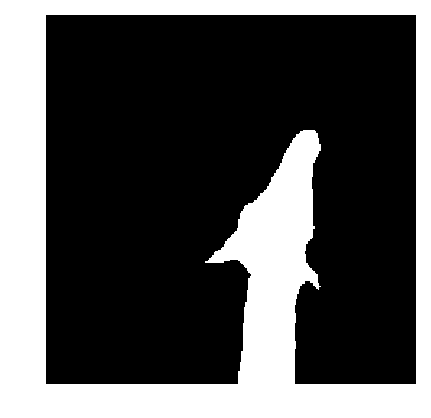} & 
			\includegraphics[width=1.8cm]{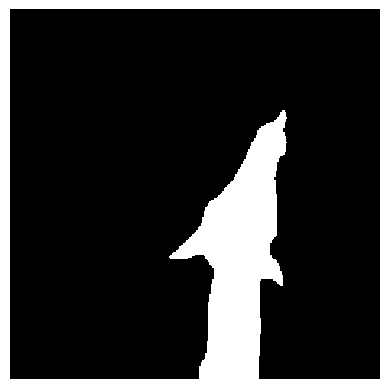} & \includegraphics[width=1.8cm]{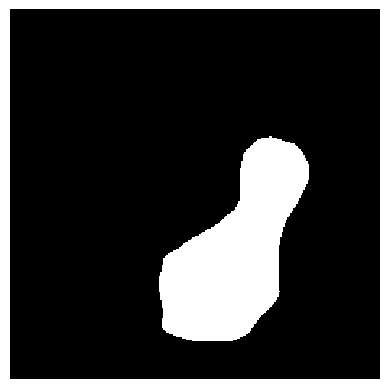} 
			& \includegraphics[width=1.8cm]{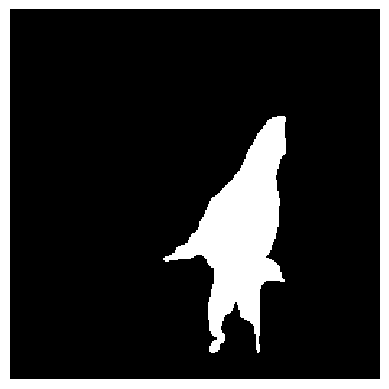} 	&

			\includegraphics[width=1.8cm, height=1.8cm]{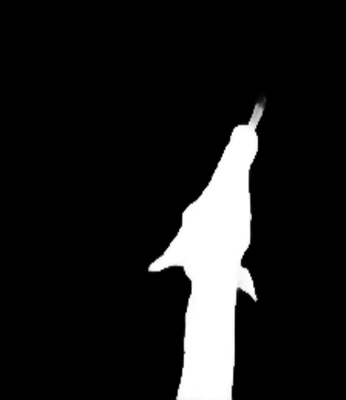} & 		\includegraphics[width=1.8cm, height=1.8cm]{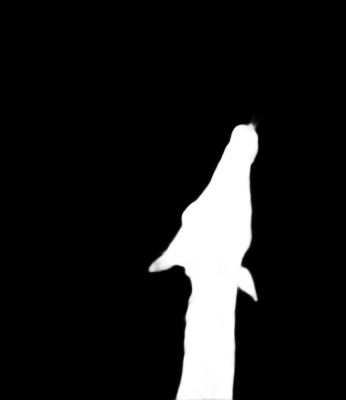}
			&\includegraphics[width=1.8cm, height=1.8cm]{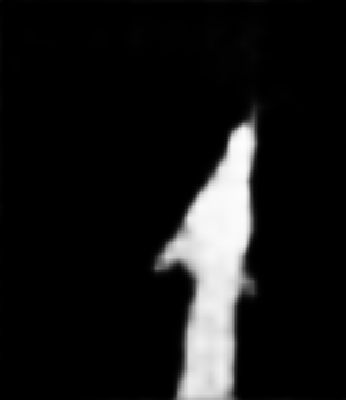} & 		\includegraphics[width=1.8cm, height=1.8cm]{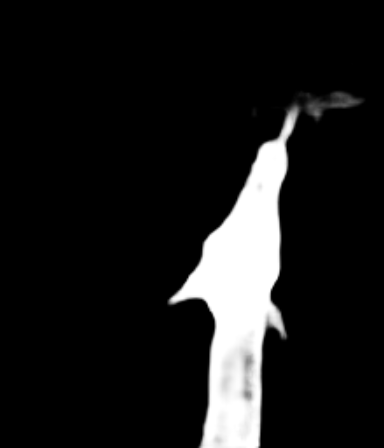}  & 	\includegraphics[width=1.8cm, height=1.8cm]{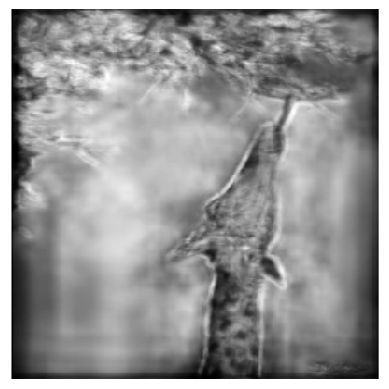} (PFAN)\\
			
			\bottomrule
		\end{tabular}
	\end{adjustbox}
	\caption{Visual Results for High-fidelity and Low-fidelity Data Training, and Comparison with State-of-the-art methods }
	\label{tbl:sup_results}

\end{figure*}

\vspace*{5em}

\begin{table}[h!]
\scriptsize
	\begin{tabular}{ c c c  | c c c }
		
		\toprule
		\multicolumn{3}{c}{ResNet-CAS} & \multicolumn{3}{c}{ResNet-CE} \\
		\hline
	
		\% of low-fidelity data & $F_\beta \uparrow$ & MAE $\downarrow$ & \% of low-fidelity data & $F_\beta \uparrow$ & MAE $\downarrow$    \\
		\hline  \\ 
		2 & 0.\textbf{903 }& 0.055  & 2 & \textbf{0.903} & \textbf{0.051}\\  

		5 &\textbf{ 0.921}  & \textbf{0.046}  &  5 & 0.901 & 0.055  \\
10 & \textbf{0.904 }& \textbf{0.053 }& 10 & 0.858 & 0.068     \\
30 &  \textbf{0.906} & \textbf{0.520}    & 30 & 0.647 & 0.159 \\
50 & \textbf{0.920} & \textbf{0.038} & 50 & 0.691 & 0.140

	\end{tabular}
\caption{Results of models trained on different percentages of low-fidelity MSRA-B} 
\label{table:lfd_perc}
\vspace{5em}
\end{table}

\begin{table*}[h!]
	\centering
	\scriptsize
	\begin{tabular}{ccc}
		\includegraphics[scale=0.2]{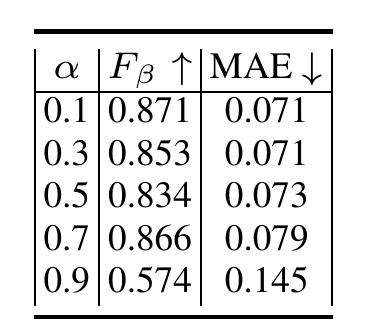}  &		  \includegraphics[scale=0.22]{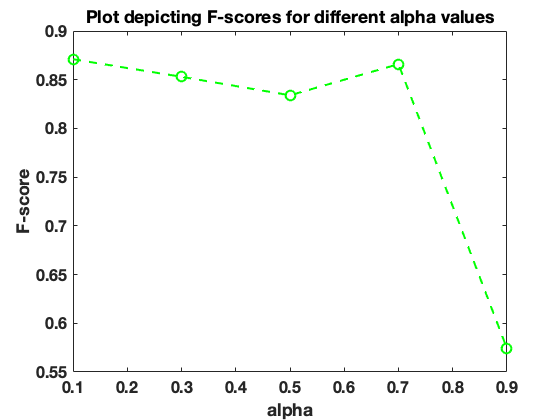} 
		& \includegraphics[scale=0.22]{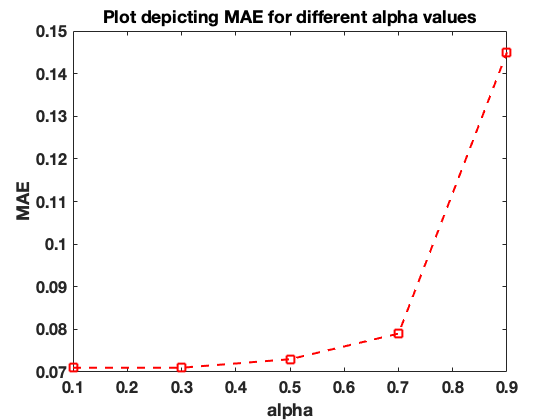}

	\end{tabular}
	\caption{Results different values of $\alpha$ trained and tested on DUTS dataset with FCN-ResNet101 }
	\label{table:alpha}
\end{table*}

\begin{table}[h!]
	\centering
	{\scriptsize

		\begin{tabular}{p{4cm}|p{1cm}p{1cm}|p{1cm}p{1cm}p{1cm}p{1cm}p{1cm}p{1cm}p{1cm}}
			& \multicolumn{2}{c|}{Contour} & \multicolumn{6}{c}{Region metrics} \\ \hline
			&  \multicolumn{2}{c|}{F-meas.}  & \multicolumn{2}{c}{GT-cov.} & \multicolumn{2}{c}{Rand.~Index} & \multicolumn{2}{c}{Var.~Info.} \\ 
			& ODS  & OIS  & ODS & OIS &  ODS & OIS & ODS & OIS \\ \hline\hline \\
			FCN-ResNet101-a-CE  &  0.04 &  0.04 & 0.79 &0.79 & 0.55 &  0.55 & 1.39 &  1.39 \\
			FCN-ResNet101-a-CAS &  {\bf 0.48} & {\bf 0.48} & {\bf 0.87} &{\bf 0.87} & {\bf 0.87} &  {\bf 0.87} & {\bf 0.50} & {\bf 0.50 }\\
			FCN-ResNet101-t-CAS&  0.17&  0.17 & 0.83 &0.83 & 0.66 &  0.66 & 0.94 &  0.94 \\
			DeepLab-d-CE   & 0.18 & 0.18 & 0.82 & 0.82 & 0.67 & 0.67 & 1.35 & 1.35 \\
			DeepLab-d-CAS & 0.07 & 0.07 & 0.73 & 0.73 & 0.54 & 0.54 & 1.64 & 1.64 
		\end{tabular}
	}
	\begin{tablenotes}
		\item[] \scriptsize -a- denotes trained on the 7 saliency datasets; -t- denotes trained on texture data; \\ -d- denotes trained on DUTS-TR data

	\end{tablenotes}
	\caption{{\bf Results on Texture Segmentation
			Datasets of Deep Networks}. Algorithms are evaluated using contour and region
		metrics. Higher F-measure for the contour
		metric, ground truth covering (GT-cov), and rand index indicate
		better fit to the ground truth, and lower variation of information
		(Var.~Info) indicates a better fit to ground truth. }
	
	\label{tab:results_Nets}

\end{table}

\vspace{5em}

\def\fWidRTex{.0635\linewidth}
\def\fHeiRTex{.04\linewidth}
\begin{figure*}[h!]
	\centering
	\scriptsize
	\begin{tabular}{c}
		Images  \\
		\includegraphics[width=\fWidRTex,height=\fHeiRTex]{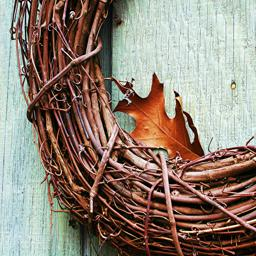}
		\includegraphics[width=\fWidRTex,height=\fHeiRTex]{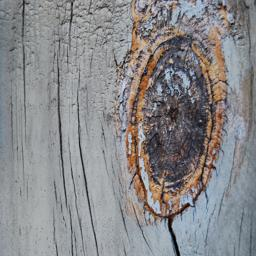}
		\includegraphics[width=\fWidRTex,height=\fHeiRTex]{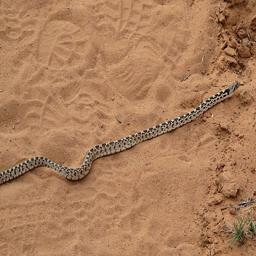}
		\includegraphics[width=\fWidRTex,height=\fHeiRTex]{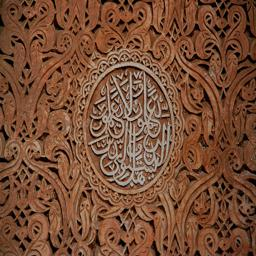}
		\includegraphics[width=\fWidRTex,height=\fHeiRTex]{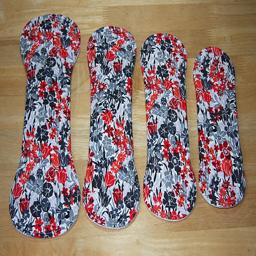}
		\includegraphics[width=\fWidRTex,height=\fHeiRTex]{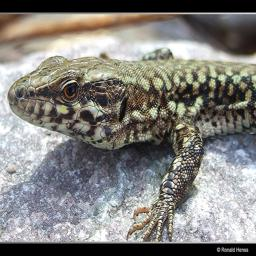}
		\includegraphics[width=\fWidRTex,height=\fHeiRTex]{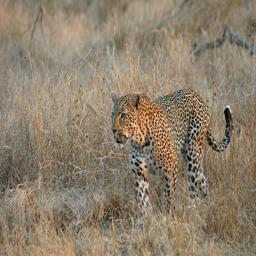}
		\includegraphics[width=\fWidRTex,height=\fHeiRTex]{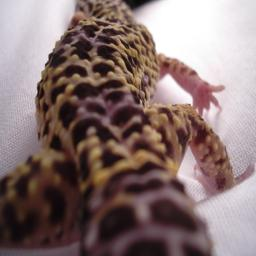}
		\includegraphics[width=\fWidRTex,height=\fHeiRTex]{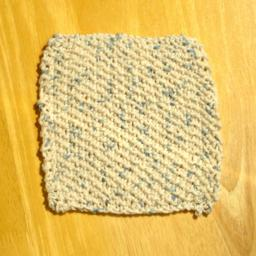}
		\includegraphics[width=\fWidRTex,height=\fHeiRTex]{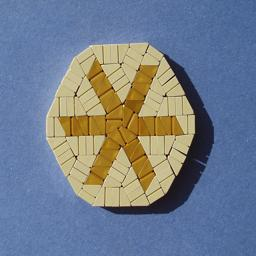}
		\includegraphics[width=\fWidRTex,height=\fHeiRTex]{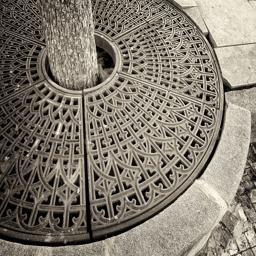}
		\includegraphics[width=\fWidRTex,height=\fHeiRTex]{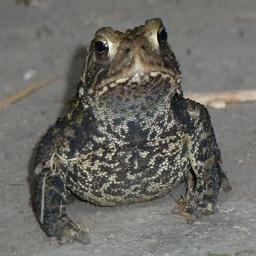}
		\includegraphics[width=\fWidRTex,height=\fHeiRTex]{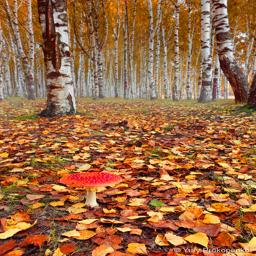}
		\includegraphics[width=\fWidRTex,height=\fHeiRTex]{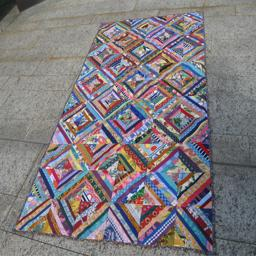}  \\
		Ground Truth  \\
		\includegraphics[width=\fWidRTex,height=\fHeiRTex]{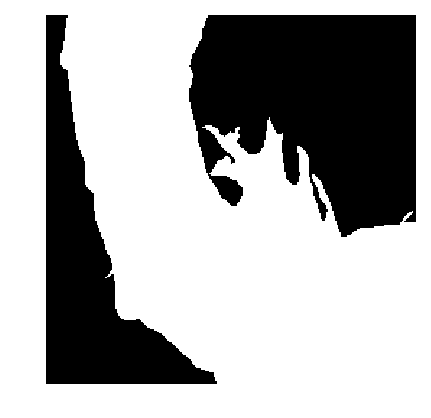}
		\includegraphics[width=\fWidRTex,height=\fHeiRTex]{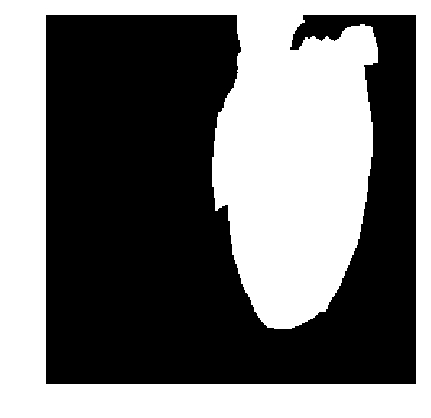}
		\includegraphics[width=\fWidRTex,height=\fHeiRTex]{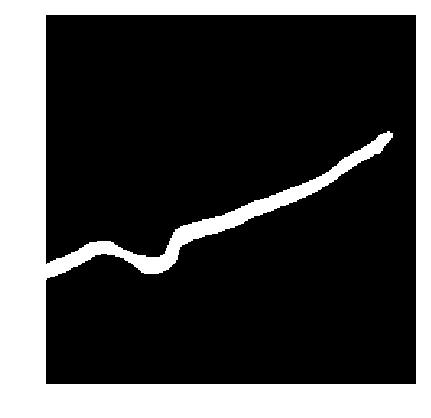}
		\includegraphics[width=\fWidRTex,height=\fHeiRTex]{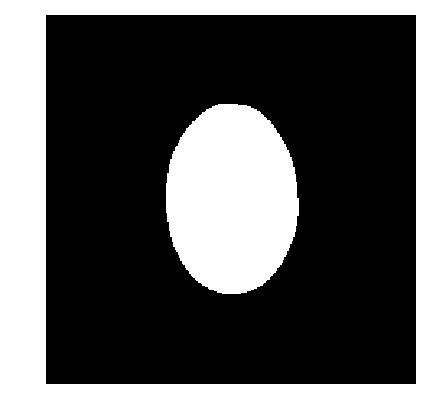}
		\includegraphics[width=\fWidRTex,height=\fHeiRTex]{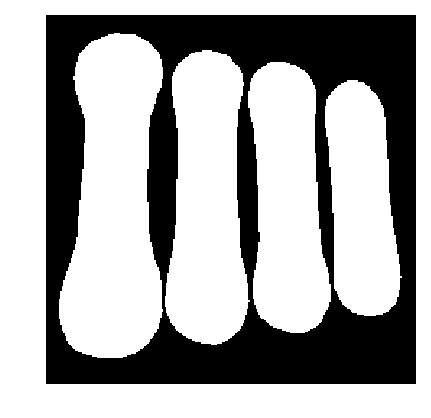}
		\includegraphics[width=\fWidRTex,height=\fHeiRTex]{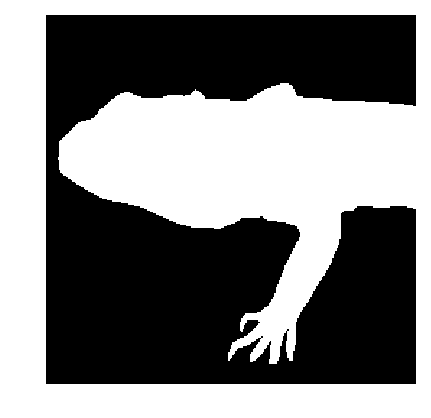}
		\includegraphics[width=\fWidRTex,height=\fHeiRTex]{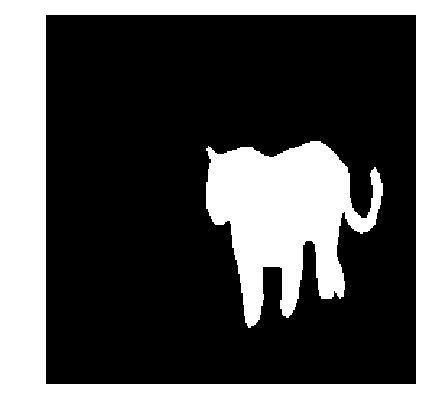} 
		\includegraphics[width=\fWidRTex,height=\fHeiRTex]{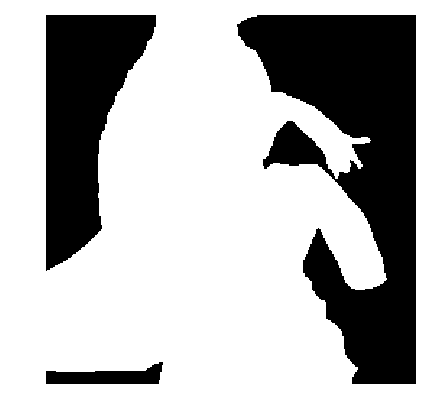}
		\includegraphics[width=\fWidRTex,height=\fHeiRTex]{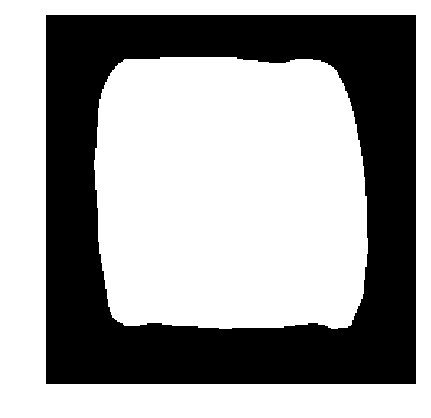}
		\includegraphics[width=\fWidRTex,height=\fHeiRTex]{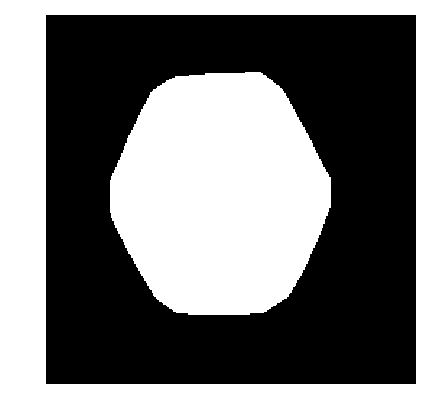}
		\includegraphics[width=\fWidRTex,height=\fHeiRTex]{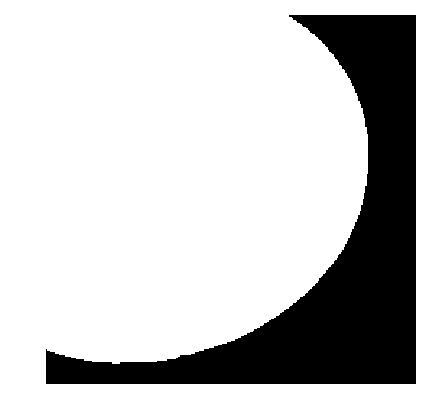}
		\includegraphics[width=\fWidRTex,height=\fHeiRTex]{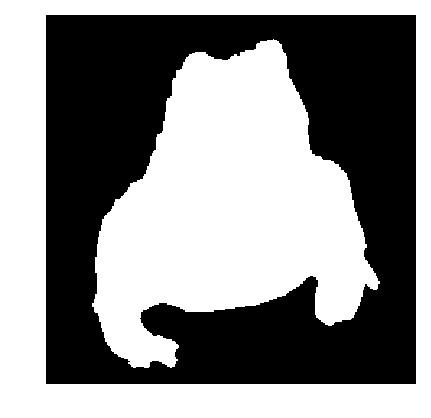}
		\includegraphics[width=\fWidRTex,height=\fHeiRTex]{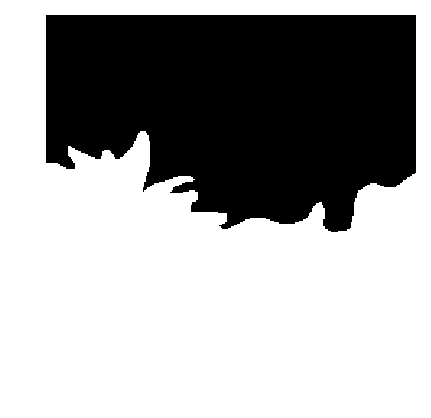}
		\includegraphics[width=\fWidRTex,height=\fHeiRTex]{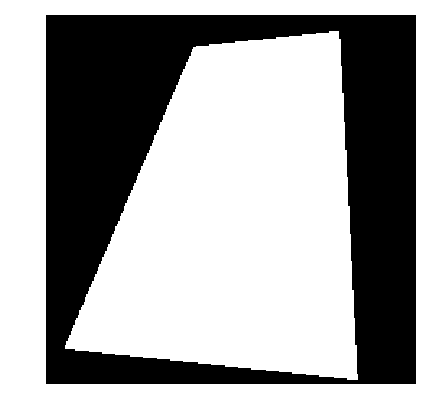}\\
		FCN-ResNet101-a-CE \\
		\includegraphics[width=\fWidRTex,height=\fHeiRTex]{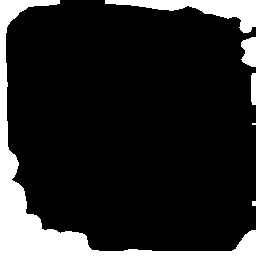}
		\includegraphics[width=\fWidRTex,height=\fHeiRTex]{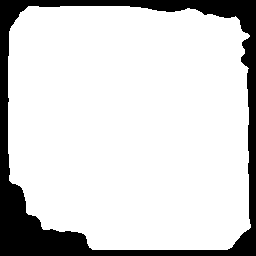}
		\includegraphics[width=\fWidRTex,height=\fHeiRTex]{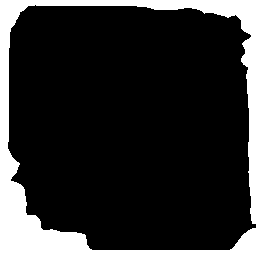}
		\includegraphics[width=\fWidRTex,height=\fHeiRTex]{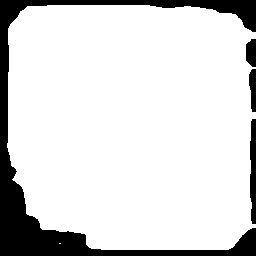}
		\includegraphics[width=\fWidRTex,height=\fHeiRTex]{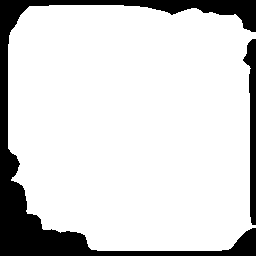}
		\includegraphics[width=\fWidRTex,height=\fHeiRTex]{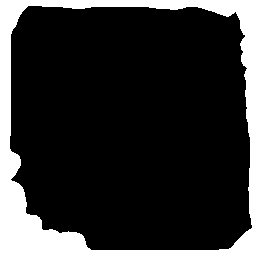}
		\includegraphics[width=\fWidRTex,height=\fHeiRTex]{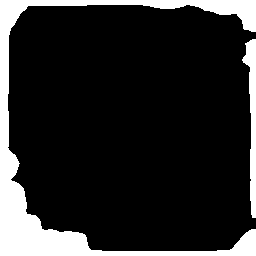} 
		\includegraphics[width=\fWidRTex,height=\fHeiRTex]{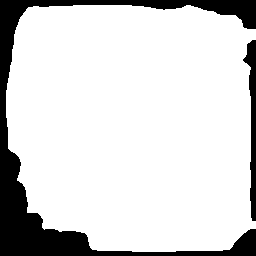}
		\includegraphics[width=\fWidRTex,height=\fHeiRTex]{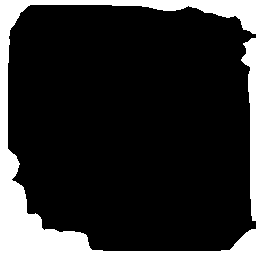}
		\includegraphics[width=\fWidRTex,height=\fHeiRTex]{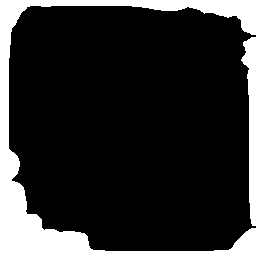}
		\includegraphics[width=\fWidRTex,height=\fHeiRTex]{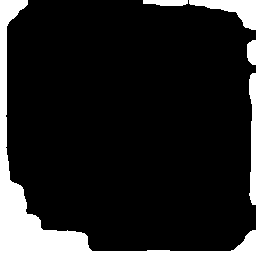}
		\includegraphics[width=\fWidRTex,height=\fHeiRTex]{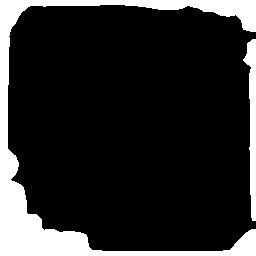}
		\includegraphics[width=\fWidRTex,height=\fHeiRTex]{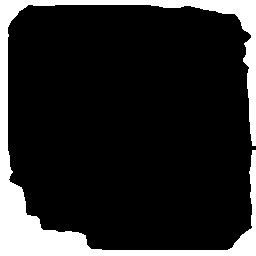}
		\includegraphics[width=\fWidRTex,height=\fHeiRTex]{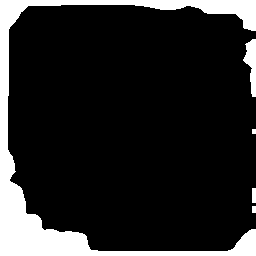}\\ 
		
		FCN-ResNet101-a-CAS \\
		\includegraphics[width=\fWidRTex,height=\fHeiRTex]{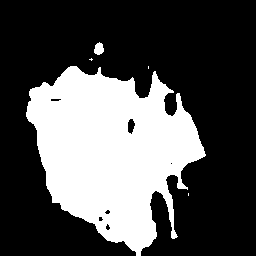}
		\includegraphics[width=\fWidRTex,height=\fHeiRTex]{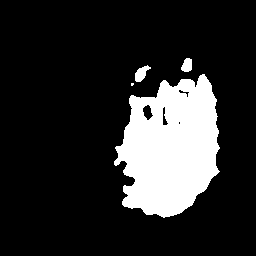}
		\includegraphics[width=\fWidRTex,height=\fHeiRTex]{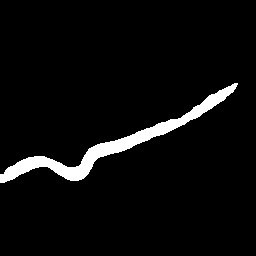}
		\includegraphics[width=\fWidRTex,height=\fHeiRTex]{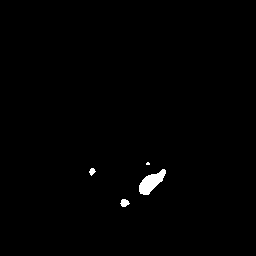}
		\includegraphics[width=\fWidRTex,height=\fHeiRTex]{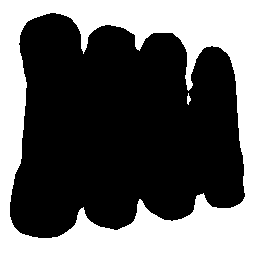}
		\includegraphics[width=\fWidRTex,height=\fHeiRTex]{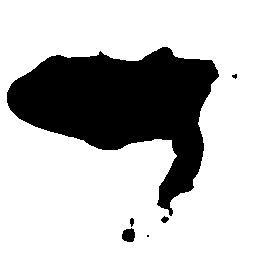}
		\includegraphics[width=\fWidRTex,height=\fHeiRTex]{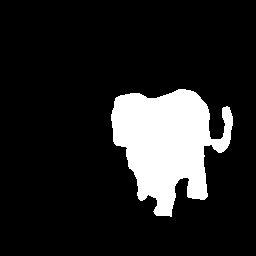} 
		\includegraphics[width=\fWidRTex,height=\fHeiRTex]{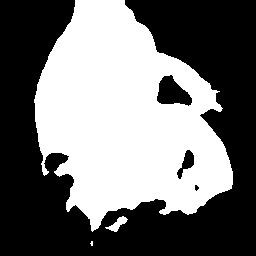}
		\includegraphics[width=\fWidRTex,height=\fHeiRTex]{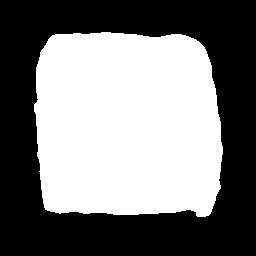}
		\includegraphics[width=\fWidRTex,height=\fHeiRTex]{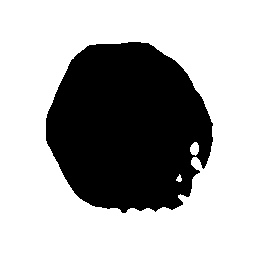}
		\includegraphics[width=\fWidRTex,height=\fHeiRTex]{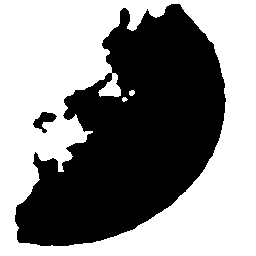}
		\includegraphics[width=\fWidRTex,height=\fHeiRTex]{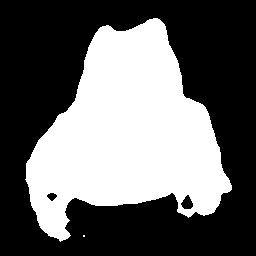}
		\includegraphics[width=\fWidRTex,height=\fHeiRTex]{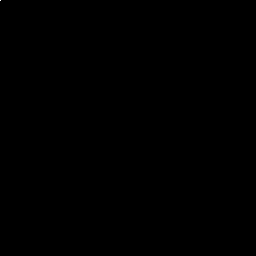}
		\includegraphics[width=\fWidRTex,height=\fHeiRTex]{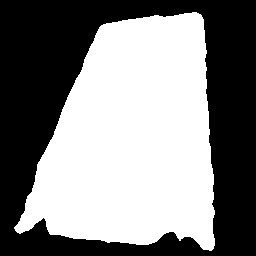}

	\end{tabular}
	\caption{{\bf Sample representative results on Real-World
			Texture Dataset}: Visual results for texture segmentation experiments; -a- denotes trained on the 7 saliency datasets }
	\label{fig:deep_results}
\end{figure*}

\vspace*{5em}
\section{Conclusion}

In this work we made several noteworthy contributions. We  presented class-agnostic segmentation loss function which allows us to cast the problem of region-based general segmentation problem with deep networks. We tested on 7 salient object segmentation datasets against 15 methods and on a challenging general segmentation dataset in 2 data fidelity settings.  Using the class-agnostic segmentation loss function we tackled the problem of salient object segmentation in low-fidelity training data case (where class labels are either not available or not reliable) and showed state-of-the-art results, around 50\% better than the next best methods. This huge performance gain is due to the fact that CAS loss forces learning of descriptors through the deep network which are invariant on similar looking regions. We also applied CAS loss to high-fidelity training data case as well as texture segmentation and showed state-of-the-art results. This shows that our CAS loss performs well in multiple scenarios for general segmentation across domains. Although standard CE based loss functions perform satisfactorily in salient object detection with high-fidelity training data, they fail completely in low-fidelity training data case with a performance drop of around 50\%. Likewise the CE based loss functions fail to learn any significant feature in texture segmentation dataset.

\clearpage
\bibliography{library.bib}
\bibliographystyle{unsrt}
\newpage

\begin{center}
	\textbf{\Large{Appendix}
	}
\end{center}

\appendix

\section{Gradient Calculation for CAS loss} \label{a_cal}

Here we describe the calculation of gradient of the class-agnostic segmentation loss function with respect to the weights $\omega$ of the network: \\

The class-agnostic segmentation loss is defined as:
\begin{align} 
	\textit{CAS} = \sum_{i = 1}^{N} \int_{r_i}^{} \underbrace{\dfrac{\alpha ||\mathbf{s}(x)-\mathbf{\hat{s}}(r_i)||_{2}^{2}}{|r_i|}dx}_\text{Uniformer} \notag \\ - \sum_{i=1}^{N}\sum_{\substack{j=1 \\ i\neq j}}^{N} (1-\alpha)\underbrace{||\mathbf{\hat{s}}(r_i) - \mathbf{\hat{s}}(r_j)||_{2}^{2}}_\text{Discriminator}
\end{align}
where $N$ is the number of regions in the ground truth mask;     $r_1,..., r_i, r_j, ..., r_N$ denotes the regions of the ground truth mask (a particular segment);    $|r_i|$ denotes the number of pixels in the region $r_i$; $\mathbf{s} = \{s^1, ..., s^m,..., s^M\}$ is  a vector of output descriptor components (or softmax output) of the network; $m \in \{1,...,M\}$ where $M$ denotes the number of output (softmax) channels i.e., number of units in the last layer of the network;   $\alpha \in [0, 1]$ is a scalar, a weighing hyper-parameter which assigns weight to each term; for a region $r$  we have that,  $\mathbf{\hat{s}}(r) = \{\hat{s}(r)^{1}, ...,\hat{s}^{m}(r),..., \hat{s}(r)^{M}\}$ is a vector containing channel-wise mean of the descriptor values; where for a channel $m$,  $\hat{s}^{m}(r)= \frac{1}{|r|}\int_{r} s^{m} (x)\ dx$. In our formulation $\hat{s}^{m}(r)$ acts as a proxy for class label. \\

We compute the derivative of the loss with respect to the weights $\omega$ of the neural network,
\begin{equation}
\nabla_{\omega} \sum_{i = 1}^{N} \int_{r_i} \alpha \frac{||\mathbf{s}(x)-\mathbf{\hat{s}}(r_i)||_{2}^{2}}{|r_i|}dx -  \nabla_{\omega} (1-\alpha)||\mathbf{\hat{s}}(r_i) - \mathbf{\hat{s}}(r_j)||_{2}^{2}
\end{equation}
using Leibniz rule, we can interchange the order of integral and gradient, we get,
\begin{equation}
\sum_{i = 1}^{N} \int_{r_i} \nabla_{\omega} \alpha \frac{||\mathbf{s}(x)-\mathbf{\hat{s}}(r_i)||_{2}^{2}}{|r_i|}dx -  \nabla_{\omega} (1-\alpha)||\mathbf{\hat{s}}(r_i) - \mathbf{\hat{s}}(r_j)||_{2}^{2}
\end{equation}

Next, we simply apply chain rule to get,

\begin{align} \label{casl_b1}
	& \nabla_{\boldsymbol{\omega}} CAS = \sum_{i=1}^{N} \int_{r_i}^{} 2\dfrac{\alpha  (\mathbf{s}(x)-\mathbf{\hat{s}}(r_i))(\nabla_{\boldsymbol{\omega}}\mathbf{s}(x) - \nabla_{\boldsymbol{\omega}}\mathbf{\hat{s}}(r_i))}{|r_i|}dx  \notag \\
	&-  \sum_{i=1}^{N}\sum_{\substack{j=1 \\ i\neq j}}^{N} 2(1-\alpha)(\mathbf{\hat{s}}(r_i) - \mathbf{\hat{s}}(r_j))(\nabla_{\boldsymbol{\omega}}\mathbf{\hat{s}}(r_i) - \nabla_{\boldsymbol{\omega}}\mathbf{\hat{s}}(r_j))  \notag
\end{align}
From Equation (1) we've, $
\nabla_{\boldsymbol{\omega}} \hat{s}^{m}(r_i) = \dfrac{1}{|r_i|}\int_{r_i}^{} \nabla_{\boldsymbol{\omega}} s^m(x) dx$,  where we have $\nabla_{\boldsymbol{\omega}} s^m(x)$ from the backpropagation of the network as $s^m(x)$ is the softmax output of the network.

\section{Detailed Results for Properties of CAS loss} \label{a_prop}
\subsection{Sparsity}
To prove the property of sparsity of the network when using CAS loss, we show the application of loss function in binary segmentation case where we have two outputs from the softmax channel. If we fix the value of $\alpha$, the problem simplifies to,

\begin{equation}
\text{min} -||{\bf a}-{\bf b}||_2^2 = \text{max} \quad||{\bf a}-{\bf b}||_2^2
\end{equation}
subject to,
\begin{equation}
\begin{split}
\sum_{j=1}^2{\bf a}(j)&=1 \qquad \sum_{j=1}^2{\bf b}(j)=1 \\
{\bf a}(j)&\geq 0 \qquad {\bf b}(j)\geq 0 \qquad \forall j 
\end{split}
\end{equation}

\begin{figure}[h!]
	\includegraphics[scale=0.35]{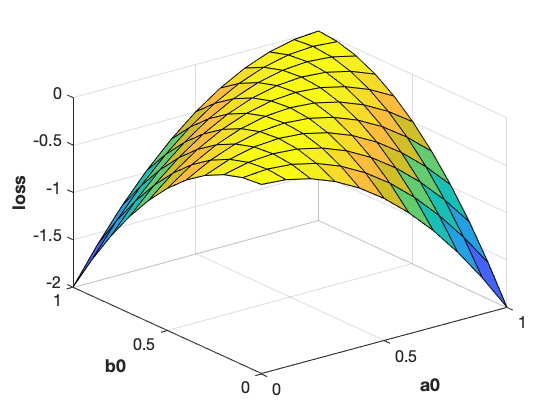}
	\caption{Profile of Loss Function}
	\label{fig:loss}
\end{figure}

Here, for simplicity of notation we are using $a$ and $b$ to denote the average descriptors on foreground and background respectively. We show the profile for loss function in Figure \ref{fig:loss}. Notice that the minima for loss function exists at the the points where output descriptors are sparse and different.\\

\noindent Writing the Lagrangian for the system we get,

\begin{align}
	L(a0,a1,b0,b1, \lambda1, \lambda2, \mu1, \mu2, \mu3, \mu4) &= (a0-b0)^2+(a1-b1)^2 -\lambda1 (a0+a1-1)\notag \\ 
	&-\lambda2 (b0+b1-1) + \mu1 a0 + \mu2 a1 + \mu2 b0 \notag\\ & + \mu4 b1
\end{align}

where $\lambda$ are the Lagrange multipliers for equality constraints and $\mu$ are the Lagrange multipliers for the inequality constraints.

\begin{equation}
\begin{aligned}
\nabla_{a0} L=2(a0-b0)-\lambda1 +\mu1=0\\
\nabla_{a1} L=2(a1-b1)-\lambda1 +\mu2=0\\
\nabla_{b0} L=-2(a0-b0)-\lambda2 +\mu3=0\\
\nabla_{b1} L=-2(a1-b1)-\lambda2 +\mu4=0\\
a0+a1=1\\
b0+b1=1\\
\mu1 \cdot a0=0\\
\mu2 \cdot a1=0\\
\mu3 \cdot b0=0\\
\mu4 \cdot b1=0
\end{aligned}
\end{equation}
For a point to be maximum of the above constraint optimization it has to satisfy the KKT conditions. Here we write the KKT conditions and show that the sparse and different descriptors satisfy the KKT conditions. Sparse and different descriptors $a0=1, a1=0, b0=0, b1=1$, $\lambda1=\lambda2=2, \mu2=\mu3=4$ satisfy the KKT conditions. Similarly, we can show the same for the other sparse and different solution. The same can be extended to multiple dimension with the only condition being that the number of softmax channels should be greater than or equal to the number of classes.

\begin{figure*}[h!]
	\centering
	\begin{adjustbox}{width=1\textwidth}
		\begin{tabular}{p{1.5cm}||cccccccc}
			\tiny Channel 0 & \includegraphics[width=0.8cm]{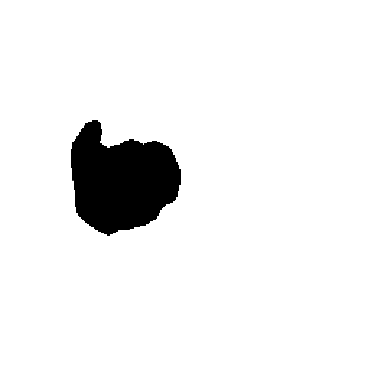} & 
			\includegraphics[width=0.8cm]{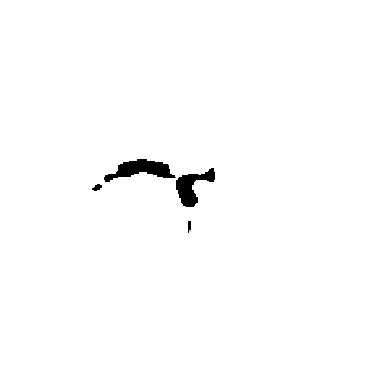} & \includegraphics[width=0.8cm]{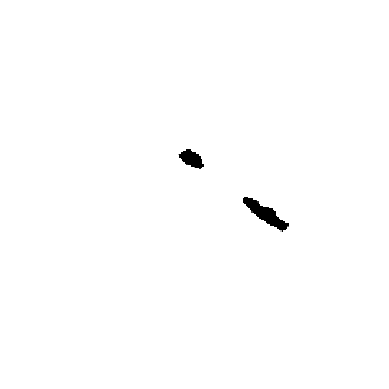} & \includegraphics[width=0.8cm]{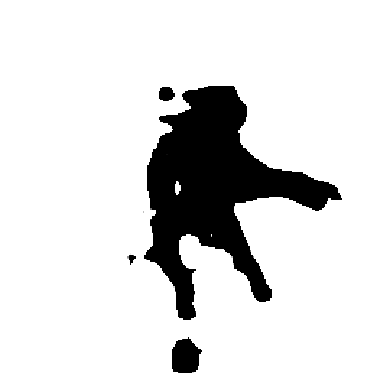} &
			\includegraphics[width=0.8cm]{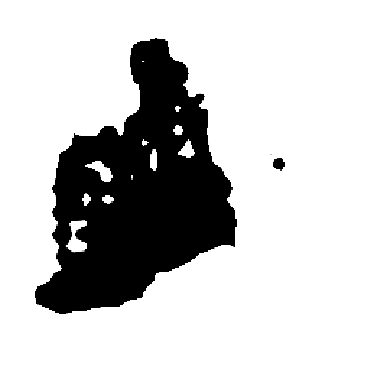} & \includegraphics[width=0.8cm]{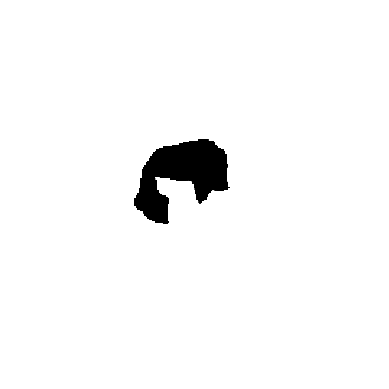} & \includegraphics[width=0.8cm]{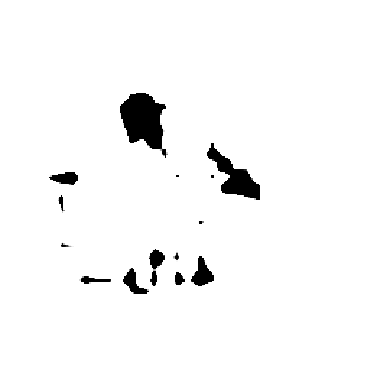} & \includegraphics[width=0.8cm]{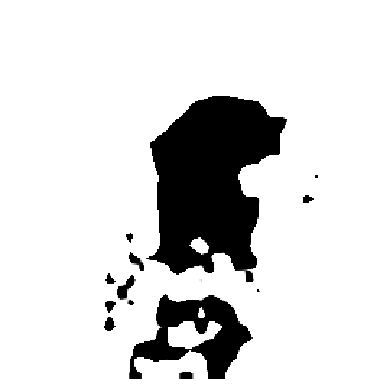} \\
			\tiny Channel 1 & \includegraphics[width=0.8cm]{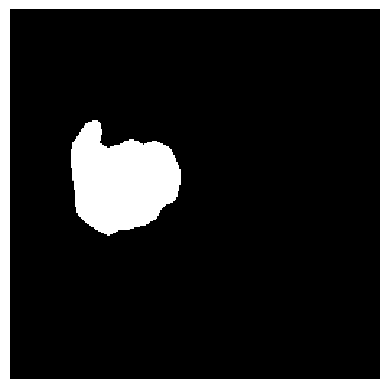} & 
			\includegraphics[width=0.8cm]{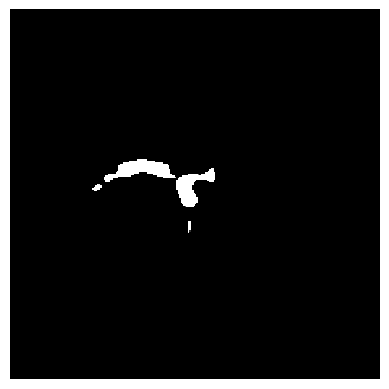} & \includegraphics[width=0.8cm]{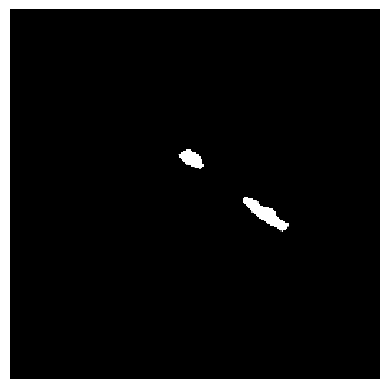} & \includegraphics[width=0.8cm]{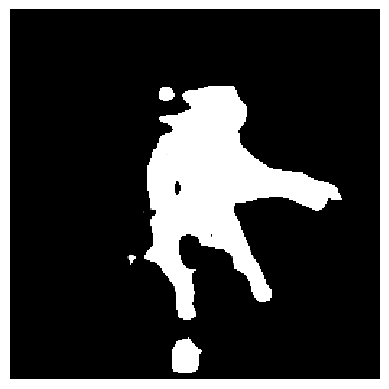} &
			\includegraphics[width=0.8cm]{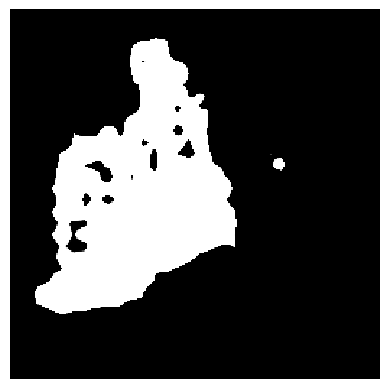} & \includegraphics[width=0.8cm]{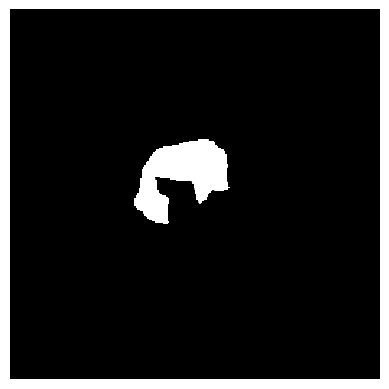} & \includegraphics[width=0.8cm]{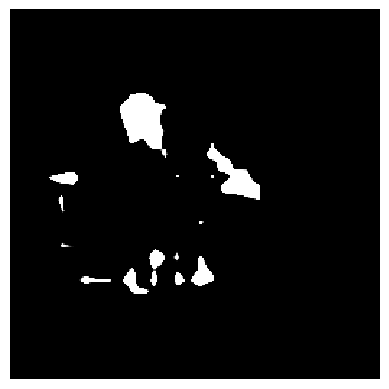} & \includegraphics[width=0.8cm]{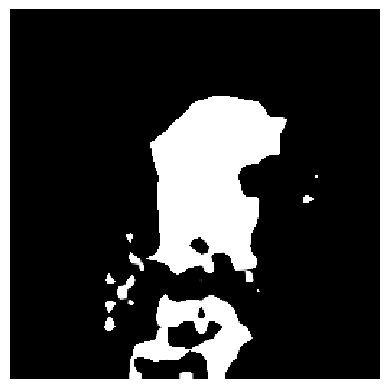} \\

			\hline
			\tiny Channel 0 & \includegraphics[width=0.8cm]{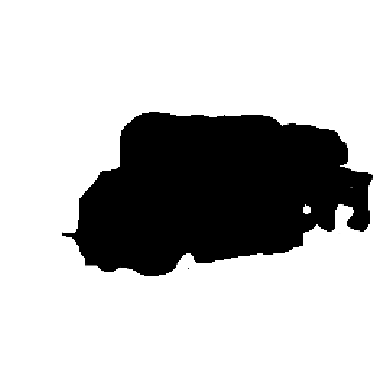} & 
			\includegraphics[width=0.8cm]{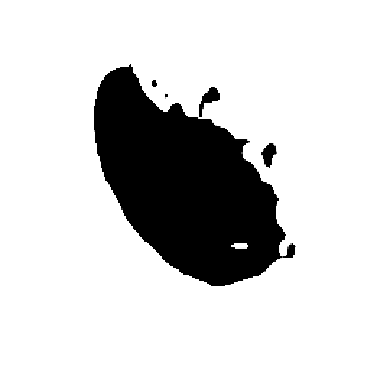} & \includegraphics[width=0.8cm]{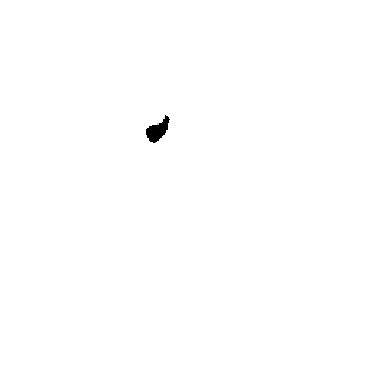} & \includegraphics[width=0.8cm]{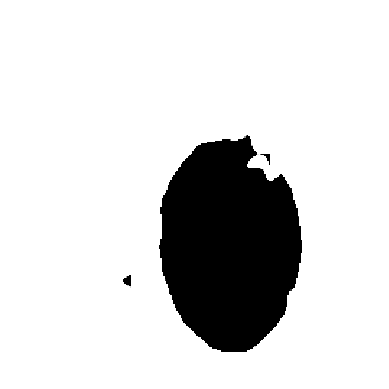} &
			\includegraphics[width=0.8cm]{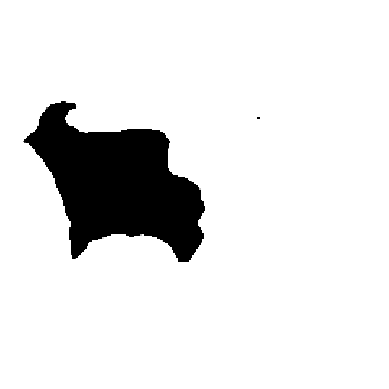} & \includegraphics[width=0.8cm]{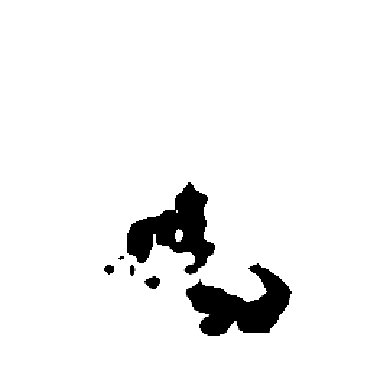} & \includegraphics[width=0.8cm]{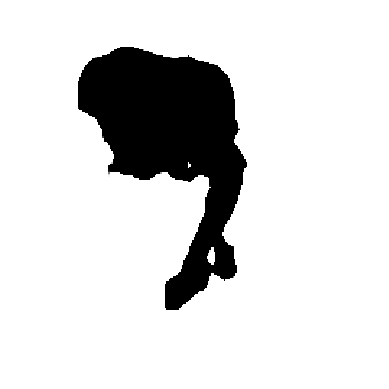} & \includegraphics[width=0.8cm]{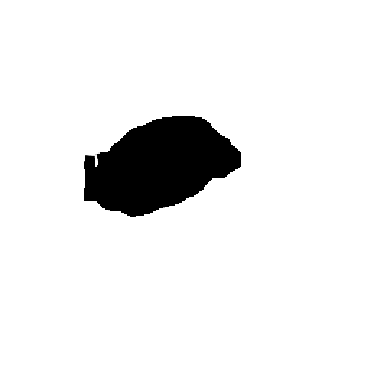} \\
			\tiny Channel 1 & \includegraphics[width=0.8cm]{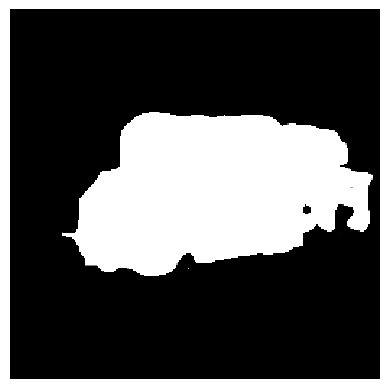} & 
			\includegraphics[width=0.8cm]{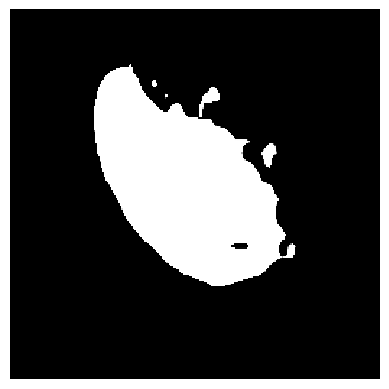} & \includegraphics[width=0.8cm]{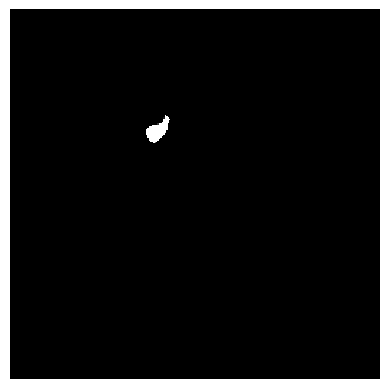} & \includegraphics[width=0.8cm]{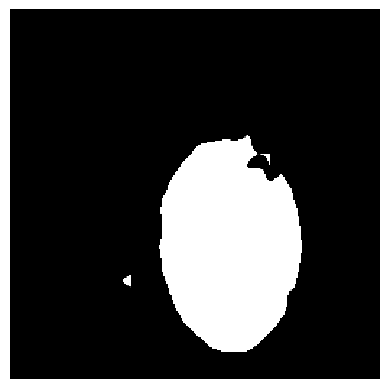} &
			\includegraphics[width=0.8cm]{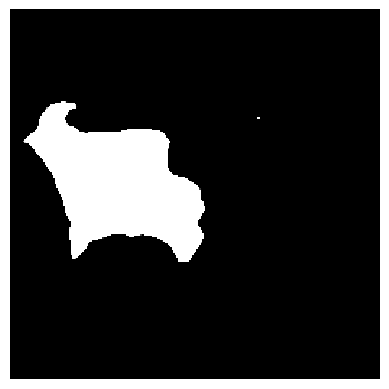} & \includegraphics[width=0.8cm]{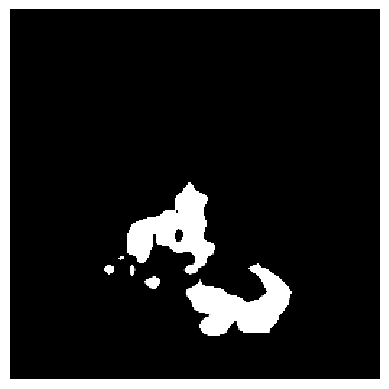} & \includegraphics[width=0.8cm]{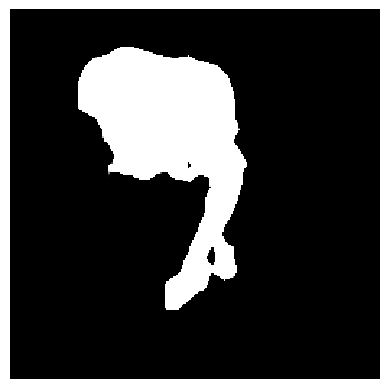} & \includegraphics[width=0.8cm]{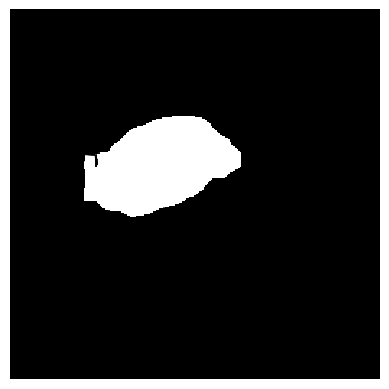} \\
			
			\hline
			
			\tiny Channel 0 & \includegraphics[width=0.8cm]{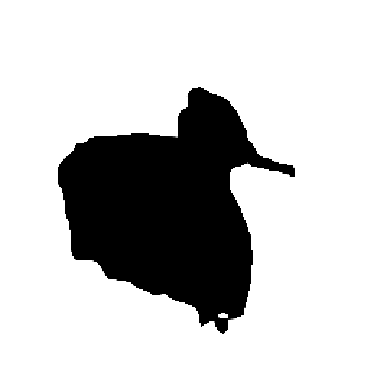} & 
			\includegraphics[width=0.8cm]{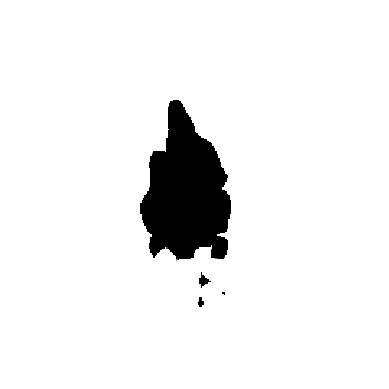} & \includegraphics[width=0.8cm]{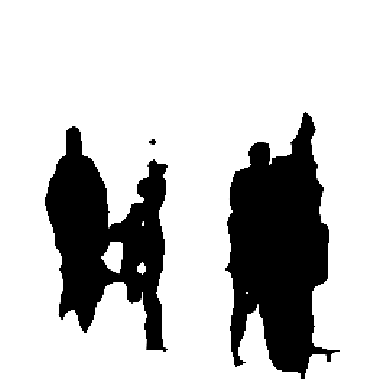} & \includegraphics[width=0.8cm]{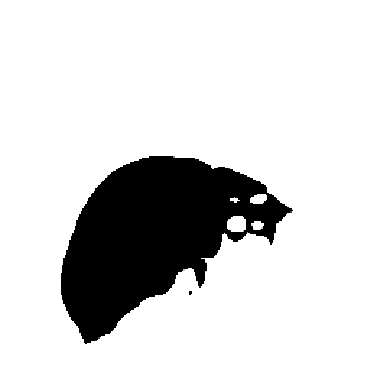} &
			\includegraphics[width=0.8cm]{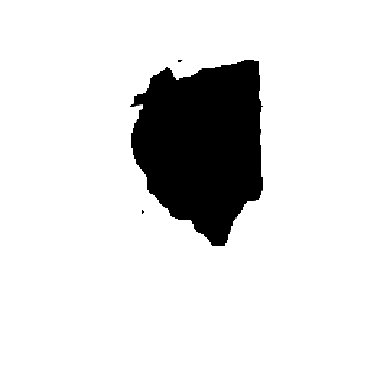} & \includegraphics[width=0.8cm]{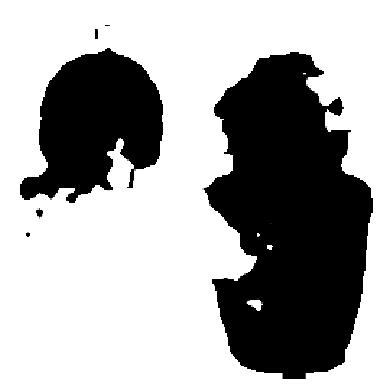} & \includegraphics[width=0.8cm]{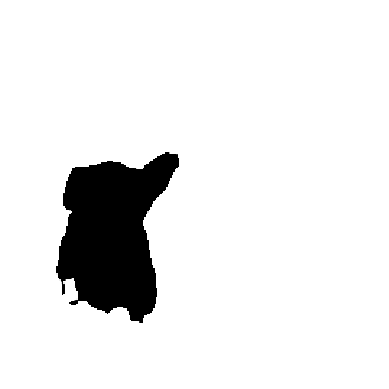} & \includegraphics[width=0.8cm]{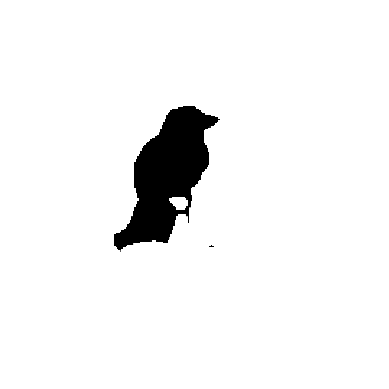} \\
			\tiny Channel 1 & \includegraphics[width=0.8cm]{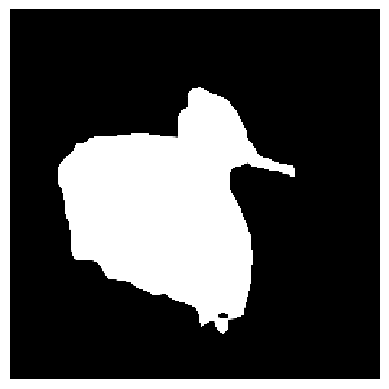} & 
			\includegraphics[width=0.8cm]{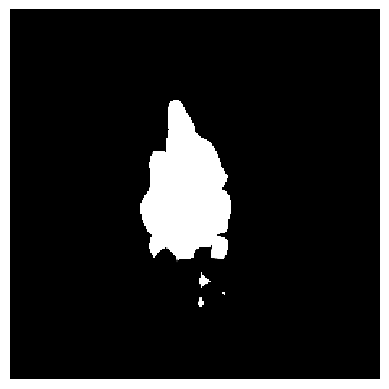} & \includegraphics[width=0.8cm]{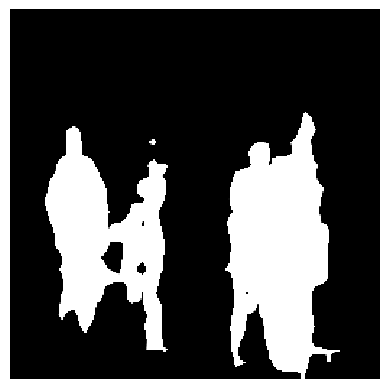} & \includegraphics[width=0.8cm]{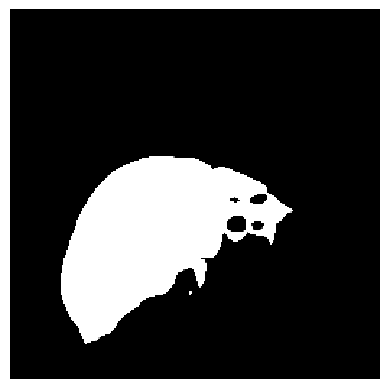} &
			\includegraphics[width=0.8cm]{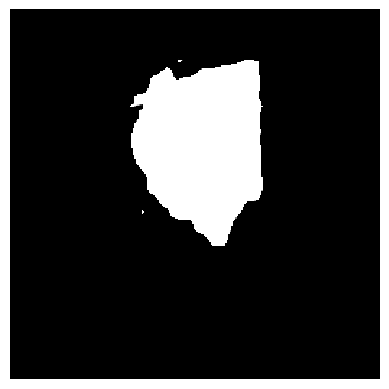} & \includegraphics[width=0.8cm]{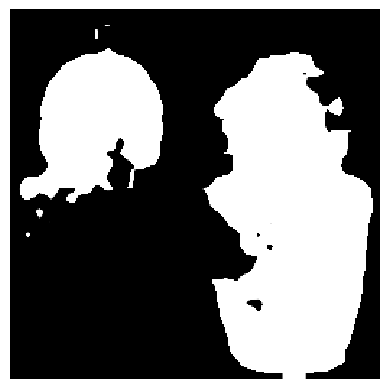} & \includegraphics[width=0.8cm]{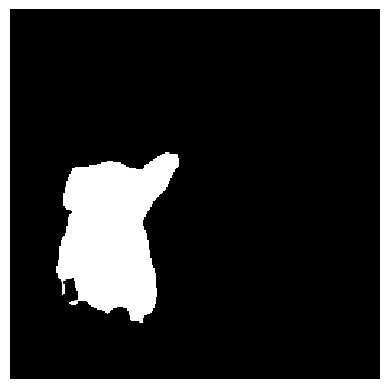} & \includegraphics[width=0.8cm]{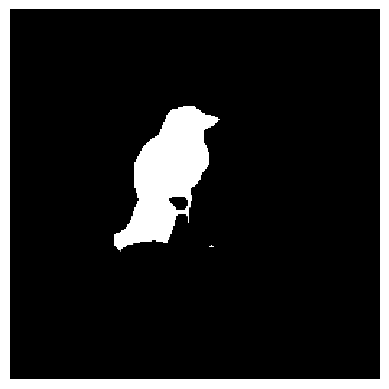} \\
			\hline 
			
			\tiny Channel 0 & \includegraphics[width=0.8cm]{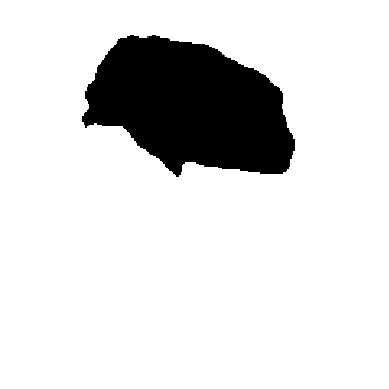} & 
			\includegraphics[width=0.8cm]{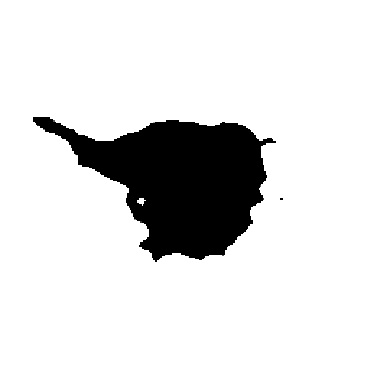} & \includegraphics[width=0.8cm]{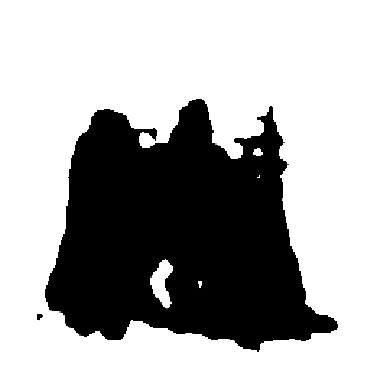} & \includegraphics[width=0.8cm]{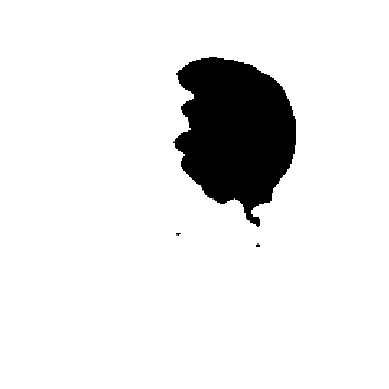} &
			\includegraphics[width=0.8cm]{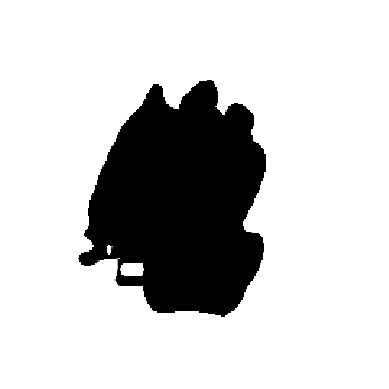} & \includegraphics[width=0.8cm]{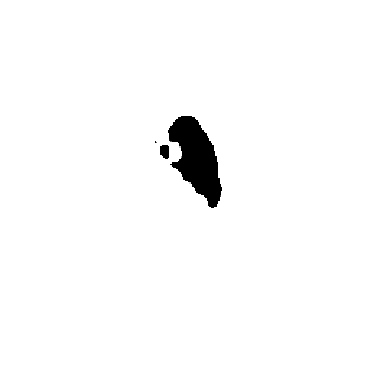} & \includegraphics[width=0.8cm]{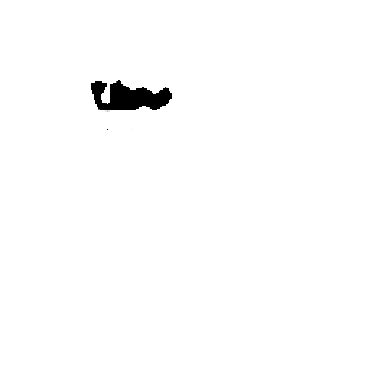} & \includegraphics[width=0.8cm]{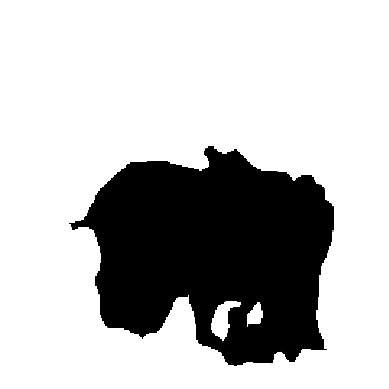} \\
			\tiny Channel 1 & \includegraphics[width=0.8cm]{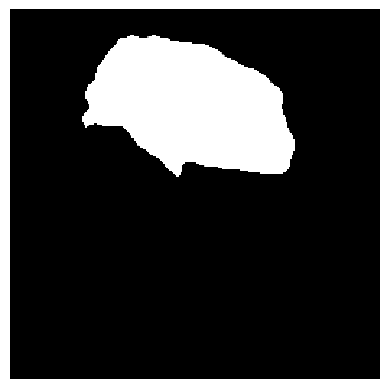} & 
			\includegraphics[width=0.8cm]{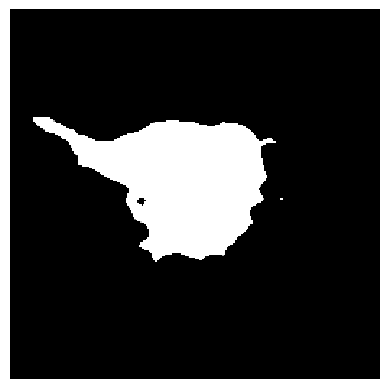} & \includegraphics[width=0.8cm]{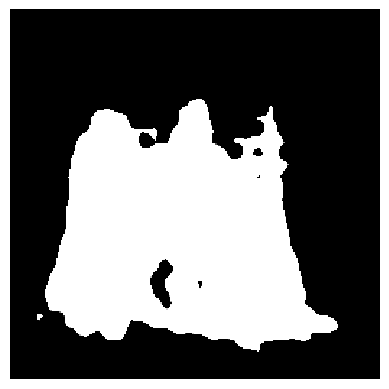} & \includegraphics[width=0.8cm]{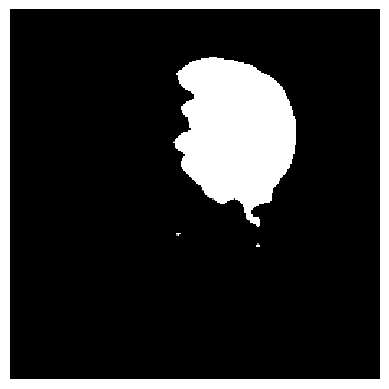} &
			\includegraphics[width=0.8cm]{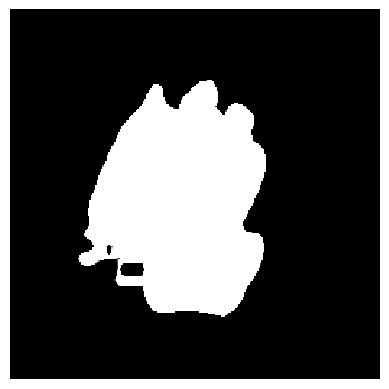} & \includegraphics[width=0.8cm]{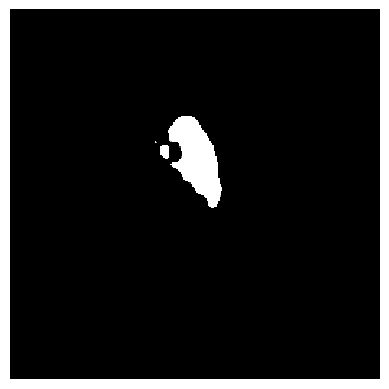} & \includegraphics[width=0.8cm]{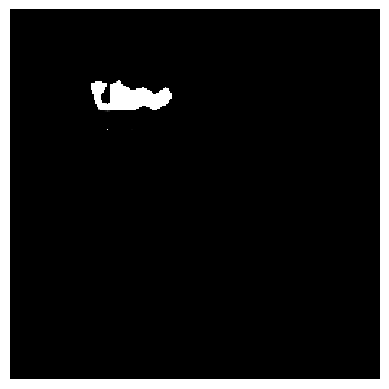} & \includegraphics[width=0.8cm]{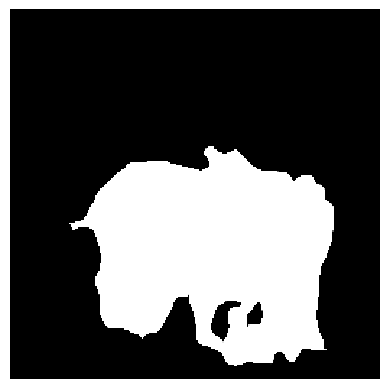} \\
		\end{tabular}
	\end{adjustbox}
	\caption{Sparse outputs of the network using CAS loss }
	
	\label{fig:sprse}
\end{figure*}

The network using CAS loss has sparse solutions which can be visualised as sparse outputs as shown in Figure \ref{fig:sprse}.

\subsection{Class-Imbalance} \label{sm_ci}
\begin{figure} [h!]
	\includegraphics[scale=0.35]{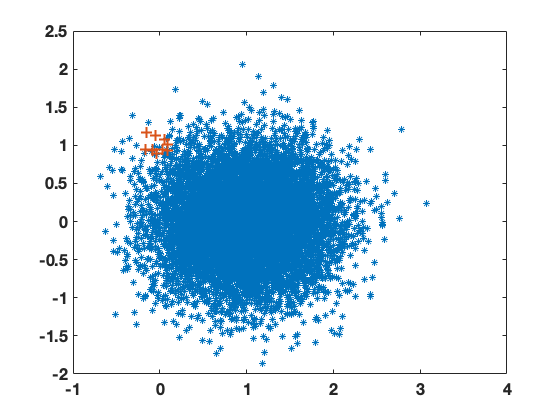}
	\caption{Data Sample}
	\label{fig:classIB}
\end{figure}

To test the accuracy of the class-agnostic segmentation loss in tackling class imbalance, we test on an artificially generated toy example. This allows us to specifically focus on the class agnostic property of the loss function while eliminating other factors. We generate data for two classes in 2D where class one is centred around point $(1,0)$ and class two is centred around point $(0,1)$. The samples for both classes have random Gaussian noise added to each component and hence are scattered around the class centres with variance value of 0.2. The two classes are also highly unbalanced with class 1 having 10000 data points and class 2 having only 10 points (see Figure \ref{fig:classIB}).

We generate 2 sets of data for training and testing respectively. We train a 2 layer fully connected networks with 10 hidden units and test on testing data. The results are summarized in Table \ref{ci_t}. The CE loss fails to perform in this case where data is highly unbalanced and assigns all output labels to belong to class 1. On the other hand, CAS loss is immune to this class imbalance and performs better.  

\begin{table}[h!]
	\centering
	\caption{CE and CAS comparison}

	\begin{tabular}{c|c|c}
		
		\multicolumn{3}{c}{CE Results} \\
		\hline
		\hline
		output $\backslash$ label  & class 1   & class 2  \\
		\hline
		class 1 & 10000 & 10 \\
		class 2 & 0 & 0
	\end{tabular}
	\quad
	\begin{tabular}{c|c|c}
		\multicolumn{3}{c}{CAS Results} \\
		\hline
		\hline 
		output $\backslash$ label&  class 1  & class 2 \\
		\hline
		class 1 & 9899 & 0 \\
		class 2 & 101 & 10
	\end{tabular}
	\label{ci_t}
\end{table}

\section{Low-fidelity data setting explanation} \label{a_lfd}
\begin{figure*}[h!]
	\centering
	\begin{adjustbox}{width=1\textwidth}
		\begin{tabular}{p{1.5cm}|cccccccc}
			\tiny Image & \includegraphics[width=0.8cm, height=0.8cm]{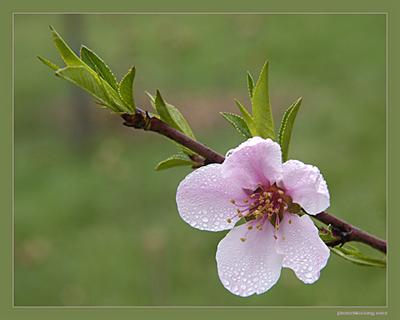} &
			\includegraphics[width=0.8cm, height=0.8cm]{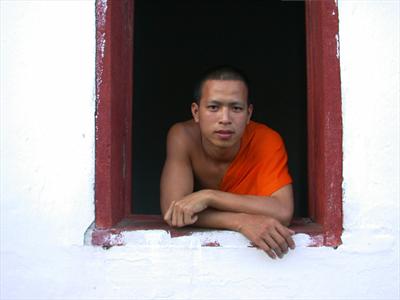} &
			\includegraphics[width=0.8cm, height=0.8cm]{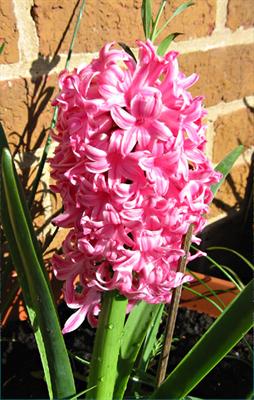} &
			\includegraphics[width=0.8cm, height=0.8cm]{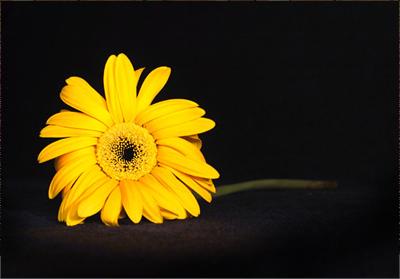} &\includegraphics[width=0.8cm, height=0.8cm]{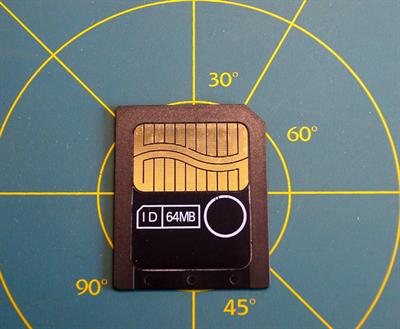} &\includegraphics[width=0.8cm, height=0.8cm]{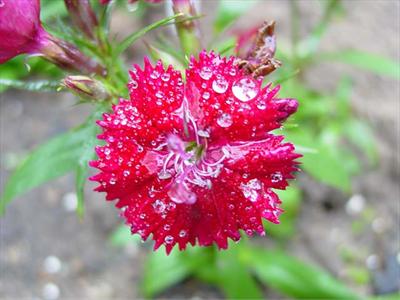} &\includegraphics[width=0.8cm, height=0.8cm]{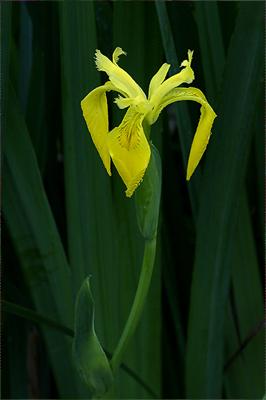} &
			\includegraphics[width=0.8cm, height=0.8cm]{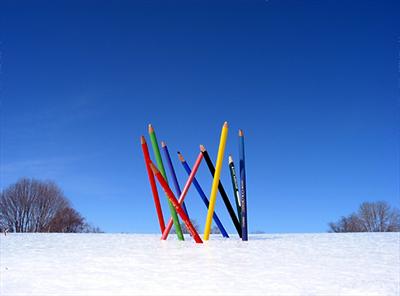} \\
			
			\tiny Ground Truth & \includegraphics[width=0.8cm, height=0.8cm]{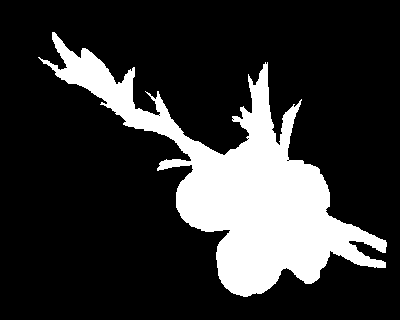} &
			\includegraphics[width=0.8cm, height=0.8cm]{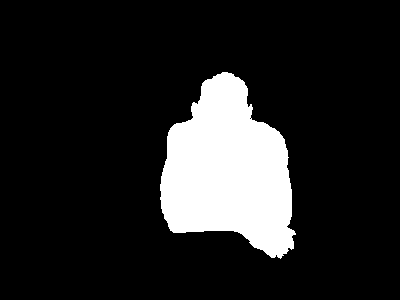} &
			\includegraphics[width=0.8cm, height=0.8cm]{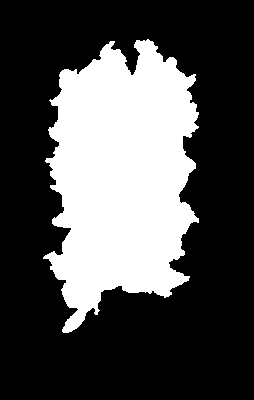} &
			\includegraphics[width=0.8cm, height=0.8cm]{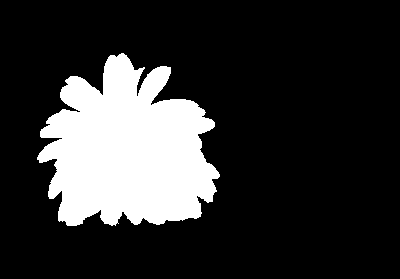} &\includegraphics[width=0.8cm, height=0.8cm]{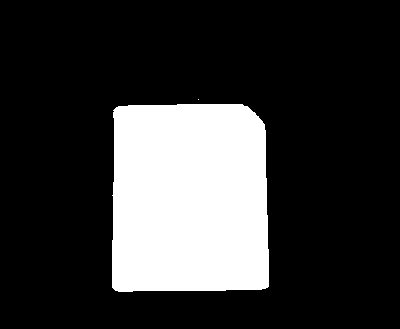} &\includegraphics[width=0.8cm, height=0.8cm]{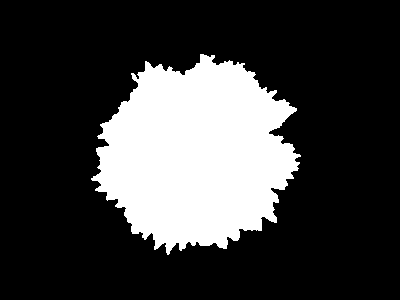} &\includegraphics[width=0.8cm, height=0.8cm]{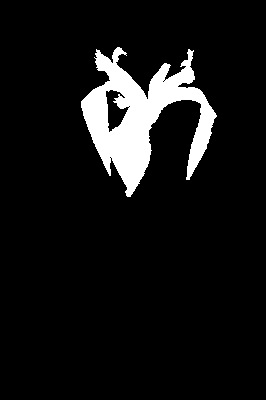} &
			\includegraphics[width=0.8cm, height=0.8cm]{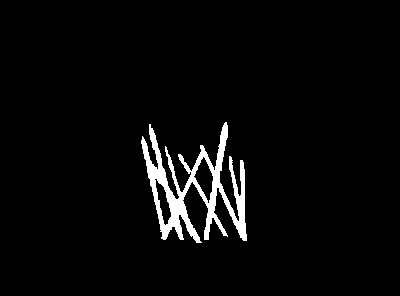} \\

			\hline
			\tiny Image & \includegraphics[width=0.8cm, height=0.8cm]{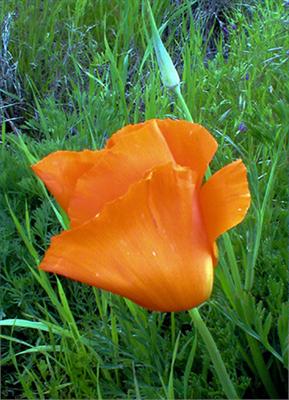} &
			\includegraphics[width=0.8cm, height=0.8cm]{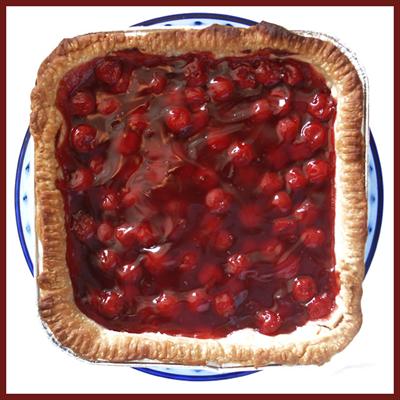} &
			\includegraphics[width=0.8cm, height=0.8cm]{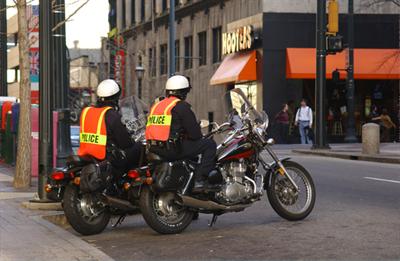} &
			\includegraphics[width=0.8cm, height=0.8cm]{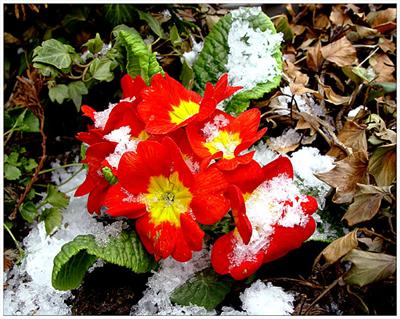} &\includegraphics[width=0.8cm, height=0.8cm]{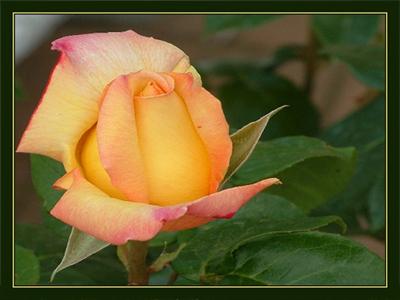} &\includegraphics[width=0.8cm, height=0.8cm]{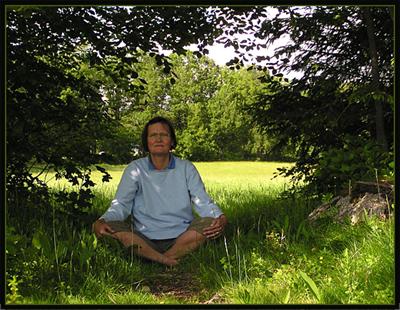} &\includegraphics[width=0.8cm, height=0.8cm]{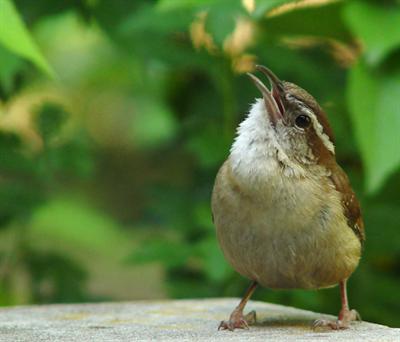} &
			\includegraphics[width=0.8cm, height=0.8cm]{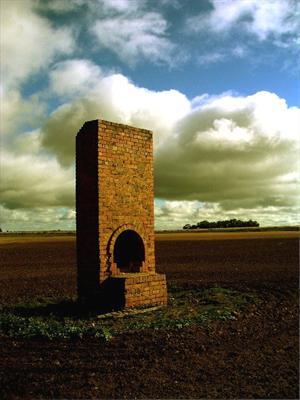} \\
			
			\tiny Ground Truth & \includegraphics[width=0.8cm, height=0.8cm]{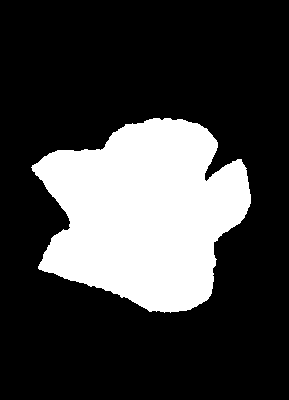} &
			\includegraphics[width=0.8cm, height=0.8cm]{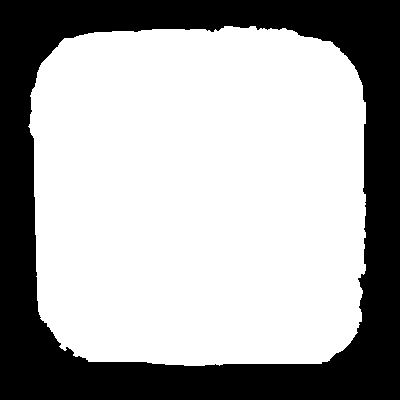} &
			\includegraphics[width=0.8cm, height=0.8cm]{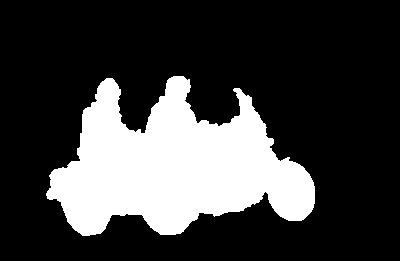} &
			\includegraphics[width=0.8cm, height=0.8cm]{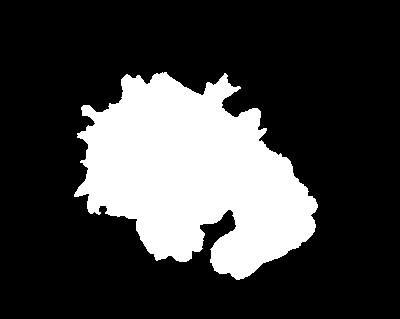} &\includegraphics[width=0.8cm, height=0.8cm]{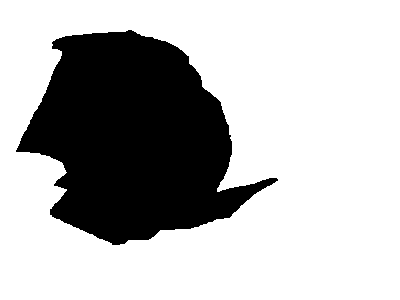} &\includegraphics[width=0.8cm, height=0.8cm]{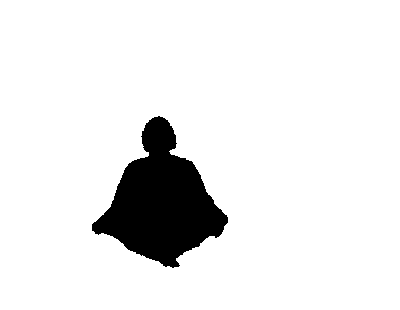} &\includegraphics[width=0.8cm, height=0.8cm]{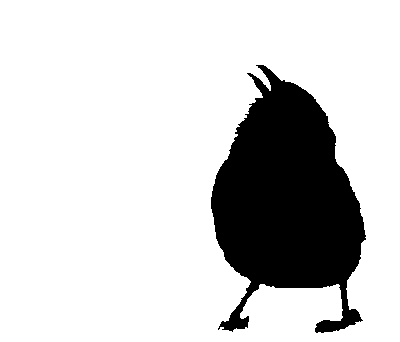} &
			\includegraphics[width=0.8cm, height=0.8cm]{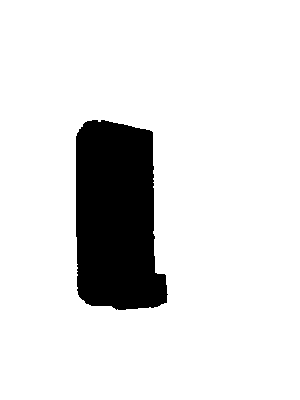} \\

			\hline
			\tiny Image & \includegraphics[width=0.8cm, height=0.8cm]{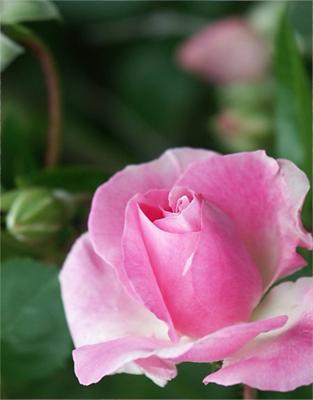} &
			\includegraphics[width=0.8cm, height=0.8cm]{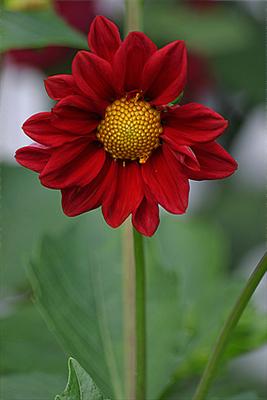} &
			\includegraphics[width=0.8cm, height=0.8cm]{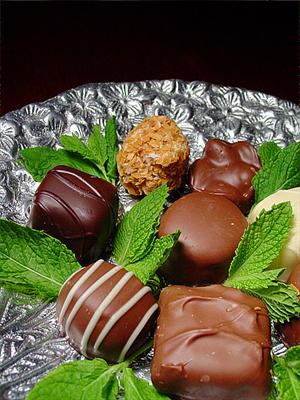} &
			\includegraphics[width=0.8cm, height=0.8cm]{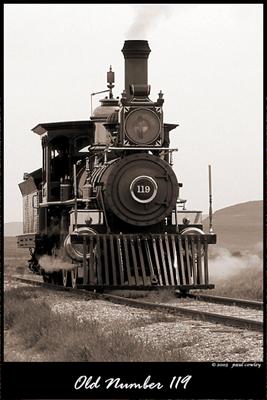} &\includegraphics[width=0.8cm, height=0.8cm]{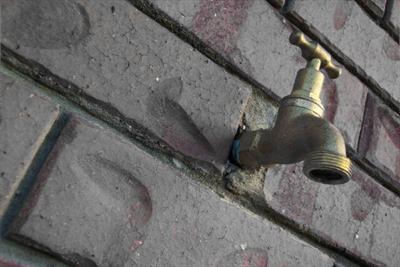} &\includegraphics[width=0.8cm, height=0.8cm]{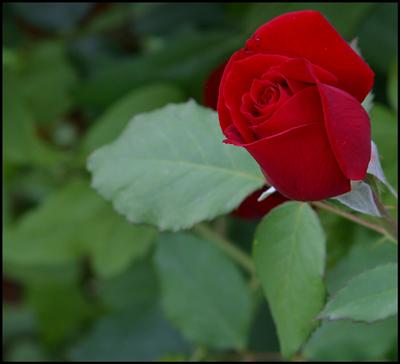} &\includegraphics[width=0.8cm, height=0.8cm]{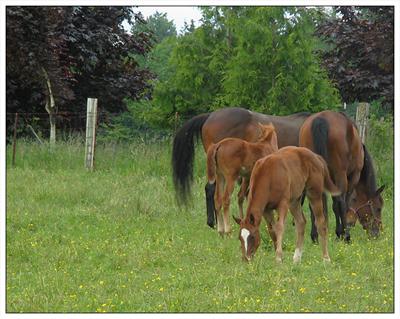} &
			\includegraphics[width=0.8cm, height=0.8cm]{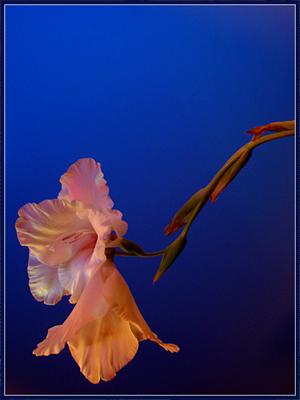} \\
			\tiny Ground Truth & \includegraphics[width=0.8cm, height=0.8cm]{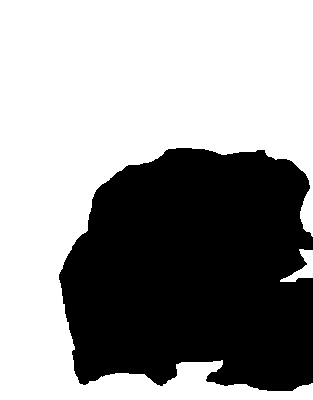} &
			\includegraphics[width=0.8cm, height=0.8cm]{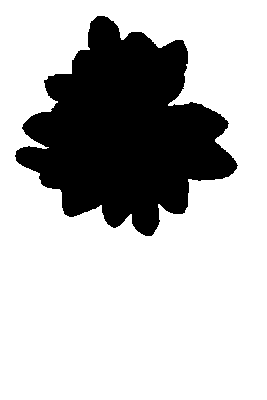} &
			\includegraphics[width=0.8cm, height=0.8cm]{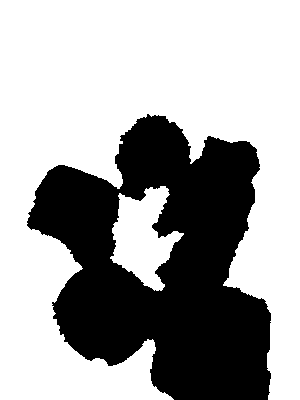} &
			\includegraphics[width=0.8cm, height=0.8cm]{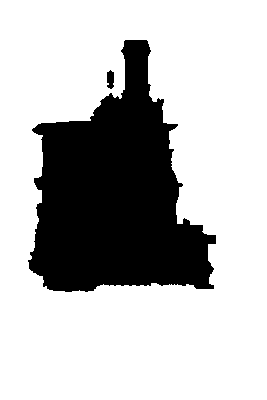} &\includegraphics[width=0.8cm, height=0.8cm]{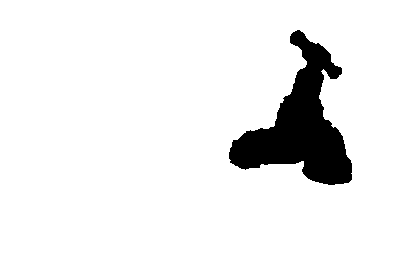} &\includegraphics[width=0.8cm, height=0.8cm]{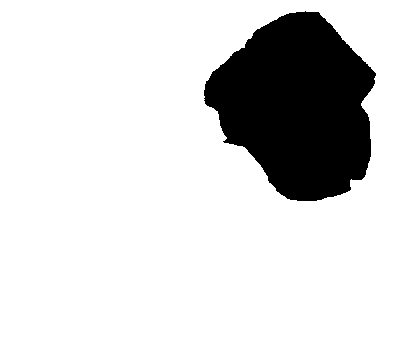} &\includegraphics[width=0.8cm, height=0.8cm]{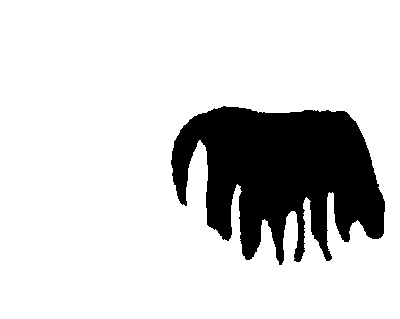} &
			\includegraphics[width=0.8cm, height=0.8cm]{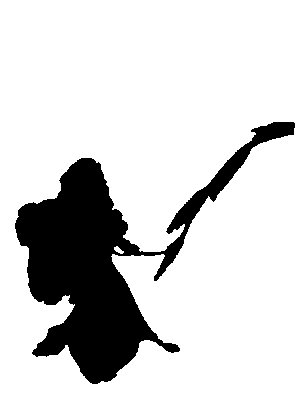} \\
			
		\end{tabular}
	\end{adjustbox}
	\caption{Low-fidelity training data sample}
	
	\label{fig:lfd_data}
\end{figure*}

An example of what the low-fidelity data looks like is shown in Figure \ref{fig:lfd_data}, where 50\% of the data was normal and half was flipped.

\section{Dataset-wise models' Results} \label{a_d}
In Table \ref{table:sup_full} the quantitative results of all our models trained on different datasets and tested on 7 saliency datasets are shown, along with comparison with state-of-the-art methods. 

As seen in Table \ref{table:sup_full}, all the models, even the state-of-the-art ones, perform better when tested on the datasets belonging to the training set domain. This is the anticipated issue of dataset bias \cite{Torralba}, which is a shortcoming of the breadth of the various datasets. 

\section{Visual Results for DeepLab model for Salient Object Detection and  Texture Segmentation} \label{a_tex}
\begin{figure*}[h!]
	\centering
	\begin{adjustbox}{width=1\textwidth}
		\begin{tabular}{cc||c|c||c}
			
			\tiny Image & \tiny Ground Truth & \tiny DeepLab-CAS (hfd)&  \tiny DeepLab-CE (hfd)  & \tiny DeepLab-CAS (lfd)\\

			\midrule
			
			\includegraphics[width=0.8cm]{all/img/image_0.png} & \includegraphics[width=0.8cm]{all/gt/target_0.png}  &\includegraphics[width=0.8cm]{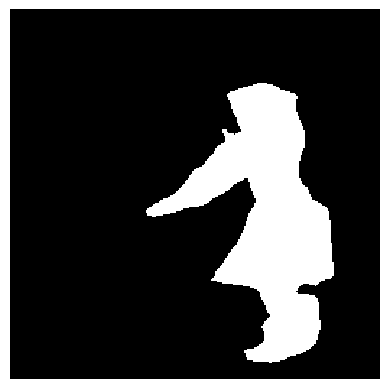} &  \includegraphics[width=0.8cm]{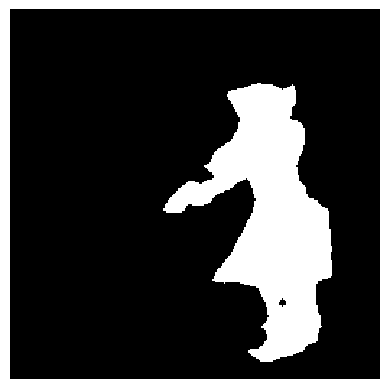}   & \includegraphics[width=0.8cm]{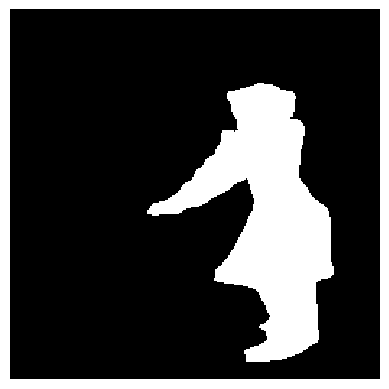}\\

			\includegraphics[width=0.8cm]{all/img/ILSVRC2012_test_00000649.jpg} & \includegraphics[width=0.8cm]{all/gt/target_65} &\includegraphics[width=0.8cm]{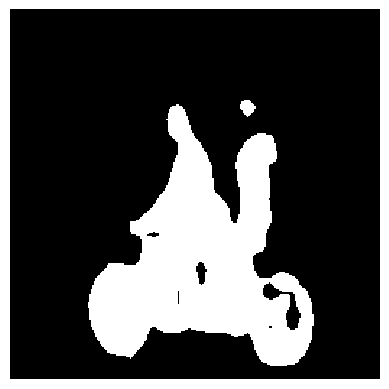} &  \includegraphics[width=0.8cm]{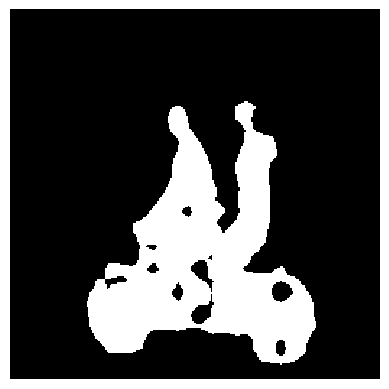}  &  \includegraphics[width=0.8cm]{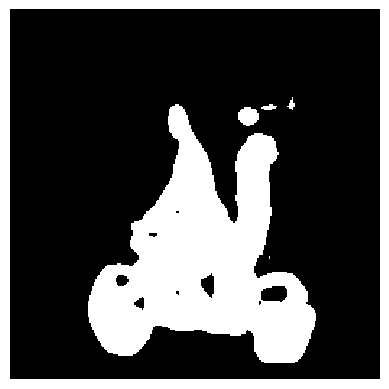}  \\

			\includegraphics[width=0.8cm]{all/img/ILSVRC2012_test_00000998} & \includegraphics[width=0.8cm]{all/gt/target_108.png}   &\includegraphics[width=0.8cm]{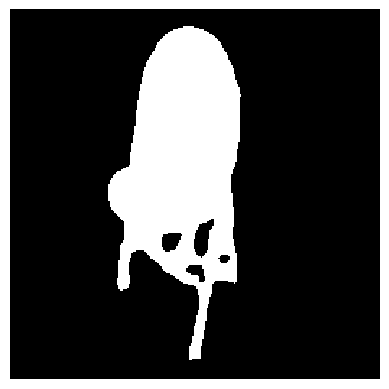} &  \includegraphics[width=0.8cm]{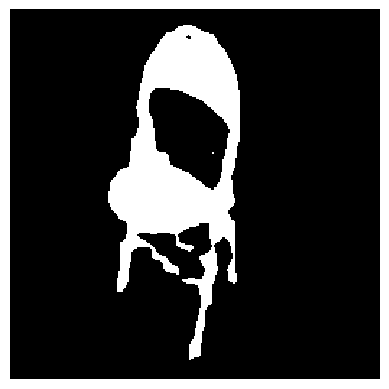} &  \includegraphics[width=0.8cm]{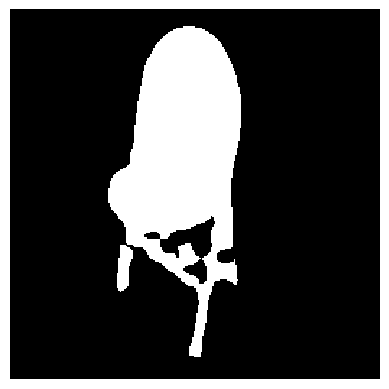}		\\

			\includegraphics[width=0.8cm]{all/img/0016.png} & \includegraphics[width=0.8cm]{all/gt/target_11.png} &\includegraphics[width=0.8cm]{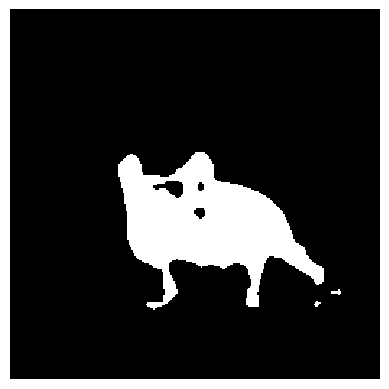} &  \includegraphics[width=0.8cm]{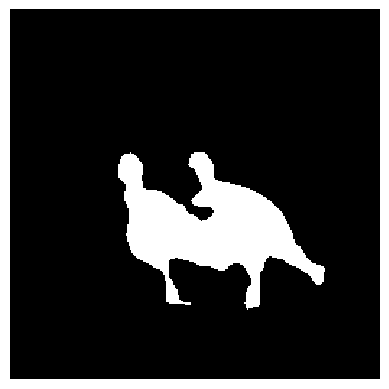}  &  \includegraphics[width=0.8cm]{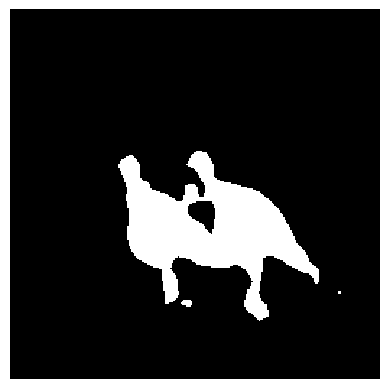}   \\

			\includegraphics[width=0.8cm, height=0.8cm]{all/img/0935.jpg} & \includegraphics[width=0.8cm]{all/gt/target_934.png} &\includegraphics[width=0.8cm]{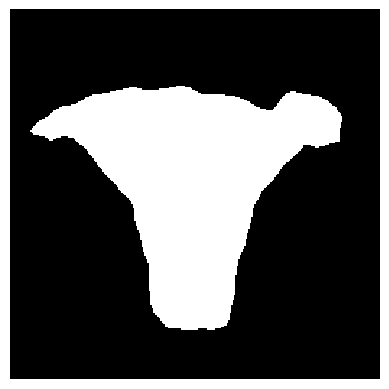} &  \includegraphics[width=0.8cm]{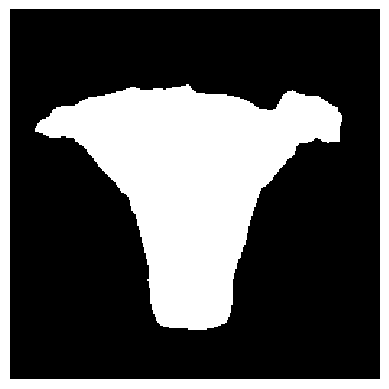}&  \includegraphics[width=0.8cm]{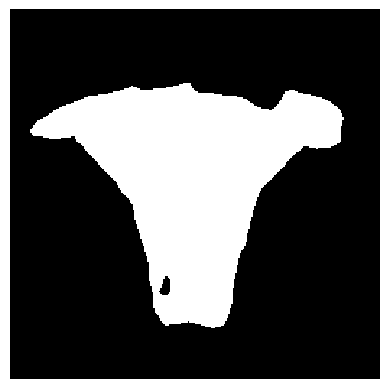}  \\

			\includegraphics[width=0.8cm]{all/img/image_6.png} & \includegraphics[width=0.8cm]{all/gt/target_6.png}  &\includegraphics[width=0.8cm]{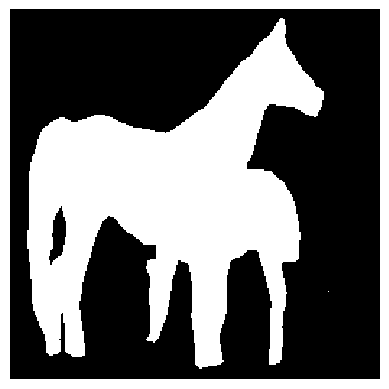} &  \includegraphics[width=0.8cm]{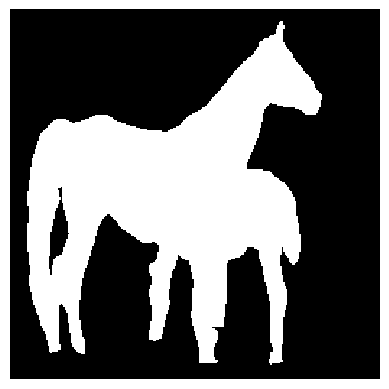}  & \includegraphics[width=0.8cm]{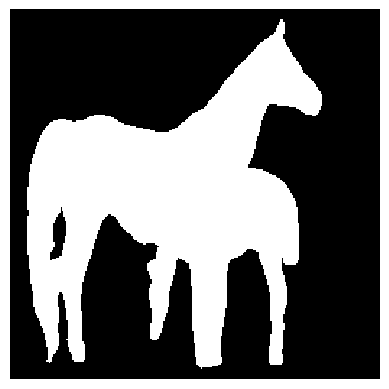}  \\

			\includegraphics[width=0.8cm]{all/img/ILSVRC2012_test_00000633.jpg} & \includegraphics[width=0.8cm]{all/gt/target_63.png} &\includegraphics[width=0.8cm]{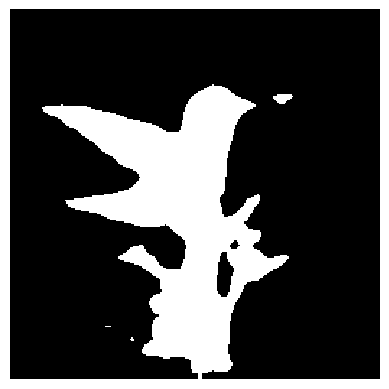} &  \includegraphics[width=0.8cm]{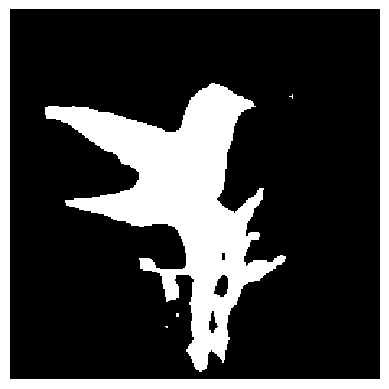} &  \includegraphics[width=0.8cm]{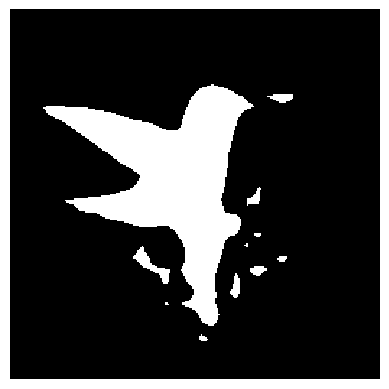}   \\

			\includegraphics[width=0.8cm]{all/img/0151.jpg} & \includegraphics[width=0.8cm]{all/gt/target_150}  &\includegraphics[width=0.8cm]{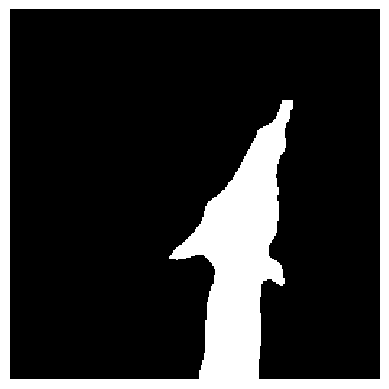} &  \includegraphics[width=0.8cm]{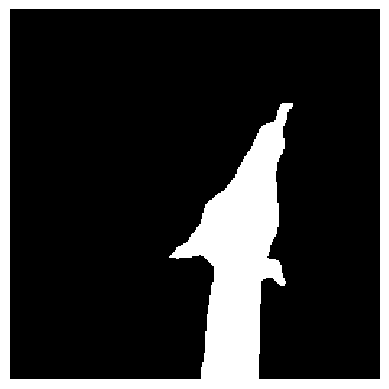} &  \includegraphics[width=0.8cm]{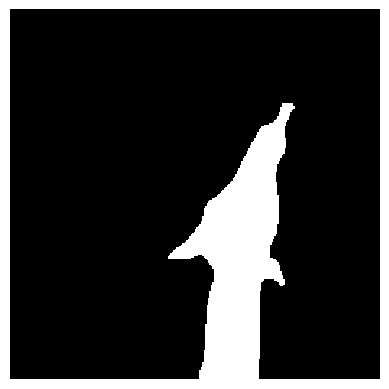}  \\
			
			\includegraphics[width=0.8cm]{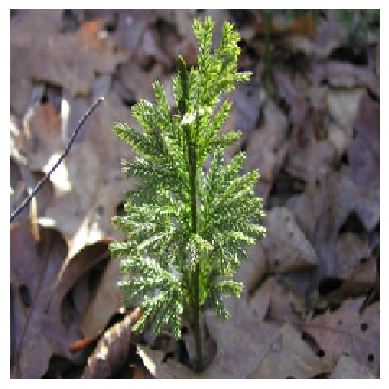} & \includegraphics[width=0.8cm]{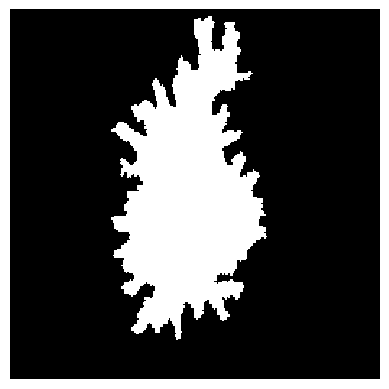}  &\includegraphics[width=0.8cm]{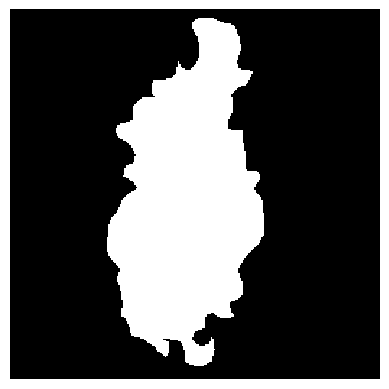} &  \includegraphics[width=0.8cm]{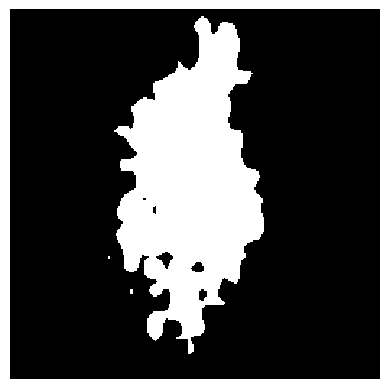}
			&  \includegraphics[width=0.8cm]{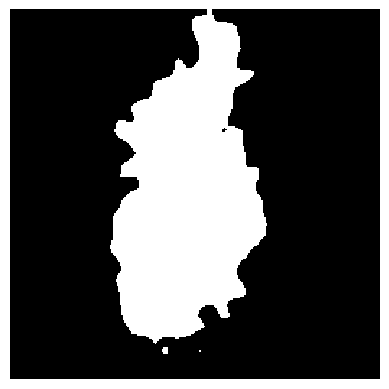}   \\

			\includegraphics[width=0.8cm]{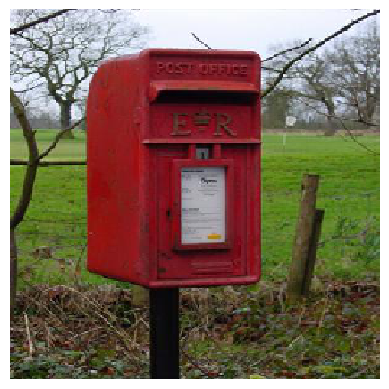} & \includegraphics[width=0.8cm]{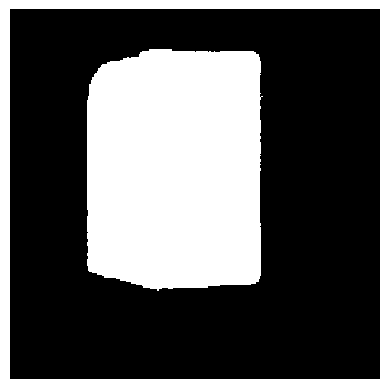}  &\includegraphics[width=0.8cm]{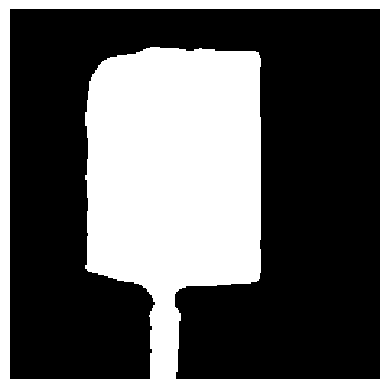} &  \includegraphics[width=0.8cm]{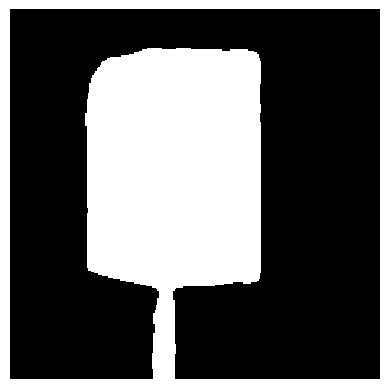} &  \includegraphics[width=0.8cm]{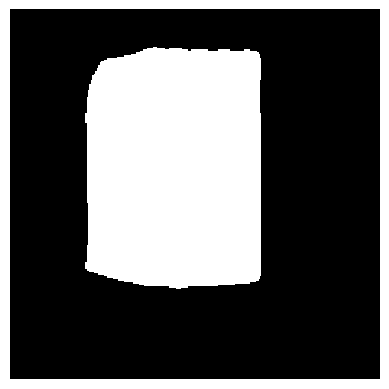} \\

			\includegraphics[width=0.8cm]{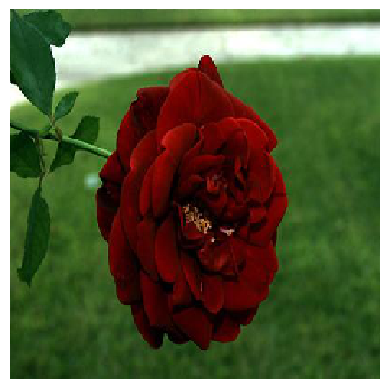} & \includegraphics[width=0.8cm]{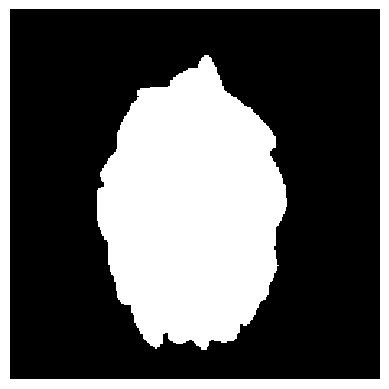}  &\includegraphics[width=0.8cm]{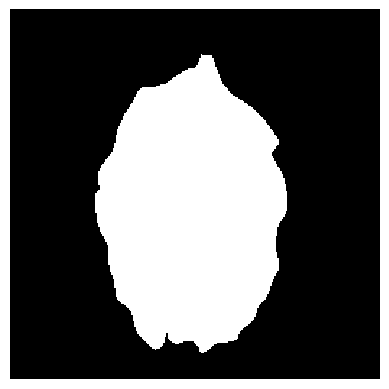} &  \includegraphics[width=0.8cm]{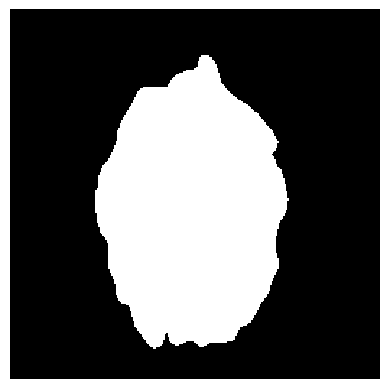}  &  \includegraphics[width=0.8cm]{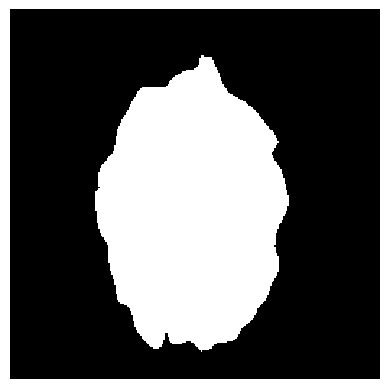}  \\

		\end{tabular}
	\end{adjustbox}
	\caption{Visual Results for models trained with DeepLab-v3 architecture (hfd - denotes trained on high-fidelity data; lfd - denotes trained on low-fidelity data)}
	\label{tbl:vis}
\end{figure*}

Following from Section 6 in the paper, the visual results for  DeepLab-CE model i.e., DeepLab-v3 architecture with cross-entropy loss function and DeepLab-CAS model i.e., DeepLab-v3 architecture with class-agnostic segmentation loss are shown in Figure \ref{tbl:vis} for salient object detection and in Figure \ref{fig:text} for texture segmentation. The quantitative results were summarised in Table 2 in the paper for salient object detection and Table 4 for texture segmentation.

All these results concur with those performed on FCN-ResNet-101 architecture in Section 4 in the paper. These verify our claims empirically, about the working of our CAS loss function with any neural network. Also, the performance of the CAS loss is   comparable and majority of times better than the cross-entropy loss function, for both the tasks of salient object detection and segmentation.

\begin{table*}[h!]
	\begin{center}
		\begin{adjustbox}{width=1\textwidth}	
			\begin{threeparttable}
				\begin{tabular}{|p{3.5cm}|p{2cm}||c c|c c| c c| c c|c c| c c|cc|}
					\hline
					Model & Training Set & \multicolumn{2}{c|}{MSRA-B} & \multicolumn{2}{c|}{DUTS-TE} & \multicolumn{2}{c|}{ECSSD} & \multicolumn{2}{c|}{PASCAL-S} & \multicolumn{2}{c|}{HKU-IS} &  \multicolumn{2}{c|}{THUR15k}
					& \multicolumn{2}{c}{DUT-OMRON}\\ 
					& &  $F_\beta \uparrow$ & MAE $\downarrow$ & $F_\beta \uparrow$ & MAE $\downarrow$ & $F_\beta\uparrow$ & MAE  $\downarrow$  & $F_\beta \uparrow$ & MAE  $\downarrow$& $F_\beta\uparrow$ & MAE $\downarrow$ & $F_\beta\uparrow$ & MAE  $\downarrow$  & $F_\beta\uparrow$ & MAE  $\downarrow$  \\
					\hline\hline
					ResNet-pre-CAS (ours) & MSRA-B  & \textcolor{red}{0.985}& \textcolor{red}{0.010} & 0.836 & 0.088 & 0.874 & 0.071 & 0.818 & 0.112 & 0.887 & 0.056 & 0.891 & 0.075 & \textcolor{blue}{0.876} & 0.066 \\
					\hline
					ResNet-m-CE (ours)& MSRA-B & \textcolor{blue}{0.958}  & 0.030  & \textcolor{blue}{0.910}  &0.067  & 0.905 & 0.068 & 0.868 & 0.100 & 0.921 & 0.053 & 0.928 & \textcolor{blue}{0.065}  & \textcolor{red}{0.920} & 0.059\\ 
					ResNet-m-CAS (ours)& MSRA-B &  0.944 & 0.037  & 0.853 & 0.091 & 0.876 & 0.079 & 0.811& 0.124 & 0.931 & 0.057 & 0.930 & 0.075 & 0.863 & 0.080\\ 
					\hline 
					ResNet-d-CE (ours)& DUTS-TR&  0.947 &0.0.65  &  \textcolor{red}{0.919}&  0.055& 0.867 & 0.077& 0.876 & 0.091 & 0.928 & 0.044 &  \textcolor{red}{0.935} & \textcolor{red}{0.057} & 0.867 & 0.062\\ 
					ResNet-d-CAS  (ours)& DUTS-TR &0.932 & 0.046& 0.871 & 0.071 & 0.888 & 0.075 & 0.840 & 0.121 & \textcolor{red}{0.939} & 0.050 & \textcolor{blue}{0.931} & 0.073 & 0.875 & 0.071\\
					\hline 
					DeepLab-CAS (ours)& DUTS-TR & 0.931 & 0.040 & 0.850 & 0.070 & 0.864 & 0.072 & 0.800 & 0.111 & 0.882 &0.054 & 0.888 & 0.069 & 0.865 & 0.060\\
					
					DeepLab-CE (ours)&  DUTS-TR & 0.928  & 0.039 & 0.847 & 0.070 & 0.867 & 0.069 & 0.805 & 0.110 & 0.880 & 0.052 & 0.881 & 0.070 & 0.856 & 0.061\\ 
					\hline
					BAS-Net \cite{Qin2019} & DUTS-TR & -  & - & 0.860   &    0.047   &\textcolor{blue}{0.942}   &  0.037 & 0.854 & 0.076 & 0.921 & 0.039 & - & - & 0.805 & 0.056\\
					
					PoolNet \cite{Liu2019}	 & MSRA-B + HKU-IS  &-&-& 0.892 & \textcolor{red}{0.036}& \textcolor{red}{0.945} & 0.038 & \textcolor{blue}{0.880}&\textcolor{red}{0.065} & \textcolor{blue}{0.935}&\textcolor{blue}{0.030}&-&- & 0.833 & \textcolor{blue}{0.053}\\
					
					CPSNet \cite{Zeng2019}	 & COCO+DUT  &-&-&-&-& 0.878& 0.096 & 0.790 &0.134   &-&-&-&- & 0.718 & 0.114\\
					
					PFAN \cite{Zhao2019} & DUTS-TR&  - & - & 0.870 & \textcolor{blue}{0.040} & 0.931 & \textcolor{red}{0.032} & \textcolor{red}{0.892} & \textcolor{blue}{0.067} & 0.926 & 0.032 & -&- & 0.855 & \textcolor{red}{0.041} \\		
					PAGENET+CRF\cite{Wang2019} & THUS10k& - & - & 0.817 & 0.047  &  0.926 & \textcolor{blue}{0.035}  & 0.835 & 0.074 & 0.920 & \textcolor{blue}{0.030} & - & -  & 0.770 & 0.063 \\ 
					PAGENET\cite{Wang2019}  & THUS10k  & -   & -   & 0.815 & 0.051  & 0.924 & 0.042 &  0.835 & 0.078 & 0.918 & 0.037 & -&- & 0.770 & 0.066\\ 
					
					HED \cite{Houa}	 & MSRA-B  & 0.927 & \textcolor{blue}{0.028}& -   &  -  & 0.915 & 0.052  & 0.830&0.080 & 0.913 & 0.039 &-&-&0.764 & 0.070 \\
					
					DNA 	\cite{Liu2019a} & DUTS-TR  & - &-& 0.873 & \textcolor{blue}{0.040} & 0.938 & 0.040 & -&-& 0.934 &\textcolor{red}{0.029}&  0.796 & 0.068 & 0.805 & 0.056\\

					\hline
					
				\end{tabular}
				\begin{tablenotes}
					\item[] -pre- represents model pre-trained using cross-entropy loss and then trained using CAS loss; -m- represents the model trained on MSRA-B dataset;  -d- represents the model trained on DUTS-TR dataset ; \item[] \textcolor{red}{red} represents the  best score value on the dataset; \textcolor{blue}{blue} represents the second best score on the dataset; - represents the dataset was not tested by the method
				\end{tablenotes}
			\end{threeparttable}
		\end{adjustbox}
		
		\caption{Numerical Results on High-Fidelity Data for all the models trained on different datasets}
		\label{table:sup_full}
	\end{center}
\end{table*}

\def\fWidRTex{0.7\linewidth} 
\def\fHeiRTex{.5\linewidth}  
\begin{figure*}[h!]
	\begin{adjustbox}{width=1\textwidth}
		\begin{tabular}{c}
			
			\Huge Images \\
			\includegraphics[width=\fWidRTex,height=\fHeiRTex]{{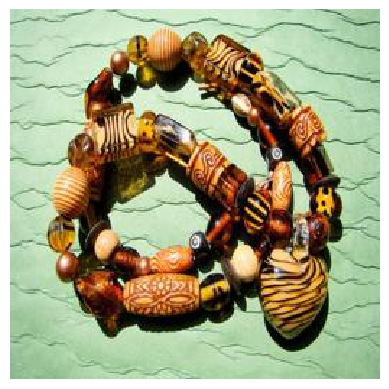}} \qquad \qquad
			\includegraphics[width=\fWidRTex,height=\fHeiRTex]{{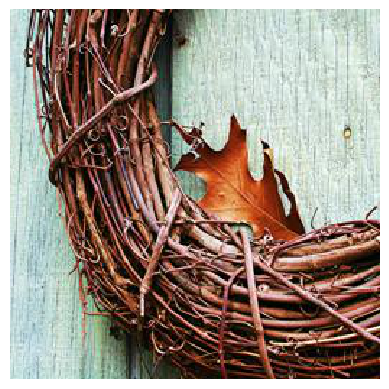}}\qquad \qquad
			\includegraphics[width=\fWidRTex,height=\fHeiRTex]{{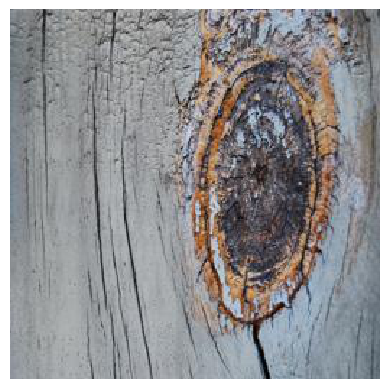}}\qquad \qquad
			\includegraphics[width=\fWidRTex,height=\fHeiRTex]{{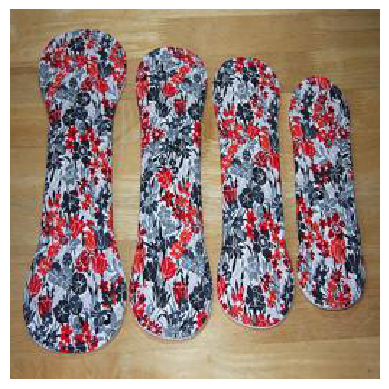}}\qquad \qquad
			\includegraphics[width=\fWidRTex,height=\fHeiRTex]{{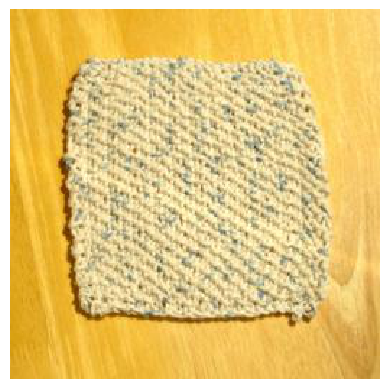}}\qquad \qquad
			\includegraphics[width=\fWidRTex,height=\fHeiRTex]{{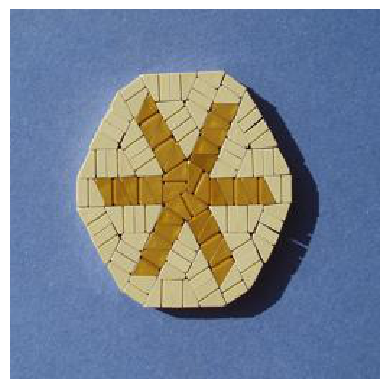}} \qquad \qquad
			\includegraphics[width=\fWidRTex,height=\fHeiRTex]{{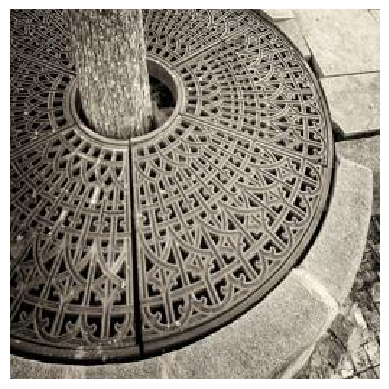}}  
			\\ \\
			
			\Huge Ground Truth\\ 
			
			\includegraphics[width=\fWidRTex,height=\fHeiRTex]{{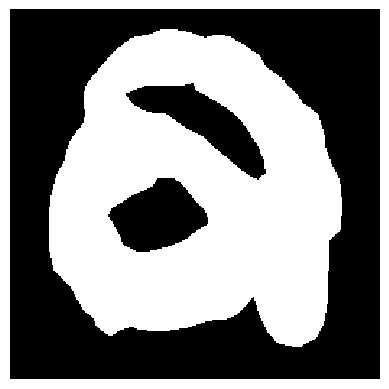}} \qquad \qquad
			\includegraphics[width=\fWidRTex,height=\fHeiRTex]{{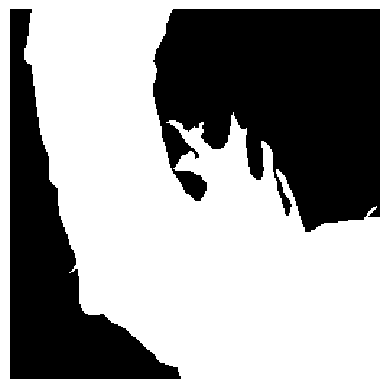}}\qquad \qquad
			\includegraphics[width=\fWidRTex,height=\fHeiRTex]{{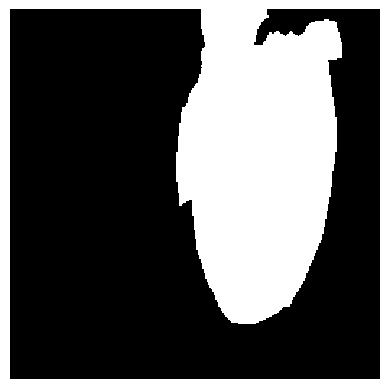}}\qquad \qquad
			\includegraphics[width=\fWidRTex,height=\fHeiRTex]{{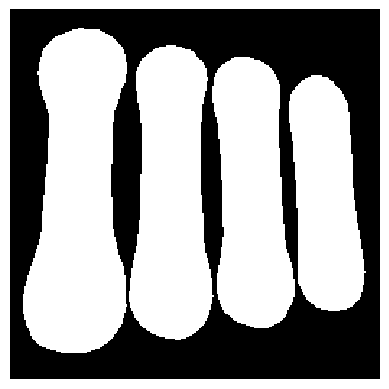}}\qquad \qquad
			\includegraphics[width=\fWidRTex,height=\fHeiRTex]{{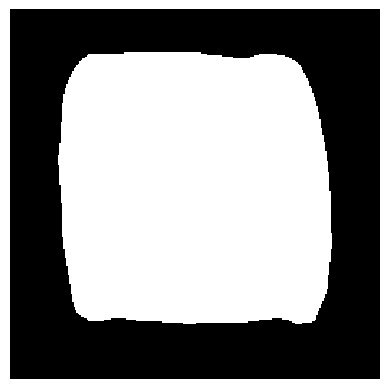}}\qquad \qquad
			
			\includegraphics[width=\fWidRTex,height=\fHeiRTex]{{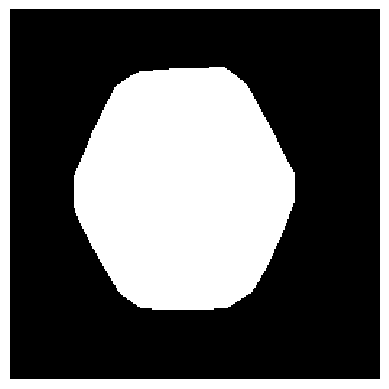}} \qquad \qquad
			\includegraphics[width=\fWidRTex,height=\fHeiRTex]{{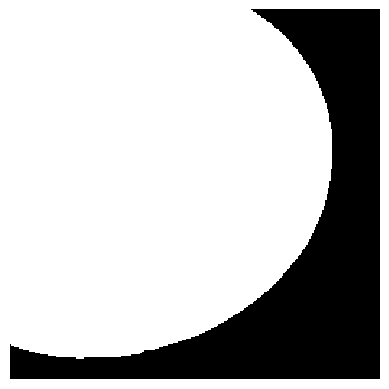}} \\ \\ 

			\Huge		DeepLab-a-CE \\
			\includegraphics[width=\fWidRTex,height=\fHeiRTex]{{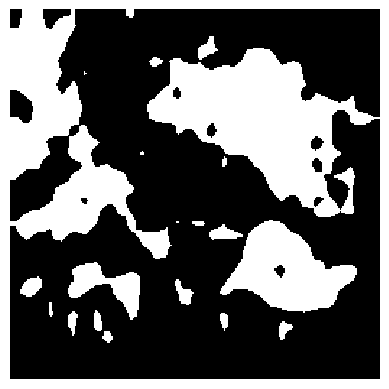}} \qquad \qquad
			\includegraphics[width=\fWidRTex,height=\fHeiRTex]{{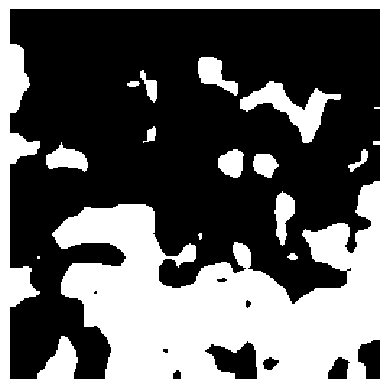}}\qquad \qquad
			\includegraphics[width=\fWidRTex,height=\fHeiRTex]{{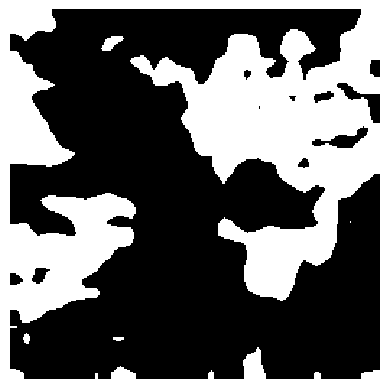}}\qquad \qquad
			\includegraphics[width=\fWidRTex,height=\fHeiRTex]{{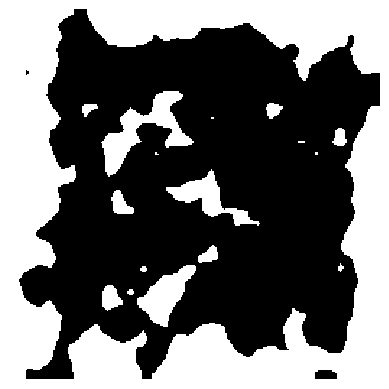}}\qquad \qquad
			\includegraphics[width=\fWidRTex,height=\fHeiRTex]{{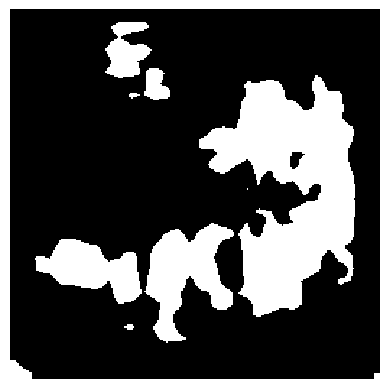}}\qquad \qquad
			
			\includegraphics[width=\fWidRTex,height=\fHeiRTex]{{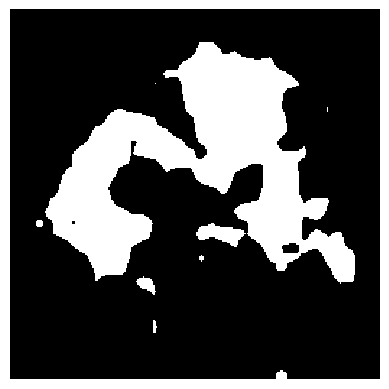}} \qquad \qquad
			\includegraphics[width=\fWidRTex,height=\fHeiRTex]{{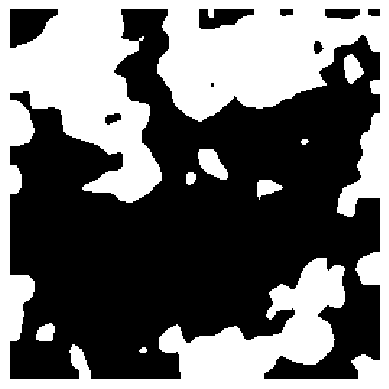}}\\ \\

			\Huge DeepLab-a-CAS \\
			\includegraphics[width=\fWidRTex,height=\fHeiRTex]{{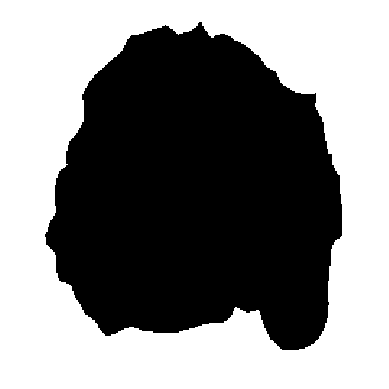}} \qquad \qquad
			\includegraphics[width=\fWidRTex,height=\fHeiRTex]{{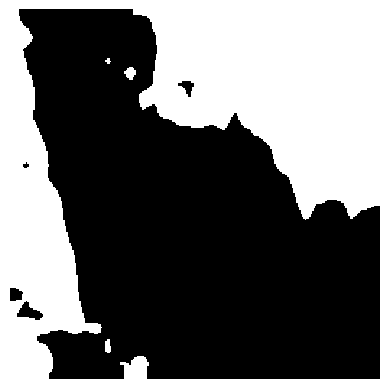}}\qquad \qquad
			\includegraphics[width=\fWidRTex,height=\fHeiRTex]{{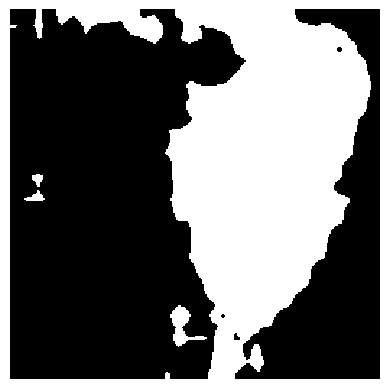}}\qquad \qquad
			\includegraphics[width=\fWidRTex,height=\fHeiRTex]{{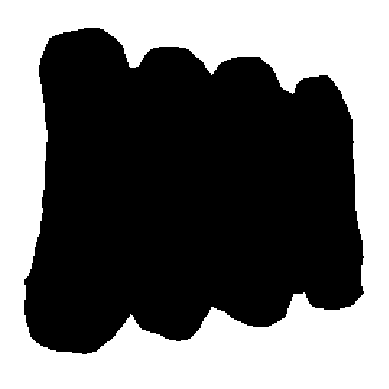}}\qquad \qquad
			\includegraphics[width=\fWidRTex,height=\fHeiRTex]{{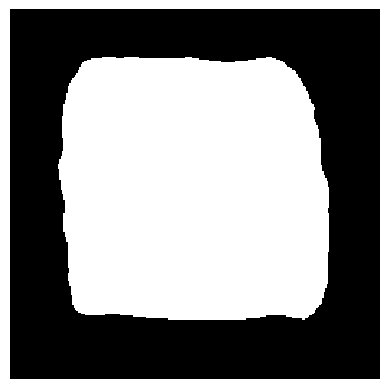}}\qquad \qquad
			
			\includegraphics[width=\fWidRTex,height=\fHeiRTex]{{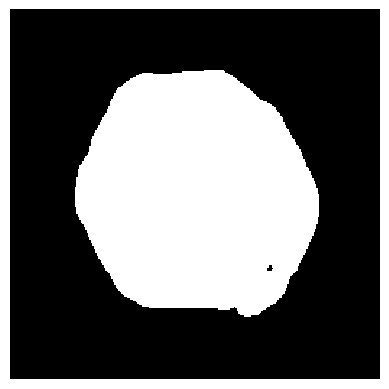}}  \qquad \qquad
			\includegraphics[width=\fWidRTex,height=\fHeiRTex]{{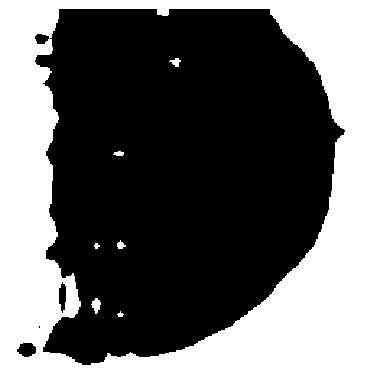}} \\ \\ \\ \\ 

			\Huge Images \\
			\includegraphics[width=\fWidRTex,height=\fHeiRTex]{{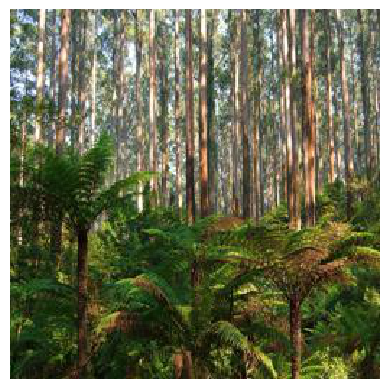}} \qquad \qquad
			\includegraphics[width=\fWidRTex,height=\fHeiRTex]{{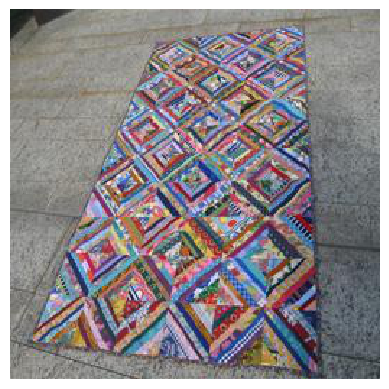}}\qquad \qquad
			\includegraphics[width=\fWidRTex,height=\fHeiRTex]{{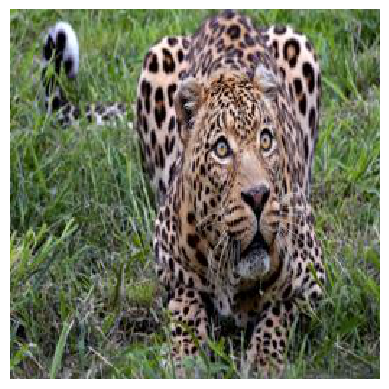}}\qquad \qquad
			\includegraphics[width=\fWidRTex,height=\fHeiRTex]{{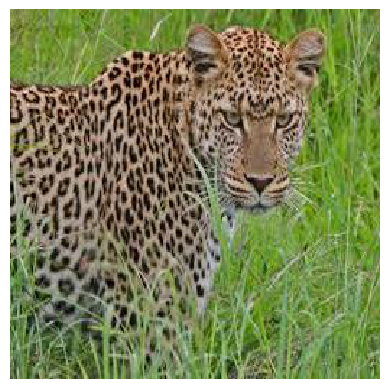}}\qquad \qquad
			\includegraphics[width=\fWidRTex,height=\fHeiRTex]{{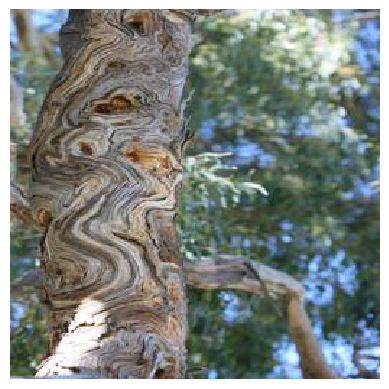}}\qquad \qquad
			
			\includegraphics[width=\fWidRTex,height=\fHeiRTex]{{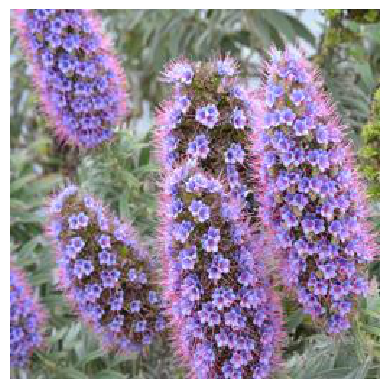}}\qquad \qquad
			\includegraphics[width=\fWidRTex,height=\fHeiRTex]{{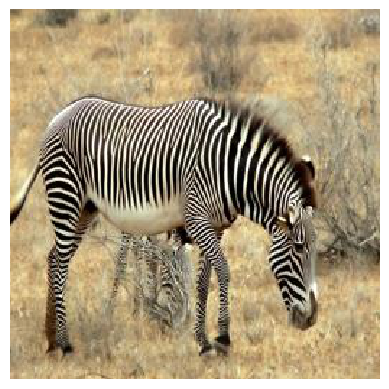}}\\ \\ 

			\Huge Ground Truth\\ 
			
			\includegraphics[width=\fWidRTex,height=\fHeiRTex]{{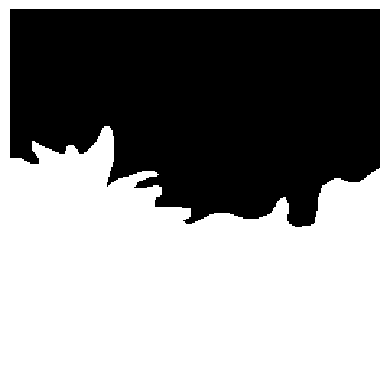}} \qquad \qquad
			\includegraphics[width=\fWidRTex,height=\fHeiRTex]{{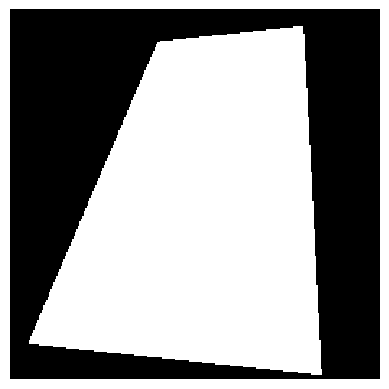}}\qquad \qquad
			\includegraphics[width=\fWidRTex,height=\fHeiRTex]{{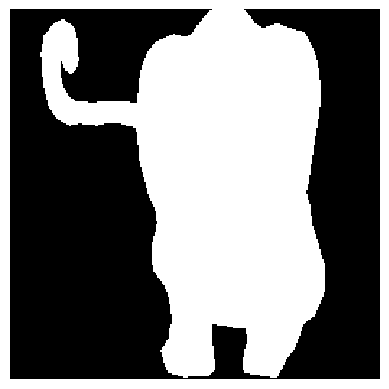}}\qquad \qquad
			\includegraphics[width=\fWidRTex,height=\fHeiRTex]{{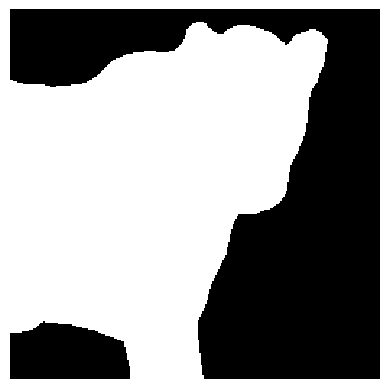}}\qquad \qquad
			\includegraphics[width=\fWidRTex,height=\fHeiRTex]{{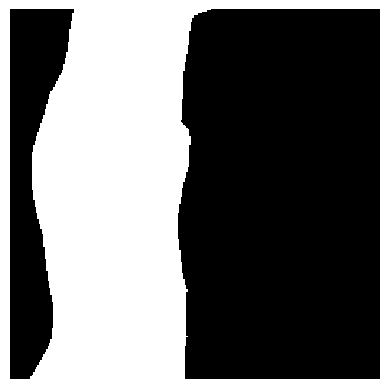}}\qquad \qquad
			
			\includegraphics[width=\fWidRTex,height=\fHeiRTex]{{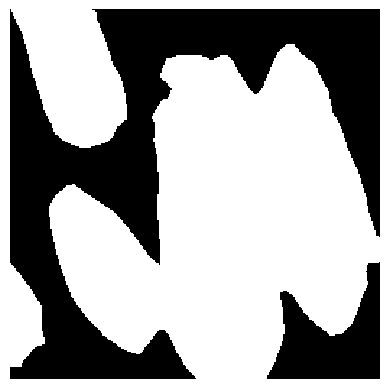}}\qquad \qquad
			\includegraphics[width=\fWidRTex,height=\fHeiRTex]{{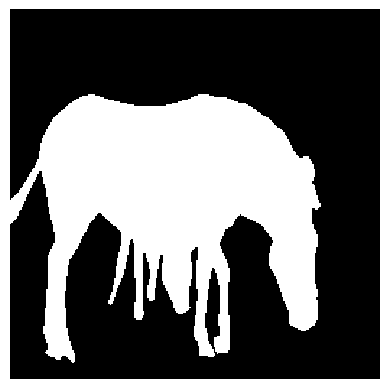}}\\ \\ 

			\Huge DeepLab-a-CE \\
			\includegraphics[width=\fWidRTex,height=\fHeiRTex]{{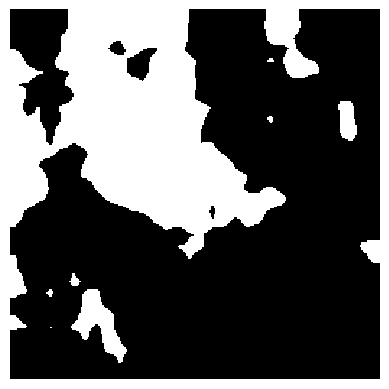}} \qquad \qquad
			\includegraphics[width=\fWidRTex,height=\fHeiRTex]{{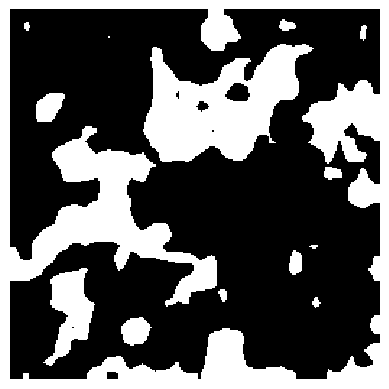}}\qquad \qquad
			\includegraphics[width=\fWidRTex,height=\fHeiRTex]{{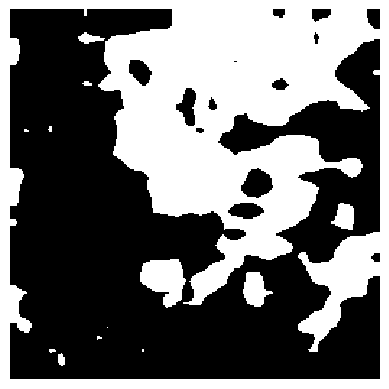}}\qquad \qquad
			\includegraphics[width=\fWidRTex,height=\fHeiRTex]{{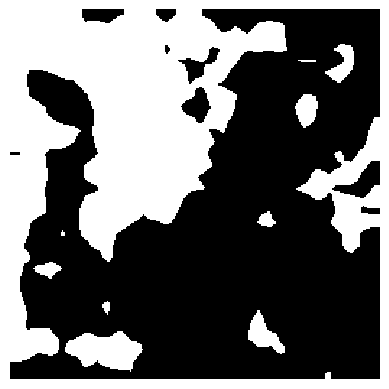}}\qquad \qquad
			\includegraphics[width=\fWidRTex,height=\fHeiRTex]{{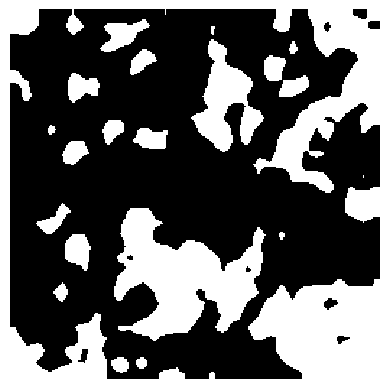}}\qquad \qquad
			
			\includegraphics[width=\fWidRTex,height=\fHeiRTex]{{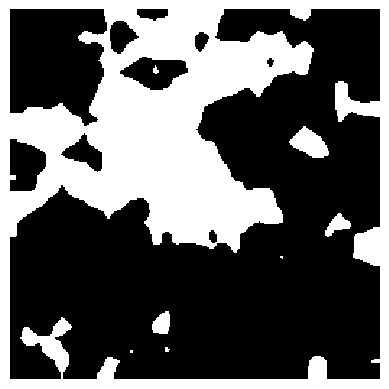}}\qquad \qquad
			\includegraphics[width=\fWidRTex,height=\fHeiRTex]{{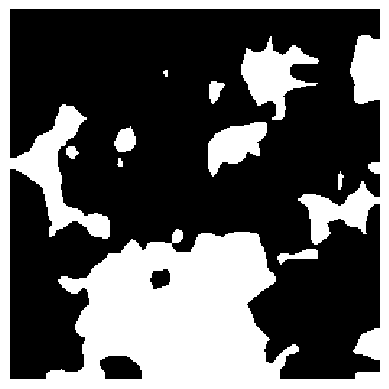}}\\ \\

			\Huge DeepLab-a-CAS \\
			\includegraphics[width=\fWidRTex,height=\fHeiRTex]{{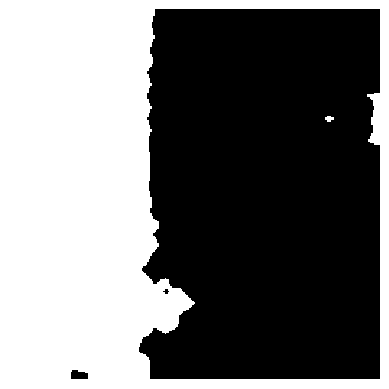}} \qquad \qquad
			\includegraphics[width=\fWidRTex,height=\fHeiRTex]{{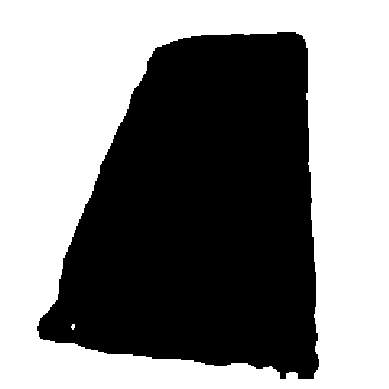}}\qquad \qquad
			\includegraphics[width=\fWidRTex,height=\fHeiRTex]{{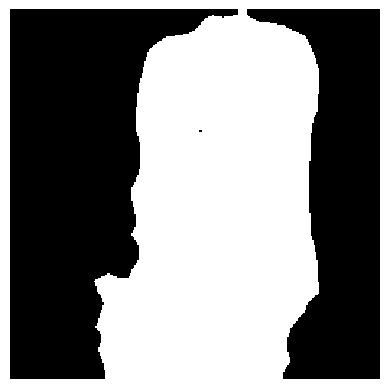}}\qquad \qquad
			\includegraphics[width=\fWidRTex,height=\fHeiRTex]{{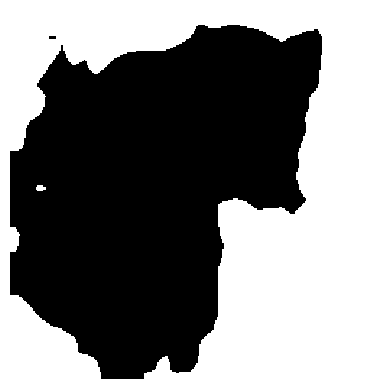}}\qquad \qquad
			\includegraphics[width=\fWidRTex,height=\fHeiRTex]{{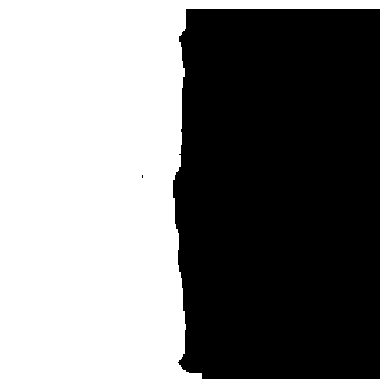}}\qquad \qquad
			
			\includegraphics[width=\fWidRTex,height=\fHeiRTex]{{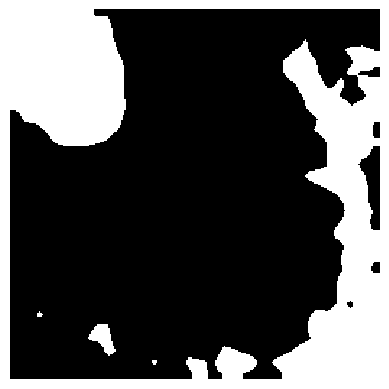}}\qquad \qquad
			\includegraphics[width=\fWidRTex,height=\fHeiRTex]{{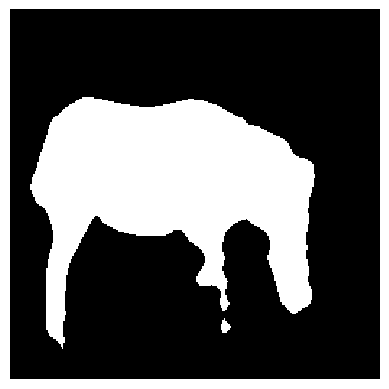}}\\ \\ \\

			\Huge Images \\
			\includegraphics[width=\fWidRTex,height=\fHeiRTex]{{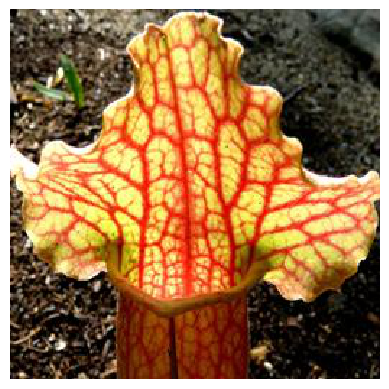}} \qquad \qquad
			\includegraphics[width=\fWidRTex,height=\fHeiRTex]{{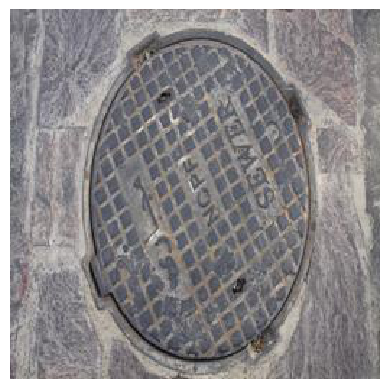}}\qquad \qquad
			\includegraphics[width=\fWidRTex,height=\fHeiRTex]{{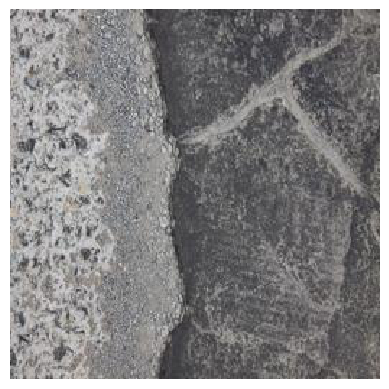}}\qquad \qquad
			\includegraphics[width=\fWidRTex,height=\fHeiRTex]{{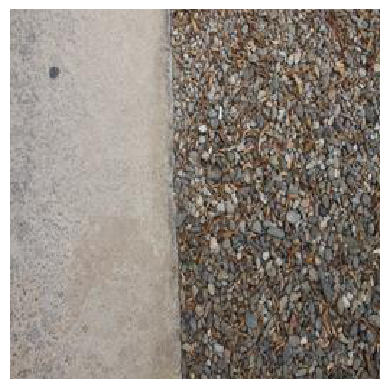}}\qquad \qquad
			
			\includegraphics[width=\fWidRTex,height=\fHeiRTex]{{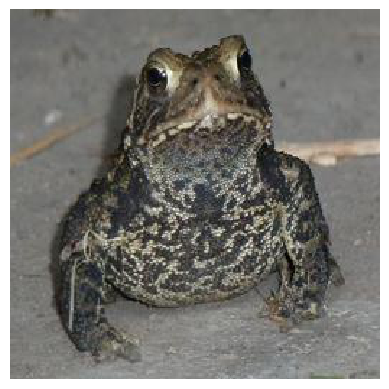}}\qquad \qquad
			\includegraphics[width=\fWidRTex,height=\fHeiRTex]{{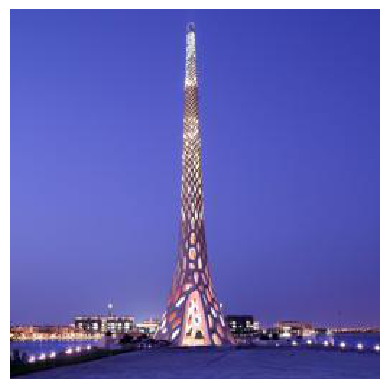}} \qquad \qquad 
			\includegraphics[width=\fWidRTex,height=\fHeiRTex]{{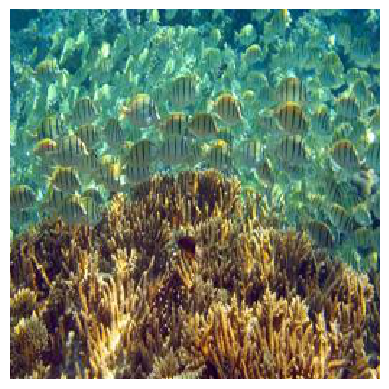}}\\ \\
			
			\Huge Ground Truth\\ 
			
			\includegraphics[width=\fWidRTex,height=\fHeiRTex]{{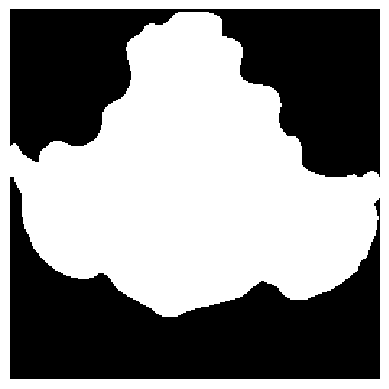}}\qquad \qquad
			\includegraphics[width=\fWidRTex,height=\fHeiRTex]{{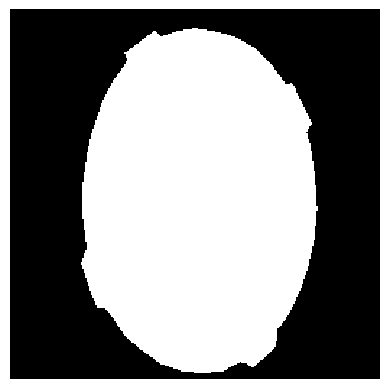}}\qquad \qquad
			\includegraphics[width=\fWidRTex,height=\fHeiRTex]{{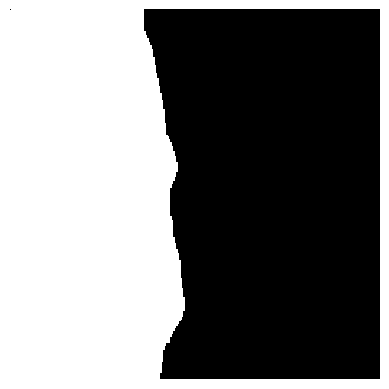}}\qquad \qquad
			\includegraphics[width=\fWidRTex,height=\fHeiRTex]{{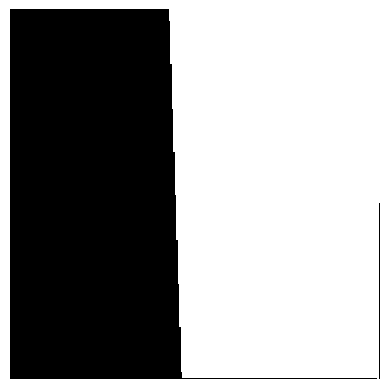}}\qquad \qquad
			\includegraphics[width=\fWidRTex,height=\fHeiRTex]{{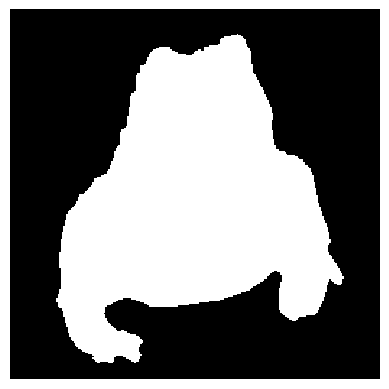}}\qquad \qquad
			
			\includegraphics[width=\fWidRTex,height=\fHeiRTex]{{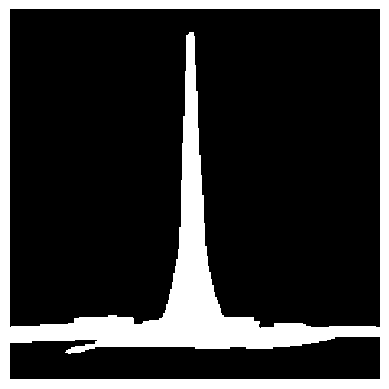}} \qquad \qquad
			\includegraphics[width=\fWidRTex,height=\fHeiRTex]{{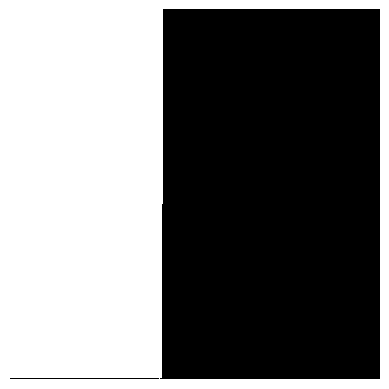}} \\ \\
			
			\Huge DeepLab-a-CE \\
			\includegraphics[width=\fWidRTex,height=\fHeiRTex]{{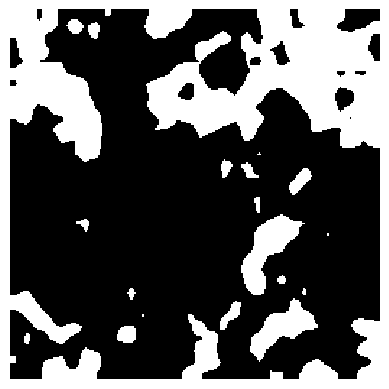}}\qquad \qquad
			\includegraphics[width=\fWidRTex,height=\fHeiRTex]{{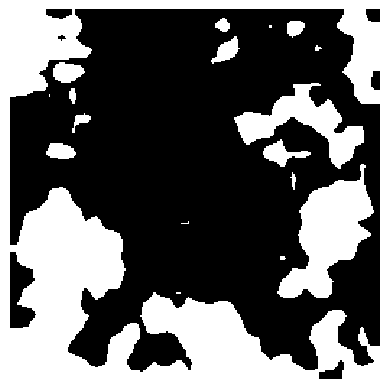}}\qquad \qquad
			\includegraphics[width=\fWidRTex,height=\fHeiRTex]{{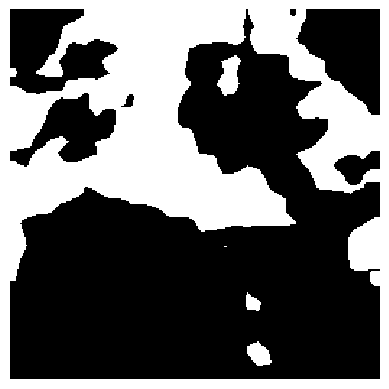}}\qquad \qquad
			\includegraphics[width=\fWidRTex,height=\fHeiRTex]{{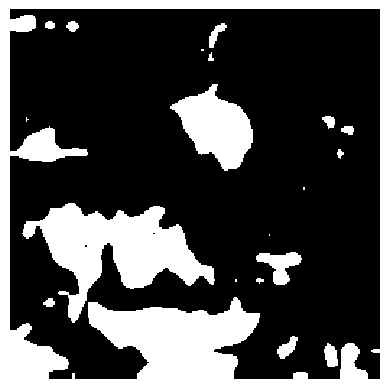}}\qquad \qquad
			\includegraphics[width=\fWidRTex,height=\fHeiRTex]{{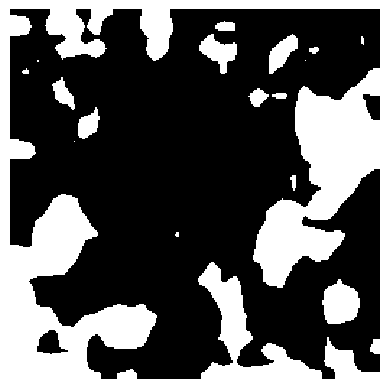}}\qquad \qquad
			
			\includegraphics[width=\fWidRTex,height=\fHeiRTex]{{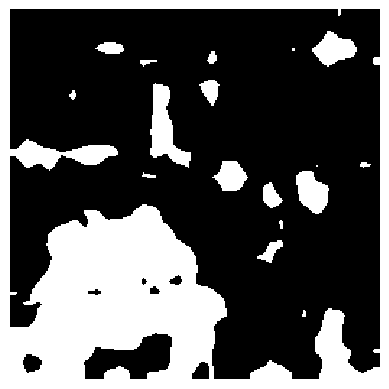}} \qquad \qquad
			
			\includegraphics[width=\fWidRTex,height=\fHeiRTex]{{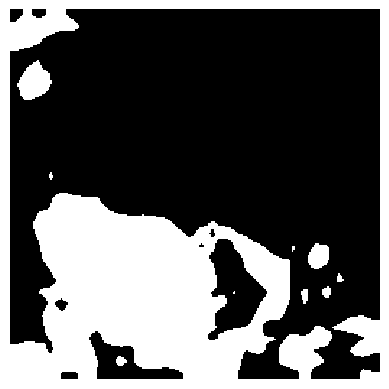}}  \\ \\
			
			\Huge DeepLab-a-CAS \\
			\includegraphics[width=\fWidRTex,height=\fHeiRTex]{{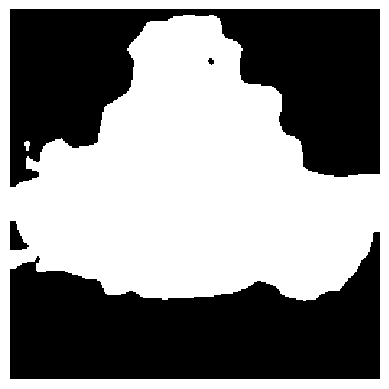}}\qquad \qquad
			\includegraphics[width=\fWidRTex,height=\fHeiRTex]{{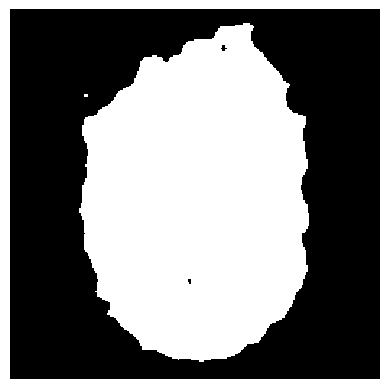}}\qquad \qquad
			\includegraphics[width=\fWidRTex,height=\fHeiRTex]{{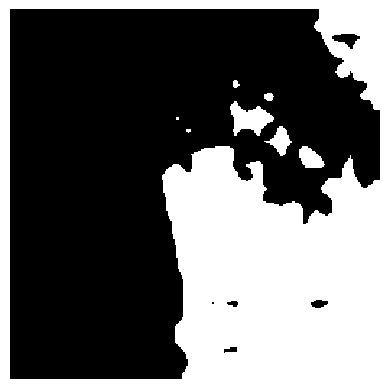}}\qquad \qquad
			\includegraphics[width=\fWidRTex,height=\fHeiRTex]{{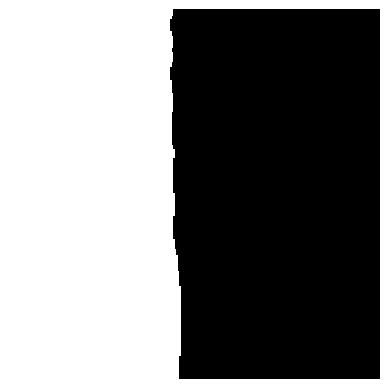}}\qquad \qquad
			\includegraphics[width=\fWidRTex,height=\fHeiRTex]{{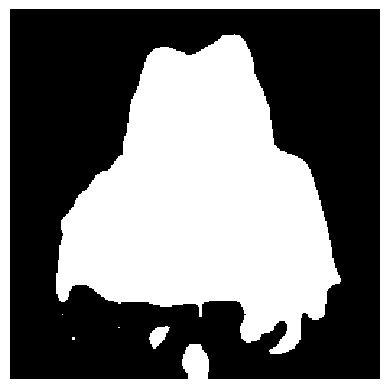}}\qquad \qquad
			
			\includegraphics[width=\fWidRTex,height=\fHeiRTex]{{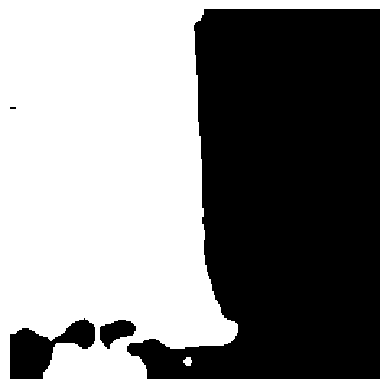}} \qquad \qquad
			
			\includegraphics[width=\fWidRTex,height=\fHeiRTex]{{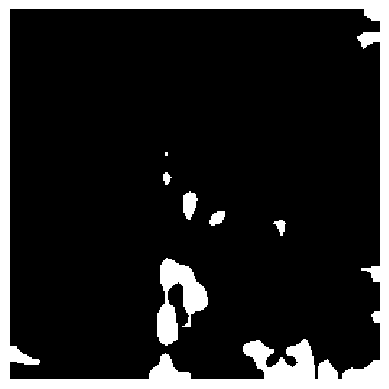}}
		\end{tabular}
		
	\end{adjustbox}
	
	\caption{{\bf Sample representative results on Real-World
			Texture Dataset}: Visual results for texture segmentation experiments, using DeepLab architecture; -a- denotes trained on the 7 saliency datasets and texture dataset }
	
	\label{fig:text}
\end{figure*}

\end{document}